\documentclass[acmtog]{acmart}

\AtBeginDocument{%
  \providecommand\BibTeX{{%
    \normalfont B\kern-0.5em{\scshape i\kern-0.25em b}\kern-0.8em\TeX}}}

\usepackage{siunitx}
\usepackage[ruled,vlined]{algorithm2e}
\usepackage{bm}
\usepackage{multirow}
\usepackage[overload]{empheq} %
\usepackage{wrapfig}
\usepackage{colortbl}

\definecolor{applegreen}{rgb}{0.44, 0.71, 0.0}

\newcommand{\x}[0]{\mathbf{x}}
\newcommand{\vel}[0]{\mathbf{v}}
\newcommand{\R}[0]{\mathcal{R}}

\newcommand{\M}[0]{\mathbf{M}}
\newcommand{\fin}[0]{\mathbf{f}_{\text{int}}}
\newcommand{\fe}[0]{\mathbf{f}_{\text{ext}}}
\newcommand{\y}[0]{\mathbf{y}}
\newcommand{\z}[0]{\mathbf{z}}
\newcommand{\bvec}[0]{\mathbf{b}}
\newcommand{\0}[0]{\mathbf{0}}
\newcommand{\p}[0]{\mathbf{p}}
\newcommand{\Gmat}[0]{\mathbf{G}}
\newcommand{\F}[0]{\mathbf{F}}
\newcommand{\A}[0]{\mathbf{A}}
\newcommand{\I}[0]{\mathcal{I}}

\newcommand{\Cset}[0]{\mathcal{C}}
\newcommand{\rf}[0]{\mathbf{r}}
\newcommand{\Sset}[0]{\mathbf{S}}
\newcommand{\avec}[0]{\mathbf{a}}
\newcommand{\Cc}[0]{\overline{\mathcal{C}}}
\newcommand{\Pmat}[0]{\mathbf{P}}
\newcommand{\Umat}[0]{\mathbf{U}}
\newcommand{\Vmat}[0]{\mathbf{V}}
\newcommand{\Imat}[0]{\mathbf{I}}
\newcommand{\e}[0]{\mathbf{e}}
\newcommand{\B}[0]{\mathbf{B}}

\usepackage{tikz}

\def\twowords#1 #2\relax{%
    \def\imgtrimwidth{#1}%
    \def\imgtrimheight{#2}%
}
\def\fourwords#1 #2 #3 #4\relax{
    \def\imgtrimleft{#1}%
    \def\imgtrimbottom{#2}%
    \def\imgtrimright{#3}%
    \def\imgtrimtop{#4}%
}

\newcommand{\plantimgwithbox}[4]{%
\begin{tikzpicture}%
\node[anchor=south west,inner sep=0](image) at (0,0){\includegraphics[trim=140 120 150 70,clip,width=#1]{#4}};%
\begin{scope}[x={(image.south east)},y={(image.north west)}]%
\definecolor{boxcolor1}{RGB}{#2}%
\definecolor{boxcolor2}{RGB}{#3}%
\draw[rounded corners=0pt,color=boxcolor1,thick] (0.5, 0.65) rectangle (0.8, 0.99);%
\draw[rounded corners=0pt,color=boxcolor2,thick] (0.05, 0.15) rectangle (0.35, 0.5);%
\end{scope}
\end{tikzpicture}
}
\begin{document}

\title{DiffPD: Differentiable Projective Dynamics}

\author{Tao Du}
\email{taodu@csail.mit.edu}
\orcid{0000-0001-7337-7667}

\author{Kui Wu}
\email{kuiwu@csail.mit.edu}

\author{Pingchuan Ma}
\email{pcma@csail.mit.edu}

\author{Sebastien Wah}
\email{sebwah@mit.edu}

\author{Andrew Spielberg}
\email{aespielberg@csail.mit.edu}

\affiliation{%
  \institution{MIT CSAIL}
}

\author{Daniela Rus}
\email{rus@csail.mit.edu}

\author{Wojciech Matusik}
\email{wojciech@csail.mit.edu}
\affiliation{%
    \institution{MIT CSAIL}
}

\renewcommand{\shortauthors}{Du et al.}

\begin{abstract}
We present a novel, fast differentiable simulator for soft-body learning and control applications. Existing differentiable soft-body simulators can be classified into two categories based on their time integration methods: Simulators using explicit time-stepping schemes require tiny time steps to avoid numerical instabilities in gradient computation, and simulators using implicit time integration typically compute gradients by employing the adjoint method and solving the expensive linearized dynamics. Inspired by Projective Dynamics (PD), we present Differentiable Projective Dynamics (DiffPD), an efficient differentiable soft-body simulator based on PD with implicit time integration. The key idea in DiffPD is to speed up backpropagation by exploiting the prefactorized Cholesky decomposition in forward PD simulation. In terms of contact handling, DiffPD supports two types of contacts: a penalty-based model describing contact and friction forces and a complementarity-based model enforcing non-penetration conditions and static friction. We evaluate the performance of DiffPD and observe it is 4-19 times faster compared with the standard Newton's method in various applications including system identification, inverse design problems, trajectory optimization, and closed-loop control. We also apply DiffPD in a reality-to-simulation (real-to-sim) example with contact and collisions and show its capability of reconstructing a digital twin of real-world scenes.
\end{abstract}




\maketitle

\section{Introduction}

The recent surge of differentiable physics witnessed the emergence of differentiable simulators as well as their success in various inverse problems that have simulation inside an optimization loop. With additional knowledge of gradients, a differentiable simulator provides more guidance on the evolution of a physics system. This extra information, when properly combined with mature gradient-based optimization techniques, facilitates the quantitative study of various downstream applications, e.g., system identification or parameter estimation, motion planning, controller design and optimization, and inverse design problems.

In this work, we focus on the problem of developing a differentiable simulator for soft-body dynamics. Despite its potential in many applications, research on differentiable soft-body simulators is still in its infancy due to the large number of degrees of freedom (DoFs) in soft-body dynamics. One learning-based approach is to approximate the true soft-body dynamics by way of a neural network for fast, automatic differentiation~\cite{li2018learning,sanchez2020learning}. For these methods, the simulation process is no longer physics-based but purely based on a neural network, which might lead to physically implausible and uninterpretable results and typically do not generalize well.

Another line of research, which is more physics-based, is to differentiate the governing equations of soft-body dynamics directly~\cite{hu2019chainqueen,Geilinger2020add,hu2019difftaichi,hahn2019real2sim}. We classify these simulators into \emph{explicit} and \emph{implicit} simulators based on their time-stepping schemes. Explicit differentiable simulators implement explicit time integration in forward simulation and directly apply the chain rule to derive any gradients involved. While explicit differentiable simulation is fast to compute and straightforward to implement, its explicit nature requires a tiny time step to avoid numerical instability. Moreover, when deriving the gradients, an output value (typically a reward or an error metric) needs to be backpropagated through all time steps. Such a process requires the state at every time step to be stored in memory regardless of the time-stepping scheme. Therefore, explicit differentiable simulators typically consume orders of magnitude more memory than their implicit counterparts, and sophisticated schemes like checkpoints are needed to alleviate the memory issue~\cite{hu2019chainqueen}.

Unlike explicit differentiable simulators, implicit simulation enables a much larger time step. It is more robust numerically and much more memory-efficient during backpropagation. However, an implicit differentiable simulator typically implements Newton's method in forward simulation and the adjoint method during backpropagation~\cite{hahn2019real2sim,Geilinger2020add}, both of which require the expensive linearization of the soft-body dynamics. Even though techniques for expediting the \emph{forward} soft-body simulation with an implicit time-stepping scheme have been developed extensively, the corresponding \emph{backpropagation} process remains a bottleneck for downstream applications in inverse problems.

In this work, we present DiffPD, an efficient differentiable soft-body simulator that implements the finite element method (FEM) and an implicit time-stepping scheme with certain assumptions on the material and contact model. We draw inspiration from Projective Dynamics (PD)~\cite{bouaziz2014projective}, a fast and stable algorithm that can be used for solving implicit time integration with FEM when the elastic energy of the material model has a specific quadratic form. The key observation we share with PD is that the computation bottleneck in both forward simulation and backpropagation is due to the nonlinearity of soft-body dynamics. By decoupling nonlinearity in the system dynamics, PD proposes a global-local solver where the global step solves a prefactorized linear system of equations and the local step resolves the nonlinearities in the physics and can be massively parallelized. Previous work in PD has demonstrated its efficacy in forward simulation and has reported significant speedup over the classical Newton's method. Our core contribution is to establish that with proper linear algebraic reformulation, the same idea of nonlinearity decomposition from PD can be fully extended to backpropagation as well.

To support differentiable contact handling, we revisit contact models used in previous PD papers, many of which choose to implement a soft contact force based on a fictitious collision energy~\cite{wang2015chebyshev, wang2016descent, dinev2018fepr, dinev2018stabilizing, liu2017quasi, li2019skeletons, bouaziz2014projective}. DiffPD naturally supports such energy-based contact and friction models as they can be seamlessly integrated into the PD framework. One notable exception is~\citet{ly2020projective}, which solves dry frictional contact in the standard PD framework. We have also explored the possibility of making the dry frictional contact model differentiable in DiffPD. Using the fact that contact vertices must be on the soft body's surface, which typically have much fewer DoFs than the interior vertices, we present a novel solution combining Cholesky factorization and low-rank matrix update to supporting differentiable \emph{static} friction and non-penetration contact.

We demonstrate the efficacy of DiffPD in various 3D applications with up to nearly $30,000$ DoFs. These applications include system identification, initial state optimization, motion planning, and end-to-end closed-loop control optimization. We compare DiffPD with both explicit and implicit differentiable FEM simulations and observe DiffPD's forward and backward calculation is 4-19 times faster than Newton's method when assumptions in PD hold. Furthermore, we embed DiffPD as a differentiable layer in a deep learning pipeline for training closed-loop neural network controllers for soft robots and report a speedup of 9-11 times in wall-clock time compared with deep reinforcement learning (RL) algorithms. Finally, we show a \emph{reality-to-simulation (real-to-sim)} application that uses our differentiable simulator to reconstruct a collision event between two tennis balls from a video input, which we hope can inspire follow-up work to solve \emph{simulation-to-reality (sim-to-real)} problems in the future.

To summarize, our paper contributes the following:
\begin{itemize}
    \item A PD-based differentiable soft-body simulator that is significantly faster than differentiable simulators using the standard Newton's method;
    \item A differentiable collision handling algorithm that handles penalty-based contact and friction forces or complementarity-based non-penetration contact and static friction;
    \item Demonstrations of the efficacy of our method on a wide range of applications, including system identification, inverse design problems, motion planning, robotics control, and a real-to-sim experiment, using material models compatible with PD and the simplified contact model stated above.
\end{itemize}

\section{Related Work}\label{sec:related}
We review methods on differentiable simulators and PD with contact handling, followed by a short summary of soft robot design and control methods. We summarize the differences between our work and previous papers in Table~\ref{tab:compare_methods}.

\begin{table*}
\caption{Summary of related work. The ``Diff.'' column refers to the differentiability of each method. In each column, green indicates the preferred property; red means the method lacks the preferred property; yellow means the method has some, but not all aspects of the preferred property.}
\label{tab:compare_methods}
\newcommand{\bad}{\cellcolor{red!20}}
\newcommand{\ok}{\cellcolor{yellow!20}}
\newcommand{\good}{\cellcolor{green!20}}
\newcommand{\non}{\cellcolor{white!20}}
\newcommand{\checked}{\cellcolor{green}}
\vspace{-1em}
\scalebox{0.89}{
\begin{tabular}{c|ccccccc}
\toprule
\non \textbf{Method}&\non\textbf{Domain}&\non\textbf{Integration}&\non\textbf{Physics-based}&\non\textbf{Speed}&\non\textbf{Memory}&\non\textbf{Diff.}&\non\textbf{Contact handling}\\
\midrule
\non ~\citet{hu2019chainqueen} & \good Soft body (MPM) & \ok Explicit & \good Yes & \good Fast & \bad High & \good Yes & \ok Solved on an Eulerian grid  \\
\non ~\citet{li2018learning} & \good Soft body (Particles) & \bad Network & \bad No & \good Fast & \ok Medium & \good Yes & \bad Trained neural-network function \\ 
\non ~\citet{liang2019differentiable} & \ok Cloth & \good Implicit & \good Yes & \bad Slow & \good Low & \good Yes & \ok Non-penetration + impulse-based friction \\
\non ~\citet{hahn2019real2sim} & \good Soft body (FEM) & \good Implicit & \good Yes & \bad Slow & \good Low & \good Yes & \ok Soft contact + penalty energy  \\ 
\non ~\citet{Geilinger2020add} & \good Multi body & \good Implicit & \good Yes & \bad Slow & \good Low & \good Yes & \ok Penalty-based frictional contact model  \\ 
\non ~\citet{ly2020projective} & \ok Cloth & \good Implicit & \good Yes & \good Fast & \good Low & \bad No & \good  Signorini-Coulomb contact model  \\ 
\non DiffPD (ours) & \good Soft body (FEM) & \good Implicit & \good Yes & \good Fast & \good Low & \good Yes & \ok Non-penetration contact + static-friction \\
\bottomrule
\end{tabular}
}
\end{table*}

\paragraph{Differentiable simulation}
Many recent advances in differentiable physics facilitate the application of gradient-based methods in robotics learning, control, and design tasks. Several differentiable simulators have been developed for rigid-body dynamics~\cite{popovic2003motion, de2018end,toussaint2018differentiable,degrave2019differentiable,Macklin2020PrimalDual}, soft-body dynamics~\cite{hu2019chainqueen,hu2019difftaichi,hahn2019real2sim,Geilinger2020add}, cloth~\cite{liang2019differentiable,qiao2020scalable}, and fluid dynamics~\cite{treuille2003keyframe, mcnamara2004fluid, wojtan2006keyframe, schenck2018spnets,holl2020learning}. Here we mainly discuss methods for soft-body dynamics as it is the focus of our work. ChainQueen~\cite{hu2019chainqueen} and its follow-up work DiffTaichi~\cite{hu2019difftaichi} introduce differentiable physics-based soft-body simulators using the material point method (MPM), which uses particles to keep track of the full states of the dynamical system and solves the momentum equations as well as collisions on a background Eulerian grid. However, the explicit time integration in their methods requires small time steps to preserve numerical stability, leading to large memory consumption during backpropagation. To resolve this issue, ChainQueen proposes to cache checkpoint steps in memory and reconstructs the states by rerunning forward simulation during backpropagation, which increases implementation complexity and introduces extra time cost. Further, solving collisions on an Eulerian grid may introduce artifacts depending on the resolution of the grid~\cite{han2019Hybrid}. Another particle-based strategy is to approximate soft-body dynamics with graph neural networks~\cite{li2018learning,sanchez2020learning}, which is naturally equipped with differentiability but may result in physically implausible behaviors. Our work is most similar to \citet{hahn2019real2sim} and~\citet{Geilinger2020add}, which use differentiable implicit FEM simulators. Compared with other differetiable deformable body simulators~\cite{hahn2019real2sim, Geilinger2020add, liang2019differentiable, qiao2020scalable} that systematically apply a combination of the adjoint method and sensitivity analysis to derive gradients, DiffPD puts a special focus on a more strategic backpropagation scheme that leverages the unique structure in Projective Dynamics, leading to empirically faster gradient computation. Finally, DiffPD is also relevant to~\citet{Li2020IPC} that introduce a contact-aware Newton-type solver and handle contact robustly with a differentiable barrier function for soft-body simulation. Although~\citet{Li2020IPC} do not discuss gradient computation or present applications that benefit from these gradients, they show that the whole contact model is differentiable in theory, from which we believe applications using differentiable simulation in the future can benefit a lot.

\paragraph{Projective Dynamics} PD was originally proposed by~\citet{bouaziz2014projective} as an attractive alternative to Newton's method for solving nonlinear system dynamics with an implicit time-stepping scheme. The standard PD algorithm has been extended to support rigid bodies~\cite{li2019skeletons}, conserve kinetic energy~\cite{dinev2018stabilizing,dinev2018fepr}, and support a wide range of hyperelastic materials~\cite{liu2017quasi}. Furthermore, prior papers have also proposed more advanced acceleration strategies including semi-iterative Chebyshev solvers~\cite{wang2015chebyshev,wang2016descent}, parallel Gauss-Seidel methods with randomized graph coloring~\cite{Fratarcangeli2016Vivace}, precomputed reduced subspace methods for the required constraint projections~\cite{brandt2018hyper}, and multigrid solvers~\cite{xian2019multigrid}. \citet{Macklin2020PrimalDual} introduce a preconditioned descent-based method of PD on GPUs with a penalty-based contact model. In DiffPD, we implement their penalty-based contact model but leave the GPU acceleration as future work. On the theoretical side, \citet{narain2016admm} and~\citet{overby2017admm} interpret PD as a special case of the alternating direction method of multipliers (ADMM) from the optimization perspective. Our work inherits the standard PD framework from previous papers and augments it with gradient computation.

\paragraph{Contact handling} The topic of handling contact and friction has been extensively studied in soft-body simulation. There exists a diverse set of collision handling algorithms with different design consideration: physical plausibility, time cost, and implementation complexity, although only a few of them take differentiability into consideration~\cite{Li2020IPC,chen2017dynamics,hu2019chainqueen,Macklin2020PrimalDual}. In the realm of PD simulation with contact, the most widespread strategy is to treat contact as soft constraints. This is achieved by either directly projecting colliding vertices onto collision surfaces at the end of each simulation step~\cite{wang2015chebyshev, wang2016descent, dinev2018fepr, dinev2018stabilizing} or imposing a fictitious collision energy~\cite{liu2017quasi, li2019skeletons, bouaziz2014projective}. Such methods introduce an artificial stiffness coefficient, which is task-dependent and requires careful tuning. Alternatively,~\citet{ly2020projective} propose to combine the more physically plausible Signorini-Coulomb contact model with PD in forward simulation. However, using such a model creates an additional challenge during backpropagation as solving the contact forces requires nontrivial iterative optimizers which either fail to maintain differentiability or cannot exploit the prefactorized Cholesky decomposition. In our work, we consider both penalty-based and complementarity-based contact models. For complementarity-based contact, we support non-penetration contact and static friction and leave sliding friction as future work. At the core of our contact handling algorithm is the assumption that contact occurs on a small fraction of the full DoFs, leading to a low-rank update on the linear system of equations. The idea of leveraging incremental, low-rank updates during contact handling can be traced back to ArtDefo~\cite{james1999artdefo}, which simulates soft bodies with the boundary element methods and resets three columns in a matrix when a new contact point becomes active. In our work, we use a similar idea to solve contact during forward simulation and extend it to compute gradients during backpropagation. Our work is also relevant to the recent work on sparse Cholesky updates~\cite{herholz2020sparse}, which provides an alternative to our low-rank update strategy based on Woodbury matrix identity.

\paragraph{Soft robot design and control}
As soft-body simulation becomes faster, more robust, and more expressive in the way of gradient information, the field of computational soft robotics is emerging as a solution to soft robot design and control problems. One of the earliest such simulators applied to computational robotics is VoxCAD \cite{hiller2014dynamic}, which is not differentiable but runs quickly on CPUs. This feature makes it well-catered to genetic algorithms, leading to computer-designed soft robots that can walk, swim, and grow \cite{cheney2013unshackling, corucci2016evolving, bongard2016material}. More recently, \citet{min2019softcon} have presented an FEM simulator which couples soft bodies with a hydrodynamical model to enable simulation and training of swimming creatures. Like VoxCAD, this simulator is a gradient-free approach, meaning training such controllers can take hours or days.

Competing with these gradient-free approaches are gradient-based approaches, which are typically more efficient due to the additional information provided by gradients. \citet{sin2013vega} and \citet{allard2007sofa} have presented fast, partially differentiable FEM simulation that exploits modal dynamics to accelerate simulation, at the cost of accuracy. Still, these approaches are robust enough to allow interactive simulation of soft characters \cite{barbivc2005real} and control of simulated and real robots \cite{thieffry2018control}. More recently, similar finite-element-based simulation techniques are shown to be effective in generating gaits for real-world foam quadrupeds \cite{bern2019trajectory}.  In each of these works, only control is examined.  Similar to our work, \citet{lee2018dexterous} demonstrate how PD can be applied to dexterous manipulation using an actuation model based on the human muscular system.  While their framework is flexible and can handle soft-rigid coupling, it can only be applied to quasi-static motions. By contrast, our simulator handles fully dynamical systems. A competing approach, differentiable MPM, is applied to soft robotics in \citet{spielberg2019learning}, co-designing soft robots over closed- and open-loop control and material distributions.

\section{Background}\label{sec:background}

In this section, we review the basic concepts of the implicit time-stepping scheme and PD. Let $n$ be the number of 3D nodes in a deformable body after FEM-discretization. After time discretization, we use $\x_i\in\R^{3n}$ and $\vel_i\in\R^{3n}$ to indicate the nodal positions and velocities at the $i$-th time step.

\paragraph{Implicit time integration} In this paper, we focus on the implicit time integration:
\begin{align}
    \x_{i+1} = &\; \x_{i} + h\vel_{i+1} \label{eq:implicit_x},\\
    \vel_{i+1} = &\; \vel_i + h\M^{-1}[\fin(\x_{i+1}) + \fe] \label{eq:implicit_v},
\end{align}
where $h$ is the time step, $\M\in\R^{3n\times3n}$ a lumped mass matrix, and  
$\fin$ and $\fe$ the sum of the internal and external forces, respectively. Substituting $\vel_{i+1}$ in $\x_{i+1}$ gives the following nonlinear system of equations:
\begin{align}\label{eq:time_integration}
    \frac{1}{h^2}\M(\x_{i+1}-\y_i) - \fin(\x_{i+1}) = \0,
\end{align}
where $\y_i=\x_i+h\vel_i+h^2\M^{-1}\fe$ is evaluated at the beginning of each time step. We drop the indices from $\x$ and $\y$ for simplicity:
\begin{align}\label{eq:time_integration_no_indices}
    \frac{1}{h^2}\M(\x-\y)-\fin(\x)=\0.
\end{align}
At each time step, our goal is to find $\x$ satisfying the equation above with the given $\y$.

As pointed out by~\citet{stuart1996dynamical} and ~\citet{martin2011example}, solving $\x$ from Eqn. (\ref{eq:time_integration_no_indices}) is equivalent to finding the critical point of the following objective $g$:
\begin{align}
    g(\x)=\frac{1}{2h^2} (\x-\y)^\top\M(\x-\y)+E(\x), \label{eq:opt}
\end{align}
where $E$ is the potential energy that induces the internal force: $\fin=-\nabla E$. It is easy to check the left-hand side of Eqn. (\ref{eq:time_integration_no_indices}) is $\nabla g$:
\begin{align}
    \nabla g(\x) = \frac{1}{h^2}\M(\x -\y) + \nabla E(\x) = \frac{1}{h^2}\M(\x -\y) - \fin(\x).
\end{align}

Eqn. (\ref{eq:time_integration_no_indices}) is typically solved with Newton's method, which iteratively solves a series of linear systems of equations. Consider the $k$-th iteration in Newton's method with $\x^k$ being the guess on $\x$ so far. Newton's method computes the next guess on $\x$ as follows:
\begin{align}
    \bm{0}=\nabla g(\x)=\nabla g(\x^k+\Delta \x)\approx & \nabla g(\x^k)+ \nabla^2 g(\x^k) \Delta \x, \label{eq:newton_update1}\\
    \nabla^2 g(\x^k)\Delta\x\approx & -\nabla g(\x^k). \label{eq:newton_update2}
\end{align}
Therefore, one can let $\Delta \x=-[\nabla^2 g(\x^k)]^{-1} \nabla g(\x^k)$ and update their guess on $\x$ at the next iteration by $\x^{k+1}=\x^k+\Delta \x$. In practice, Newton's method typically employs definiteness fixes or line searches when $\nabla^2 g$ is indefinite~\cite{nocedal2006numerical}. For large-scale problems, solving Eqn. (\ref{eq:newton_update2}) at each $\x^k$ requires expensive linearization and matrix factorization, which becomes the time bottleneck.

\paragraph{Backpropagation with implicit time integration} We sketch the main idea of backpropagation with a loss function $L$ defined on $\x$ and explain how we can compute $\frac{\partial L}{\partial \y}$ from $\frac{\partial L}{\partial \x}$. Backpropagating through multiple time steps can be done by backpropagating through every single pair of $(\x, \y)$ from each time step repeatedly. As $\x$ and $\y$ are implicitly constrained by $\nabla g(\x)=\0$, we can differentiate it with respect to $\y$ and obtain the following equation:
\begin{align}\label{eq:imp_func_theorem}
    \frac{\partial \nabla g(\x)}{\partial \x} \frac{\partial \x}{\partial \y} + \frac{\partial \nabla g(\x)}{\partial \y} = \frac{\partial }{\partial \y}\0, \\
    \frac{\partial}{\partial \x}[\frac{1}{h^2}\M (\x - \y) + \nabla E(\x)]\frac{\partial \x}{\partial \y} + \frac{\partial }{\partial \y}[\frac{1}{h^2}\M(\x - \y)] = \0, \\
    \underbrace{[\frac{1}{h^2}\M + \nabla^2 E(\x)]}_{\nabla^2 g(\x)} \frac{\partial \x}{\partial \y} - \frac{1}{h^2}\M = \0, \\
    \frac{\partial \x}{\partial \y} = \frac{1}{h^2}[\nabla^2 g(\x)]^{-1}\M
\end{align}
We can solve $\frac{\partial \x}{\partial \y}$ from it and use the chain rule to obtain the following (assuming both $\frac{\partial L}{\partial \x}$ and $\frac{\partial L}{\partial \y}$ are row vectors):
\begin{align}\label{eq:chain_rule}
    \frac{\partial L}{\partial \y} = \frac{\partial L}{\partial \x}\frac{\partial \x}{\partial \y} = \frac{1}{h^2}\underbrace{\frac{\partial L}{\partial \x}[\nabla^2 g(\x)]^{-1}}_{\z^\top}\M.
\end{align}
Note that the inverse of $\nabla^2 g(\x)$ is intentionally regrouped with $\frac{\partial L}{\partial \x}$ to avoid the expensive $[\nabla^2 g(\x)]^{-1}\M$. The adjoint vector $\z$ can be solved from the following linear system of equations:
\begin{align}\label{eq:adjoint}
\nabla^2 g(\x) \z = (\frac{\partial L}{\partial \x})^\top.
\end{align}
Note that we drop the transpose of $\nabla^2 g(\x)$ because it is symmetric.

Putting them together, we have shown that backpropagation within one time step can be done by Eqns. (\ref{eq:chain_rule}) and (\ref{eq:adjoint}). It is now clear that $\nabla^2 g(\x)$ plays a crucial role in both forward simulation and backpropagation, and we write its definition explicitly below:
\begin{align}\label{eq:hessian}
    \nabla^2 g(\x) = \frac{1}{h^2}\M + \nabla^2 E(\x).
\end{align}
Similar to Newton's method in forward simulation, a direct implementation of Eqn. (\ref{eq:adjoint}) is computationally expensive because $\nabla^2 g(\x)$ needs to be reconstructed and refactorized at every time step. This motivates us to propose the novel PD-based backpropagation method in Sec.~\ref{sec:diff_pd}.

\paragraph{Projective Dynamics} PD considers a specific family of quadratic potential energies that decouple the nonlinearity in material models~\cite{bouaziz2014projective}. Specifically, PD assumes the energy $E$ is the sum of quadratic energies taking the following form:
\begin{align}
    E_c(\x)= \min_{\p_c\in\mathcal{M}_c}\;\underbrace{\frac{w_c}{2}\|\Gmat_c\x - \p_c\|_2^2}_{\tilde{E}_c(\x, \p_c)}, \label{eq:pd_energy_single}\\
    E(\x) = \sum_c E_c(\x), \label{eq:pd_energy_total}
\end{align}
where $\Gmat_c$ is a discrete differential operator in the form of a constant sparse matrix, $w_c$ a scalar that determines the stiffness of the energy, and $\mathcal{M}_c$ a constraint manifold. For example, if one wants to formulate a volume-preserving elastic energy, $\mathcal{M}_c$ can be the set of all $3\times3$ matrices whose determinant is $1$.  $E_c$ is defined as the distance from $\Gmat_c\x$ to $\mathcal{M}_c$. Following the prevalent practice in previous work~\cite{bouaziz2014projective,liu2017quasi,min2019softcon}, we assume $E_c$ is defined on each finite element with $\mathbf{G}_c$ mapping $\x$ to the local deformation gradients $\F$~\cite{sifakis2012fem}.

With the definition of $E$ at hand, PD obtains the critical point of $g$ by alternating between a local step and a global step, which essentially minimizes the following surrogate objective $\tilde{g}(\x,\p)$:
\begin{align}
    \tilde{g}(\x,\p)=\frac{1}{2h^2}(\x-\y)^\top\M(\x-\y)+\sum_c\tilde{E}_c(\x,\p_c),
\end{align}
where $\p$ stacks up all $\p_c$ from each $E_c$. The local and global steps in PD can be interpreted as running coordinate descent optimization on $\tilde{g}$. The local step fixes the current $\x$ and projects $\Gmat_c\x$ onto $\mathcal{M}_c$ to obtain $\p_c$ in each $E_c$, which can be massively parallelizable across all $E_c$. The global step fixes $\p$ and minimizes $\tilde{g}$ over $\x$, which turns out to be a quadratic function with an analytical solution solved from the following linear system of equations:
\begin{align}\label{eq:pd_global}
    \underbrace{(\frac{1}{h^2}\M+\sum_c w_c \Gmat_c^\top\Gmat_c)}_{\A}\x=\frac{1}{h^2}\M\y+\sum_c w_c \Gmat_c^\top\p_c.
\end{align}
It is easy to see that each local and global step ensures $\tilde{g}$ is non-increasing. Since $\tilde{g}$ is bounded below by 0, PD guarantees to converge to a local minimum of $\tilde{g}$ satisfying the gradient condition $\nabla_\x\tilde{g}=\0$. Interestingly,~\citet{liu2017quasi} establishes that $\nabla_\x\tilde{g}=\nabla g$ upon convergence, confirming that the solution from PD is indeed a critical point of $g$ that solves the implicit time integration. We summarize the local-global solver in Alg.~\ref{alg:forward}, which serves as a basis for our contact handling algorithm to be described in Sec.~\ref{sec:contact:compl}.

\begin{algorithm}[tb]
\caption{PD forward simulation in one time step}
Input: $\y$\;
Output: $\x$ that satisfies Eqn. (\ref{eq:time_integration_no_indices})\;
Initialize $\x=\y$\;
\While{$\x$ not converged}{
$\p_c=\arg\min_{\p_c\in\mathcal{M}_c} \tilde{E}_c(\x,\p_c)$; \textcolor{applegreen}{\texttt{ // }Local step\;}
$\bvec=\frac{1}{h^2}\M\y+\sum_c w_c\Gmat_c^\top\p_c$\;
$\x=\A^{-1}\bvec$;\textcolor{applegreen}{\texttt{ // }Global step\;}
}
\label{alg:forward}
\end{algorithm}

The source of efficiency in \emph{forward} PD simulation lies in the fact that $\A$ in the global step is a constant, symmetric positive definite matrix. Therefore, the Cholesky factorization of $\A$ can be precomputed, after which each global step requires back-substitution only. In the next section, we will show that we can also use $\A$ in \emph{backpropagation} to obtain significant speedup.

\section{Differentiable Projective Dynamics}~\label{sec:diff_pd}

We now describe our PD-based backpropagation method. Our key observation is that the bottleneck in backpropagation lies in the computation of $\nabla^2 g(\x)$ in Eqn. (\ref{eq:adjoint}). Following the same idea as in forward PD simulation, we propose to decouple $\nabla^2 g(\x)$ into a global, constant matrix and a local, massively parallelizable nonlinear component. To see this point, we compute $\nabla^2 E$ using Eqns. (\ref{eq:pd_energy_single}) and (\ref{eq:pd_energy_total}):
\begin{align}
    \nabla E(\x) = & \sum_c w_c \Gmat_c^\top(\Gmat_c\x - \p_c), \label{eq:energy_grad}\\
    \nabla^2 E(\x) = & \sum_c w_c \Gmat_c^\top\Gmat_c - \sum_c w_c \Gmat_c^\top \frac{\partial \p_c}{\partial \x}. \label{eq:energy_hessian}
\end{align}
Note that in Eqn. (\ref{eq:energy_grad}), $\frac{\partial \p_c}{\partial \x}$ can be safely ignored according to the envelope theorem (see Appendix in~\cite{liu2017quasi}). According to Eqn. (\ref{eq:hessian}), $\nabla^2 g(\x)$ now becomes:
\begin{align}
    \nabla^2 g(\x) = \frac{1}{h^2} \M + \sum_c w_c \Gmat_c^\top\Gmat_c - \underbrace{\sum_c w_c \Gmat_c^\top \frac{\partial \p_c}{\partial \x}}_{\Delta\A} = \A - \Delta \A.\label{eq:matrix_split}
\end{align}
It is now clear that $\Delta\mathbf{A}$ is the source of nonlinearity in $\nabla^2 g$. The matrix splitting of $\nabla^2 g = \A - \Delta \A$ suggests the following iterative solver for Eqn. (\ref{eq:adjoint}):
\begin{align}\label{eq:iteration}
    \A\z^{k+1} = \Delta \A\z^{k} + (\frac{\partial L}{\partial \x})^\top,
\end{align}
where $k$ indicates the iteration number. Therefore, we propose a local-global solver for Eqn. (\ref{eq:adjoint}): at the $k$-th iteration, the local step computes $\Delta\A\z^k$ across all energies $E_c$, forming the right-hand vector in Eqn. (\ref{eq:iteration}). In the global step, we solve $\z^{k+1}$ by back-substituting $\A$. Note that $\A$ is the same constant matrix in forward PD simulation, so we can reuse the Cholesky factorization of $\A$. The source of efficiency in this local-global solver is similar to what PD proposes to speed up forward simulation: essentially, this local-global solver trades the expensive matrix assembly and factorization of $\nabla^2 g$ in Eqn. (\ref{eq:adjoint}) with iterations on a constant, prefactorized linear system of equations. We summarize our PD backpropagation algorithm in Alg.~\ref{alg:backprop}.

\begin{algorithm}[tb]
\caption{PD backpropagation in one time step}
Input: $\y$, $\x$ (already computed in forward simulation), and $\frac{\partial L}{\partial \x}$\;
Output: $\frac{\partial L}{\partial \y}$\;
Initialize $\z=\0$\;
\While{$\z$ not converged}{
$\bvec=\Delta\A\z + (\frac{\partial L}{\partial \x})^\top$; 
\textcolor{applegreen}{\texttt{ // }Local step parallelizing $\Delta \A\z$\;}
$\z=\A^{-1}\bvec$; \textcolor{applegreen}{\texttt{ // }Global step\;}
}
$\frac{\partial L}{\partial \y}=\frac{1}{h^2}\z^\top\M$;
\textcolor{applegreen}{\texttt{ // }Eqn. (\ref{eq:chain_rule})\;}
\label{alg:backprop}
\end{algorithm}

\paragraph{Convergence rate} For any iterative algorithm design, the immediate follow-up questions are whether such an algorithm is guaranteed to converge, and, if so, how fast the convergence rate is. To answer these questions, we use $(\frac{\partial L}{\partial \x})^\top=\A\z - \Delta\A\z$ and Eqn. (\ref{eq:iteration}) to obtain
\begin{align}\label{eq:sp_radius}
    \A(\z^{k+1}-\z)=\Delta\A(\z^k-\z),
\end{align}
from which we conclude the error at the $k$-th iteration is $\|\z^{k}-\z\|_2 = \|(\A^{-1}\Delta\A)^k(\z^0-\z)\|_2$. It follows that the iteration in Eqn. (\ref{eq:iteration}) is guaranteed to converge from any initial guess $\z^0$ if and only if $\rho(\A^{-1}\Delta\A)<1$, where $\rho(\cdot)$ indicates the spectral radius of a matrix. It is challenging to provide more theoretical results on $\rho(\A^{-1}\Delta\A)$ because it depends heavily on the specific form of $E_c$, which we leave as future work. In practice, we do not observe convergence issues with Eqn. (\ref{eq:iteration}) in any of our experiments, which seems to imply $\rho(\A^{-1}\Delta\A)<1$ is likely to be satisfied.

\paragraph{Further acceleration with Quasi-Newton methods}
Inspired by~\citet{liu2017quasi} which apply the quasi-Newton method to speed up forward PD simulation, we now show that a similar numerical optimization perspective can also be applied to speed up our proposed local-global solver in backpropagation. Solving Eqn. (\ref{eq:adjoint}) equals finding the critical point of the following energy $s(\z)$:
\begin{align}
    s(\z) = \frac{1}{2}\z^\top \nabla^2 g(\x)\z - \frac{\partial L}{\partial \x}\z.
\end{align}
It is easy to verify that $\nabla s(\z)=\0$ is essentially Eqn. (\ref{eq:adjoint}). We stress that in backpropagation both $\nabla^2 g(\x)$ and $\frac{\partial L}{\partial \x}$ are known values computed at $\x$ solved from forward simulation. If we apply Newton's method to this critical-point problem, the update rule will be as follows (see Eqns. (\ref{eq:newton_update1}) and (\ref{eq:newton_update2}) in Sec.~\ref{sec:background}):
\begin{align}
    \z^{k+1}=\z^k-[\nabla^2 s(\z^k)]^{-1}\nabla s(\z^k).
\end{align}
The true Hessian of $s(\z)$ is $\nabla^2 g(\x)=\A - \Delta\A$ from Eqn. (\ref{eq:matrix_split}). If we approximate it with $\A$, we get the following quasi-Newton update rule:
\begin{equation}
\begin{aligned}
    \z^{k+1} = &\;\z^k-\A^{-1}\nabla s(\z^k) \\
    = &\;\z^k - \A^{-1}[(\A - \Delta\A)\z^k - (\frac{\partial L}{\partial \x})^\top]\\
    = &\;\A^{-1}\Delta\A\z^k + \A^{-1}(\frac{\partial L}{\partial \x})^\top,
\end{aligned}
\end{equation}
which is identical to the iteration in Eqn. (\ref{eq:iteration}). As a result, the local-global solver we propose can be reinterpreted as running a simplified quasi-Newton method with a constant Hessian approximation $\mathbf{A}$. By applying a full quasi-Newton method, e.g., BFGS, we can reuse the Cholesky decomposition of $\mathbf{A}$ with little extra overhead of vector products and achieve a superlinear convergence rate~\cite{nocedal2006numerical}. Moreover, similar to the previous work~\cite{liu2017quasi}, we can apply line search techniques to ensure convergence when $\rho(\A^{-1}\Delta\A)\geq 1$, even though we do not experience convergence issues in practice.
\section{Contact Handling}\label{sec:contact}

We have described the basic framework of DiffPD in Sec.~\ref{sec:diff_pd}. In this section, we propose a novel method to incorporate contact handling and contact gradients into DiffPD. The challenges in developing such a contact handling algorithm are twofold: first, it must be compatible with our basic PD framework in \emph{both} forward simulation \emph{and} backpropagation. Second, it must support differentiability. In this section, we discuss two contact options that DiffPD supports: an explicit, penalty-based contact model with static and dynamic friction and an implicit, complementarity-based contact model supporting non-penetration conditions and static friction. Both options have their advantages and disadvantages: the penalty-based method is more straightforward to implement and easier to be integrated into a machine learning framework, e.g., as an explicit neural network layer in PyTorch. However, we find it typically requires a careful, scene-by-scene tuning of its parameters. On the other hand, our complementarity-based method does not rely on scene-dependent parameters, but it is currently limited to static friction only. Our penalty-based method is more suitable for tasks that favor speed and simplicity over physical accuracy, and the complementarity-based method is more useful when non-penetration conditions need to be strictly enforced and slipping motions are rare, e.g., simulating a wheeled robot.

\subsection{Penalty-Based Contact}\label{sec:contact:penalty}
Previous papers on PD simulation typically handle contact forces with a penalty-based soft contact model~\cite{dinev2018fepr,dinev2018stabilizing,wang2015chebyshev,wang2016descent,bouaziz2014projective,liu2017quasi}. One common way to model contact forces is to add an additional, fictitious energy $E_c$ with $\mathcal{M}_c$ being the contact surface and its exterior and $\Gmat_c$ being a matrix so that $\Gmat_c\x$ selects contact nodes from $\x$. This way, whenever a node penetrates the contact surface, $E_c$ exerts a contact force that attempts to push it back to the contact surface. As such a contact model can be seamlessly integrated into PD forward simulation, our backpropagation method in Sec.~\ref{sec:diff_pd} naturally supports it.

Handling static and dynamic frictional forces with a penalty-based model in PD is slightly trickier. Since friction is typically related to nodal velocities instead of nodal positions $\x$, it is not straightforward to find an $E_c$ that characterizes it. Therefore, instead of modeling friction with an additional $E_c$, we take the penalty-based frictional forces described in~\citet{Macklin2020PrimalDual} and add them directly to $\fe$. Deriving gradients with respect to such frictional forces is still straightforward as we can easily compute $\frac{\partial L}{\partial \fe}$ using the chain rule as before.%

\subsection{Complementarity-Based Contact}\label{sec:contact:compl}

An alternative to penalty-based contact is to model contact and friction using complementarity constraints~\cite{Macklin2020PrimalDual,ly2020projective}. Complementarity-based contact models are suitable for applications requiring high physical fidelity but typically require extra computational cost.~\citet{ly2020projective} present a general framework for handling complementarity-based contact and friction in PD forward simulation. More concretely,~\citet{ly2020projective} model contact and friction with the Signorini-Coulomb law and focus on applications in cloth simulation. Our approach is relevant to~\citet{ly2020projective} but has substantial difference because our focus in this paper is on 3D volumetric deformable bodies, which typically have a sparse contact set, i.e., the nodes in contact during simulation is usually a small portion of the full set of nodes. Below we will present a differentiable, complementarity-based contact model that leverages such sparsity to gain speedup in both forward simulation and backpropagation. Our contact model ensures non-penetration conditions and, in exchange for speedup, handles static friction only. We leave differentiable, complementarity-based dynamic friction model as future work.

\paragraph{Contact model}
Let $\phi(\cdot):\R^3\rightarrow\R$ be the signed-distance function of the contact surface with $\phi<0$ indicating the space occupied by the obstacle. We require the solution $\x$ to the implicit time integration to satisfy the following complementarity condition for any node indexed by $j$:
\begin{subequations}\label{eq:lcp}
\begin{align}[left = \empheqlbrace\,]
&\phi(\x_j)>0,\rf_j=\0, \label{eq:lcp_1} \\
&\text{or } \phi(\x_j)=0,\rf_{j|N}\geq0, \label{eq:lcp_2}
\end{align}
\end{subequations}
where $\x_j$ and $\rf_j$ are 3D vectors indicating the nodal position and contact force of node $j$. The notation $\rf_{j|N}\in\R$ is the normal component of $\rf_j$ where the normal is computed from the contact surface $\phi$ at the contact location $\x_j$. In other words, for each node $j$, it must be either above the contact surface ($\phi(\x_j)>0$) with zero contact force ($\rf_j=\0$) or in contact ($\phi(\x_j)=0$) with a positive contact force along the normal direction ($\rf_{j|N}=0$). The implicit time integration in Eqn. (\ref{eq:time_integration_no_indices}) now becomes:
\begin{subequations}\label{eq:time_integration_with_contact}
\begin{align}[left = \empheqlbrace\,]
&\frac{1}{h^2}\M(\x - \y) - \fin(\x)=\rf, \label{eq:tic_1} \\
&(\x,\rf)\text{ satisfy Eqn. (}\ref{eq:lcp}\text{)}, \label{eq:tic_2}
\end{align}
\end{subequations}
where the notation $\rf$ stacks up all contact force $\rf_j$ from each node $j$.

\paragraph{Remarks on friction} Eqn. (\ref{eq:time_integration_with_contact}) does not fully constrain the solution $\x$ because, for any $\x_j$ in contact with $\phi=0$, $\x_j$ can slide on $\phi=0$ and $\rf_j$ will compensate any force needed. This can be resolved by imposing additional location constraints on $\x_j$. Some common strategies include 1) in the penalty-based model before, $\x_j$ is chosen as certain projection onto $\phi=0$, 2) gluing $\x_j$ to its original position at the beginning of the time step, and 3) setting it to the contact point computed from collision detection~\cite{chen2017dynamics}, which is usually the intersection between $\phi=0$ and the ray $\x_j+t\vel_j,0\leq t\leq h$. Any of these strategies are compatible with DiffPD as long as they can compute a target location $\x_j^*$ on $\phi=0$ if a collision detection algorithm indicates $\x_j$ is in contact with $\phi=0$. In DiffPD, we choose the third strategy mentioned above, which essentially models a very sticky contact surface that provides infinitely large static friction once $\x_j$ is in contact.

\paragraph{Time integration with contact} We first introduce some notations to better explain our solver to Eqn. (\ref{eq:time_integration_with_contact}). Let $\I=\{0,1,2,\cdots,n-1\}$ be the indices of all $n$ nodes in the system. We use $\overline{\Sset}$ to denote the complement of a set $\Sset\subseteq\I$. In other words, $\Sset$ and $\overline{\Sset}$ is a two-set partition of $\I$. For any subsets $\Sset_r,\Sset_c\subseteq\I$, we use $\A_{\Sset_r\times\Sset_c}$ to indicate the submatrix of $\A$ created by keeping entries whose row and column indices are from nodes in $\Sset_r$ and $\Sset_c$, respectively. Similarly, for any vector $\avec$, we define $\avec_{\Sset}$ as the vector generated by keeping elements whose indices are from nodes in $\Sset$. For any vectors $\avec$ and $\bvec\in\R^{3|\Sset|}$, We use $\avec_{\Sset=\bvec}$ to indicate that $\avec$ satisfies $\avec_{\Sset}=\bvec$.

The high-level idea of our time integrator with the aforementioned contact model is described in Alg.~\ref{alg:contact}, which modifies Alg.~\ref{alg:forward} to find $\x$ that satisfies Eqn. (\ref{eq:time_integration_with_contact}). We start with any collision detection algorithm that can propose a set of candidate contact nodes $\Cset$ and compute a target location $\x_j^*$ for any $j\in\Cset$. Next, we use the proposed $\Cset$ to split the complementarity condition in Eqn. (\ref{eq:tic_2}) and solve Eqn. (\ref{eq:tic_1}): for any $j\in\Cset$, we set $\x_j=\x_j^*$; for any $j\notin\Cset$, we set $\rf_j=\0$. This makes Eqn. (\ref{eq:tic_1}) a balanced system with an equal number of equations and variables. Finally, we use the solved $\x$ to compute $\rf_j$ at each $j\in\Cset$ and check if $\rf_{j|N}\geq 0$ is satisfied. If $\x$ results in some negative $\rf_{j|N}$, these nodes are removed from $\Cset$, and a new iteration begins with the updated $\Cset$. Similarly, if $\phi(\x_j)$ becomes negative, such a node $j$ is added to $\Cset$. Essentially, we are running the active-set algorithm on Eqn. (\ref{eq:tic_1}) with linear constraints, and more advanced active set schemes can potentially be used to rebuild $\Cset$ more efficiently. Using the notations above, our algorithm attempts to solve the following reduced system at each iteration:
\begin{align} \label{eq:reduced}
    \frac{1}{h^2}\M_{\Cc\times\Cc}(\x - \y)_{\Cc}-\fin(\x_{\Cset=\x^*})_{\Cc}=&\;\0,
\end{align}
where $\x^*$ stacks up $\x_j^*$ for all $j\in\Cset$. Accordingly, the definition of $g$ is updated as follows, which we rename as $g_\Cset$:
\begin{align}\label{eq:gc}
    g_\Cset(\x_{\Cset=\x^*})=\frac{1}{2h^2}(\x-\y)_{\Cc}^\top\M_{\Cc\times\Cc}(\x-\y)_{\Cc} + E(\x_{\Cset=\x^*}).
\end{align}
It is easy to check that the left-hand side of Eqn. (\ref{eq:reduced}) is identical to $\nabla_{\x_{\Cc}} g_\Cset$. Therefore, solving Eqn. (\ref{eq:reduced}) is equal to finding the critical point of this modified $g$ function, and we can still apply Newton's method but with a slightly different definition of $\nabla^2 g$:
\begin{align}
    \nabla^2_{\x_{\Cc}} g_\Cset=\frac{1}{h^2}\M_{\Cc\times\Cc}+(\nabla^2 E)_{\Cc\times\Cc}=(\nabla^2 g)_{\Cc\times\Cc}.
\end{align}
In other words, $\nabla^2_{\x_{\Cc}}g_{\Cset}$ is a submatrix of $\nabla^2 g$ in Eqn. (\ref{eq:hessian}) created by deleting rows and columns from $\Cset$.

\paragraph{Implications on forward PD} In the PD framework, $g_{\Cset}$ also induces a modified surrogate function $\tilde{g}$, which we rename as $\tilde{g}_\Cset$:
\begin{equation}
\begin{aligned}
    \;&\tilde{g}_{\Cset}(\x_{\Cset=\x^*},\p) \\
    =\;&\frac{1}{2h^2}(\x-\y)_{\Cc}^\top\M_{\Cc\times\Cc}(\x-\y)_{\Cc} +\sum_c\tilde{E}_c(\x_{\Cset=\x^*},\p_c).
\end{aligned}
\end{equation}
It is still true that the original local-global solver will ensure $\tilde{g}_\Cset$ is non-increasing and converge to a critical point of $g_\Cset$. With the constraint $\x_\Cset=\x^*$, the local step can project each $\Gmat_c\x$ to obtain $\p_c$ as before. The global step, on the other hand, requires some modification, as can be best seen after computing $\nabla_{\x_{\Cc}}\tilde{g}_\Cset$:
\begin{align}
    \nabla_{\x_{\Cc}}\tilde{g}_\Cset=\frac{1}{h^2}\M_{\Cc\times\Cc}(\x-\y)_{\Cc}+\sum_c w_c(\Gmat_c^\top)_{\Cc\times\I}(\Gmat_c\x-\p_c).
\end{align}
Setting $\nabla_{\x_{\Cc}}\tilde{g}_{\Cc}=\0$ and using the fact that $\x_\Cset=\x^*$, we obtain the new global step with a linear system modified from Eqn. (\ref{eq:pd_global}):
\begin{align}
    \A_{\Cc\times\Cc}\x_{\Cc} = [\frac{1}{h^2}\M\y+\sum_c w_c\Gmat_c^\top(\p_c-\Gmat_c\x_{\Cset=\x^*,\Cc=\0})]_{\Cc},
\end{align}
where $\x_{\Cset=\x^*,\Cc=\0}$ is a vector satisfying $\x_{\Cset}=\x^*$ and $\x_{\Cc}=\0$. Although the right-hand side seems complicated, it can still be parallelized across all $E_c$. It is the left-hand side matrix $\A_{\Cc\times\Cc}$ that deserves more attention: since $\Cset$ is a set that changes dynamically between each time step, $\A_{\Cc\times\Cc}$ randomly erases different rows and columns from $\A$, which means the Cholesky factorization of $\A$ no longer applies. Our key observation is that $\Cset$ is usually a small subset of full nodes in 3D volumetric deformable bodies. This allows us to formulate row and column deletions on $\A$ as a \emph{low-rank update}, from which we derive efficient solvers that can reuse the Cholesky factorization of $\A$.

\begin{algorithm}[tb]
\caption{PD forward simulation with contact}
Input: $\y$\;
Output: $\x$ that satisfies Eqn. (\ref{eq:time_integration_with_contact})\;
Run a collision detection algorithm to get $\Cset$ and $\x^*$\;
\While{$\Cset$ not converged}{
Initialize $\x=\y$ and set $\x_{\Cset}=\x^*$\;
\While{$\x$ not converged}{
$\p_c=\arg\min_{\p_c\in\mathcal{M}_c} \tilde{E}_c(\x,\p_c)$; \textcolor{applegreen}{\texttt{ // }Local step\;}
$\bvec=\frac{1}{h^2}\M\y+\sum_c w_c\Gmat_c^\top(\p_c-\Gmat_c\x_{\Cset=\x^*,\Cc=\0})$\;
\textcolor{applegreen}{\texttt{// }Global step\;}
Run Alg.~\ref{alg:linear} to solve $\x_{\Cc}=(\A_{\Cc\times\Cc})^{-1}\bvec_{\Cc}$;
}
$\rf = \frac{1}{h^2}\M(\x-\y)-\fin(\x)$; \textcolor{applegreen}{\texttt{ // }Eqn. (\ref{eq:tic_1})\;}
Update $\Cset$ based on $\rf_j$, $\phi(\x_j)$, and Eqn. (\ref{eq:lcp})\;
}
\label{alg:contact}
\end{algorithm}

\paragraph{Low-rank update} Define a permutation $\sigma$ on $\I$ with the following property: $\sigma$ shuffles $\I$ so that indices from $\Cset$ come before those in $\Cc$ and the internal orders inside $\Cset$ and $\Cc$ are preserved. Define $\Pmat$ as the corresponding permutation matrix: $\Pmat_{ij}=1$ if $\sigma(i)=j$ and $0$ otherwise. Now $\A\Pmat$ shuffles all columns of $\A$ so that the $i$-th column in $\A$ becomes the $\sigma(i)$-th column in $\A\Pmat$. Similarly, $\Pmat^\top\A$ shuffles all rows of $\A$ in the same way. We now rewrite $\Pmat^\top\A\Pmat$ as a $2\times2$ block matrix:
\begin{align}
\Pmat^\top\A\Pmat=\begin{pmatrix}
\A_{\Cset\times\Cset} & \A_{\Cset\times\Cc} \\
\A_{\Cc\times\Cset} & \A_{\Cc\times\Cc} \\
\end{pmatrix}.
\end{align}
Let $c=|\Cset|$ and define $\Umat\in\R^{3n\times2c}$ as follows:
\begin{align}
\Umat=\begin{pmatrix}
\Umat_L & \Umat_R \\
\end{pmatrix}=\begin{pmatrix}
\Imat & \0 \\
\0 & \A_{\Cc\times\Cset} \\
\end{pmatrix},
\end{align}
where $\Umat_L,\Umat_R\in\mathcal{R}^{3n\times c}$ represent the left and right half of $\Umat$ and $\Imat$ the identity matrix of a proper size. Similarly, we define $\Vmat\in\R^{2c\times3n}$ as follows:
\begin{align}
\Vmat=\begin{pmatrix}
\Umat_R^\top \\
\Umat_L^\top \\
\end{pmatrix}=\begin{pmatrix}
\0 & \A_{\Cset\times\Cc} \\
\Imat & \0 \\
\end{pmatrix}.
\end{align}
It is now easy to verify that the product of $\Umat\Vmat$ is the following low-rank matrix:
\begin{align}
\Umat\Vmat=\begin{pmatrix}
\0 & \A_{\Cset\times\Cc} \\
\A_{\Cc\times\Cset} & \0 \\
\end{pmatrix},
\end{align}
and subtracting it from $\Pmat^\top\A\Pmat$ results in a block-diagonal matrix:
\begin{align}
\Pmat^\top\A\Pmat-\Umat\Vmat=\Pmat^\top\underbrace{(\A-\Pmat\Umat\Vmat\Pmat^\top)}_{\A_{\Pmat}}\Pmat=\begin{pmatrix}
    \A_{\Cset\times\Cset} & \0 \\
    \0 & \A_{\Cc\times\Cc} \\
    \end{pmatrix}.
\end{align}
Therefore, we can obtain $(\A_{\Cc\times\Cc})^{-1}$ by inverting $\Pmat^\top\A\Pmat-\Umat\Vmat$. Since inverting $\Pmat$ is trivial ($\Pmat^{-1}=\Pmat^\top$), we focus on computing $\A_{\Pmat}^{-1}$ using the Woodbury matrix identity:
\begin{align}\label{eq:woodbury}
    \A_{\Pmat}^{-1}=\A^{-1}+\A^{-1}\Pmat\Umat(\Imat-\Vmat\Pmat^\top\A^{-1}\Pmat\Umat)^{-1}\Vmat\Pmat^\top\A^{-1}.
\end{align}
Since $\A$ is prefactorized, operations using $\A^{-1}$ in the matrix identity above can be executed efficiently. Moreover, with the assumption that $c\ll n$, $\Imat- \Vmat\Pmat^\top\A^{-1}\Pmat\Umat\in\R^{2c\times2c}$ is a small matrix compared to $\mathbf{A}$, and inverting it (solving a linear system whose left-hand side is this matrix) can be done efficiently. Putting everything together, we transform the problem of factorizing $\A_{\Pmat}$ into factorizing a much smaller linear system of equations.

\paragraph{Time complexity} We now consider a brute-force implementation of Eqn. (\ref{eq:woodbury}) and analyze its time complexity. The time cost is dominated by computing $\A^{-1} \Pmat\Umat$ which takes $\mathcal{O}(n^2c)$ time. While this is still asymptotically smaller than the cost of factorizing the modified matrix, which generally takes $\mathcal{O}(n^3)$ time, the speedup in practice may not be as much as predicted due to the sparsity of $\A_{\Pmat}$. Therefore, further simplification on Eqn. (\ref{eq:woodbury}) would still be desirable.

\paragraph{Further acceleration} To reduce the time cost of computing $\A^{-1}\Pmat\Umat$, we notice that $\Pmat\Umat$ shuffles all rows of $\Umat$ with the inverse mapping $\sigma^{-1}$. As a result, $\Pmat\Umat_L$, the left part of $\Pmat\Umat$, is effectively $\Imat_{\I\times\Cset}$, i.e., a collection of one-hot column vectors
\begin{math}
\e_j,j\in \Cset,
\end{math}
where the $j$-th entry in $\e_j$ is $1$. This means that we can precompute $\A^{-1}\Imat_{\I\times\Cset}$ using a maximum possible $\Cset$ (e.g., all surface nodes) before the whole simulation begins and look up $\A^{-1}\e_j,j\in \Cset$ on the fly.

It turns out that the same idea can also be used for computing $\A^{-1}\Pmat\Umat_R$, the right half of the solution, with a slight modification. Notice that $\Pmat\Umat_R$ can be obtained from $\A$ by fetching $\A_{\I\times\Cset}$ and zeroing out corresponding rows in $\Cset$:
\begin{align}
\Pmat\Umat_R=\A_{\I\times\Cset}-\Imat_{\I\times\Cset}\A_{\Cset\times\Cset}.
\end{align}
We can, therefore, compute $\A^{-1}\Pmat\Umat_R$ as follows:
\begin{equation}
\begin{aligned}
    \A^{-1}\Pmat\Umat_R=&\;\A^{-1}\A_{\I\times\Cset}-\A^{-1}\Imat_{\I\times\Cset}\A_{\Cset\times\Cset}\\
    =&\;\Imat_{\I\times\Cset}-\A^{-1}\Imat_{\I\times \Cset}\A_{\Cset\times\Cset}.
\end{aligned}
\end{equation}
Since $\A^{-1}\Imat_{\I\times\Cset}$ has been precomputed, the time complexity will be bounded by the matrix multiplication $\mathcal{O}(nc^2)$. Moreover, noting that $\A^{-1}$ is symmetric and $\Vmat$ can be obtained from $\Umat$ by swapping and transposing block matrices $\Umat_L$ and $\Umat_R$, the results derived here can also be reused to assemble $\Vmat\Pmat^\top\A^{-1}$.

In conclusion, we have reduced the time complexity of computing $\A^{-1}\Pmat\Umat$ from $\mathcal{O}(n^2c)$ to $\mathcal{O}(nc^2)$. Since the remaining operations, excluding solving $\A^{-1}$ in Eqn. (\ref{eq:woodbury}), are also bounded by $\mathcal{O}(nc^2)$, we now have reduced the \emph{overhead} of applying Eqn. (\ref{eq:woodbury}) from $\mathcal{O}(n^2c)$ to $\mathcal{O}(nc^2)$, with the overhead defined as the extra cost brought by Eqn. (\ref{eq:woodbury}) in addition to one linear solve $\A^{-1}$ with any right-hand side vector. We present the complete algorithm in pseudocode in Alg.~\ref{alg:linear}, which serves as a subroutine in Alg.~\ref{alg:contact}. We use $\B_k$ and $\avec_k$ to denote intermediate matrices and vectors respectively, with the subscript $k$ indicating the order of their first occurrence.

\begin{algorithm}[tb]
\caption{Global step in Alg.~\ref{alg:contact}.}
Input: $\Cset\subseteq\I$, $\x^*\in\R^{3|\Cset|}$, and $\bvec\in\R^{3n}$\;
Output: $\x$ such that $\x_{\Cset}=\x^*$ and $\A_{\Cc\times\Cc}\x_{\Cc}=\bvec_{\Cc}$\;
Collect $\B_1=\A^{-1}\Imat_{\I\times\Cset}$ from precomputed data\;
$\B_2=\Imat_{\I\times\Cset}-\B_1\A_{\Cset\times \Cset}$\;
$\B_3=(\B_1,\B_2)$; \textcolor{applegreen}{\texttt{ // }$\B_3=\A^{-1}\Pmat\Umat$\;}
\textcolor{applegreen}{\texttt{// }$\Vmat\Pmat^\top$ can be fetched from $\A$ without permutation\;}
\textcolor{applegreen}{\texttt{// }No need to compute $\Pmat$\;}
$\B_4=\Imat-\Vmat\Pmat^\top\B_3$\;
Solve $\avec_1$ from $\A\avec_1=\bvec$\;
$\avec_2=(\bvec^\top\B_2,\bvec^\top\B_1)$; \textcolor{applegreen}{\texttt{ // }Row vector\;}
Solve $\avec_3$ from $\B_4\avec_3=\avec_2^\top$\;
$\x=\avec_1+\B_3\avec_3$\;
Set $\x_{\Cset}=\x^*$\;
\label{alg:linear}
\end{algorithm}

\paragraph{Backpropagation}
With a contact set $\Cset$ and the corresponding $\x^*$, the backpropagation scheme in Sec.~\ref{sec:diff_pd} also needs modifications. Backpropagating from $\frac{\partial L}{\partial \x}$ to $\frac{\partial L}{\partial \y}$ now becomes trickier due to the existence of $\Cset$ and $\x^*$ from a collision detection algorithm, which splits both $\x$ and $\y$ into two vectors $\x_{\Cset}$, $\x_{\Cc}$, $\y_{\Cset}$, and $\y_{\Cc}$. Here, we will sketch the core idea by showing how gradients can be backpropagated from $\frac{\partial L}{\partial \x_{\Cc}}$ to $\frac{\partial L}{\y_{\Cc}}$. Backpropagation through other dependencies is easier to derive and therefore skipped.

From Eqns. (\ref{eq:reduced}) and (\ref{eq:gc}), we see that $\x_{\Cc}$ and $\y_{\Cc}$ are constrained by $\nabla_{\x_{\Cc}}g_{\Cset}=\0$. By differentiating Eqn. (\ref{eq:reduced}), we obtain:
\begin{align}
    \nabla^2_{\x_{\Cc}} g_{\Cset}\frac{\partial \x_{\Cc}}{\partial \y_{\Cc}}-\frac{1}{h^2}\M_{\Cc\times\Cc}=\0,
\end{align}
which is a reduced version of Eqn. (\ref{eq:imp_func_theorem}). The chain rule still applies in a similar way:
\begin{align}
    \frac{\partial L}{\partial \y_{\Cc}}=\frac{\partial L}{\partial \x_{\Cc}}\frac{\partial \x_{\Cc}}{\partial \y_{\Cc}}=\frac{1}{h^2}\underbrace{\frac{\partial L}{\partial \x_{\Cc}}[(\nabla^2 g)_{\Cc\times\Cc}]^{-1}}_{\z^\top}\M_{\Cc\times\Cc},
\end{align}
where a new adjoint vector $\z$ is defined. It should now become very clear that $\z$ is obtained from the following linear system of equations:
\begin{align}
    (\nabla^2 g)_{\Cc\times\Cc}\z=(\frac{\partial L}{\partial \x_{\Cc}})^\top.
\end{align}
Now using Eqn. (\ref{eq:matrix_split}), we see the iterative solver in Sec.~\ref{sec:diff_pd} becomes:
\begin{align}
    \A_{\Cc\times\Cc}\z^{k+1}=\Delta\A_{\Cc\times\Cc}\z^{k}+(\frac{\partial L}{\partial \x_{\Cc}})^\top,
\end{align}
from which we see a similar issue we experience in forward simulation: $\A_{\Cc\times\Cc}$ changes dynamically, so the Cholesky factorization of $\A$ is not directly applicable. This is exactly where we can use the same global solver in Alg.~\ref{alg:linear} to retain the source of efficiency in our PD backpropagation algorithm. We summarize this new backpropagation method in Alg.~\ref{alg:contact_backprop}.

\begin{algorithm}[tb]
\caption{PD backpropagation with contact}
Input: $\y$, $\x$ and $\Cset$ (from forward simulation), and $\frac{\partial L}{\partial \x_{\Cc}}$\;
Output: $\frac{\partial L}{\partial \y_{\Cc}}$\;
Initialize $\z=\0$\;
\While{$\z$ not converged}{
$\bvec=(\Delta\A)_{\Cc\times\Cc}\z + (\frac{\partial L}{\partial \x_{\Cc}})^\top$; 
\textcolor{applegreen}{\texttt{ // }Local step\;}
Run Alg.~\ref{alg:linear} to solve $\z=(\A_{\Cc\times\Cc})^{-1}\bvec$; \textcolor{applegreen}{\texttt{ // }Global step\;}
}
$\frac{\partial L}{\partial \y_{\Cc}}=\frac{1}{h^2}\z^\top\M_{\Cc\times\Cc}$;
\textcolor{applegreen}{\texttt{ // }Eqn. (\ref{eq:chain_rule})\;}
\label{alg:contact_backprop}
\end{algorithm}

\paragraph{Summary} In summary, we have presented a differentiable contact handling algorithm that ensures non-penetration conditions and imposes infinitely large static friction. Moreover, we have also discussed its implementation in forward simulation and backpropagation that can still benefit from the Cholesky factorization of $\A$. We stress that there exist more physically accurate contact handling algorithms that satisfy not only non-penetration conditions but also the Coulomb's law of friction~\cite{chen2017dynamics,ly2020projective,Li2020IPC}. However, our contact handling algorithm achieves a good trade-off between differentiability, physically plausibility, and compatibility with our differentiable PD framework.

\section{Evaluation}\label{sec:evaluation}

\begin{figure*}
    \centering
    \includegraphics[trim=70 20 200 260,clip,width=0.16\linewidth]{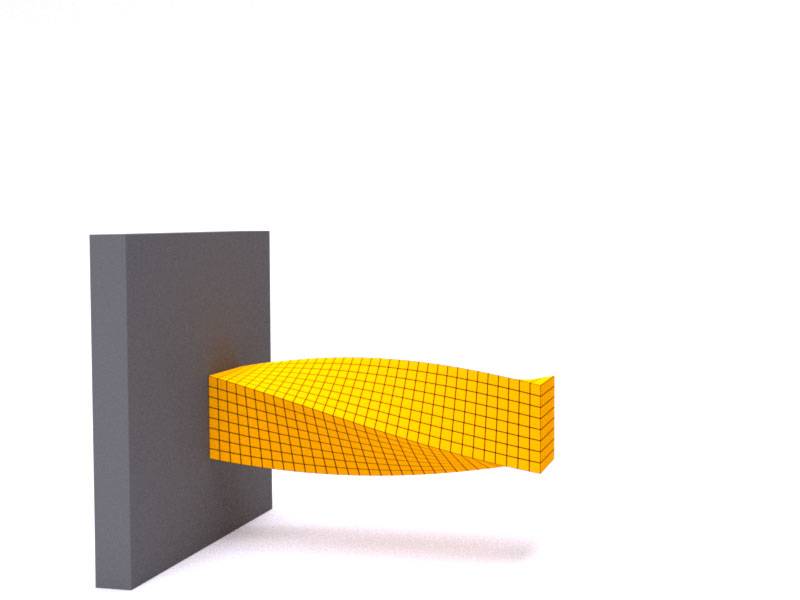}
    \includegraphics[trim=70 20 200 260,clip,width=0.16\linewidth]{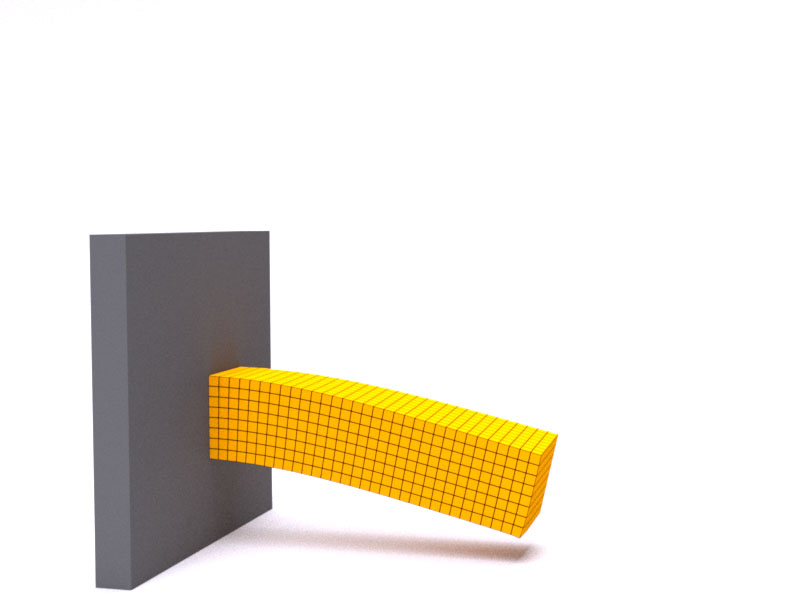}
    \includegraphics[trim=70 20 200 260,clip,width=0.16\linewidth]{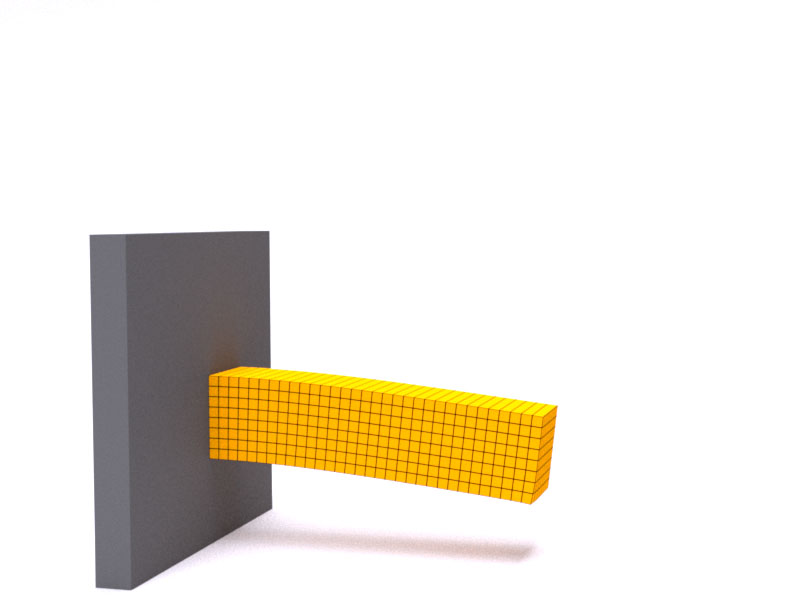}
    \quad
    \includegraphics[trim=200 45 380 400,clip,width=0.16\linewidth]{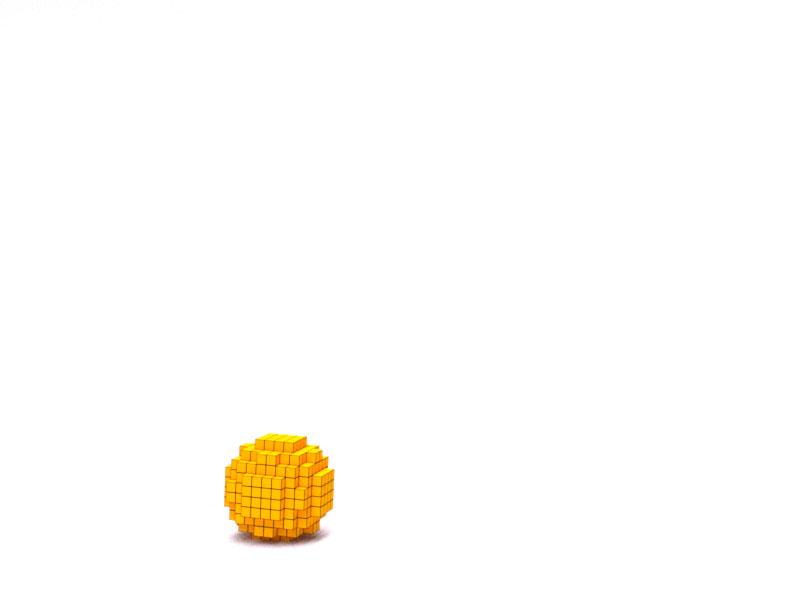}
    \includegraphics[trim=240 45 340 400,clip,width=0.16\linewidth]{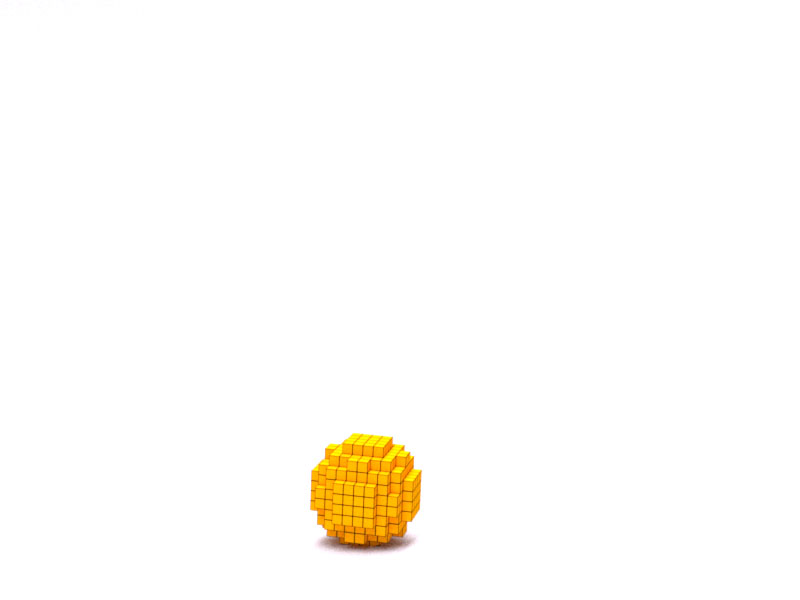}
    \includegraphics[trim=280 45 300 400,clip,width=0.16\linewidth]{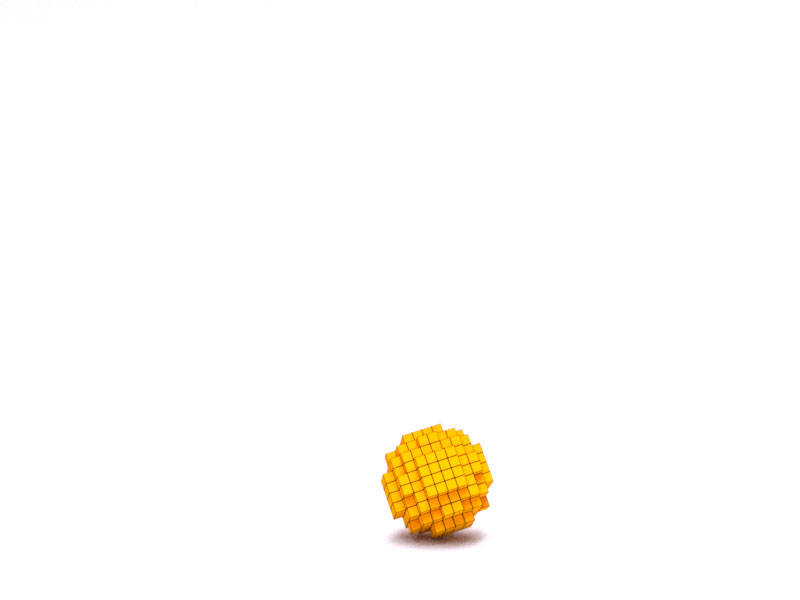}
    \vspace{-2em}
  \caption{The ``Cantilever'' and ``Rolling sphere'' examples in Sec.~\ref{sec:evaluation} designed for comparing DiffPD to the Newton's method. The ``Cantilever'' example starts with a twisted cantilever (left), oscillates, and bends downwards eventually due to gravity. In the ``Rolling sphere'' example, we roll a soft sphere on the ground (right) which constantly breaks and reestablishes contact.}
  \label{fig:benchmark_vis}
\end{figure*}

In this section, we compare DiffPD with a few baseline differentiable simulation methods and conduct ablation studies on the acceleration techniques in Sec.~\ref{sec:diff_pd} and Sec.~\ref{sec:contact}. We start by discussing the difference between implicit and explicit time-stepping schemes in backpropagation. Next, we compare our simulator with two other fully implicit simulators implemented with the Newton's method. We end this section with a discussion on the two contact models implemented in DiffPD. The end goal of this section is to evaluate the difference between different time-stepping methods and understand the source of efficiency in DiffPD. We implement both baseline algorithms and DiffPD in C++ and use Eigen~\cite{eigen} for sparse matrix factorization and linear solvers. We run all experiments in this section and next section on a virtual machine instance from Google Cloud Platform with 16 Intel Xeon Scalable Processors (Cascade Lake) @ 3.1 GHz and 64 GB memory. We use OpenMP for parallel computing and 8 threads by default unless otherwise specified.

\subsection{Comparisons with Explicit Method}
Compared with explicit time-stepping methods used in previous papers on differentiable simulation~\cite{hu2019chainqueen,hu2019difftaichi,spielberg2019learning}, implicit time integration brings two important changes to a differentiable simulator: first, implicit methods enable a much larger time step during simulation, resulting in much fewer number of frames. This is particularly beneficial for solving an inverse problem with a long time horizon as we store fewer states (nodal positions and velocities) in memory during backpropagation. Second, due to implicit damping, we can expect the landscape of the loss function defined on nodal states and their derived quantities to be smoother.

To demonstrate the memory consumption, we consider a soft cantilever discretized into $12\times3\times3$ hexahedral elements (a low-resolution version of the ``Cantilever'' example in Fig.~\ref{fig:benchmark_vis} left). We impose Dirichlet boundary conditions on one end of the cantilever and simulate its vibration after twisting the other end of the cantilever under gravity for 0.2 seconds. We define a loss function $L$ as a randomly generated weighted average of the final nodal positions and velocities. The implicit time integration in our simulator allows us to use a time step as large as 10 milliseconds, while a explicit implementation is only numerically stable in both forward simulation and backpropagation for time steps of 0.5 milliseconds. Since memory consumption during backpropagation is proportional to the number of frames, we can expect a 20$\times$ increase in memory consumption for the explicit method, requiring additional techniques like checkpoint states~\cite{hu2019chainqueen,spielberg2019learning} before the problem size can be scaled up.

\begin{figure}[ht]
  \centering
  \includegraphics[width=0.8\linewidth]{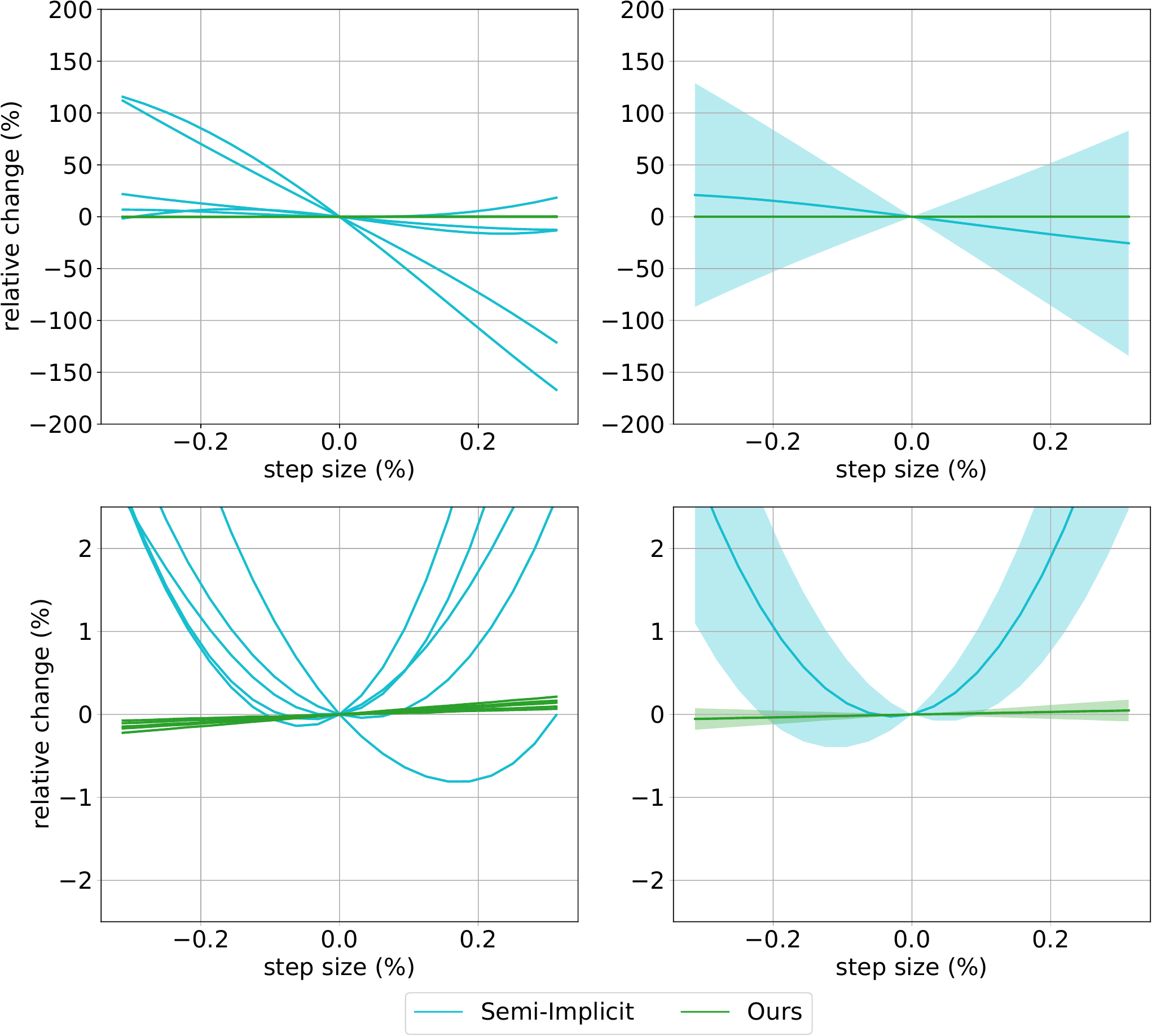}
    \vspace{-1.2em}
  \caption{The relative changes in both the loss (top left) and the magnitude of the gradient (bottom left) for the explicit method (cyan) and our method (green) for 5 out of the 16 random directions in the neighborhood of the initial nodal positions $\mathbf{x}_0$. Also shown are the means (solid curves) and standard deviations (shaded) of the percent change in loss (top right) and the magnitude of the gradient (bottom right) for all 16 random directions.}
  \label{fig:l_g_si}
  \Description{Comparison in backpropagation robustness}
\end{figure}

To demonstrate the influences of time-stepping schemes on the smoothness of the energy landscape, we visualize in Fig.~\ref{fig:l_g_si} the loss function $L$ and its gradient norm $|\nabla L|$ sliced along 16 random directions in the neighborhood of the cantilever's initial nodal positions. Specifically, let $\mathbf{x}_0$ be the initial nodal positions, and let $\mathbf{r}_1,\mathbf{r}_2,\dots,\mathbf{r}_{16}$ be the random directions. We plot $L(\mathbf{x}_0+\alpha\mathbf{r}_i)$ and $|\nabla L(\mathbf{x}_0+\alpha\mathbf{r}_i)|$ for each $\mathbf{r}_i$ (Fig.~\ref{fig:l_g_si} left) with $\alpha$ being the step size, which is uniformly sampled between $-0.3\%$ and $0.3\%$ of the cantilever beam length. The standard deviations from 16 random directions (Fig.~\ref{fig:l_g_si} right) indicate that small perturbations in $\mathbf{x}_0$ lead to much smoother loss and gradients when implicit time integration is applied, which is not surprising due to numerical damping. From the perspective of differentiable simulation, a smoother energy landscape can be more favorable as it induces more well-defined gradients to be used by gradient-based optimization techniques.

\subsection{Comparisons with Other Implicit Methods}\label{sec:evaluation:implicit}

We now compare our simulator with other implicit time-stepping schemes to evaluate its speedup in both forward and backward modes. We choose Newton's method with two standard sparse linear solvers: an iterative solver using preconditioned conjugate gradient (Newton-PCG) and a direct solver using Cholesky decomposition (Newton-Cholesky), as our baseline solvers for implicit time integration in Eqn. (\ref{eq:time_integration}). First, we compare with the Newton's method in Sec.~\ref{sec:diff_pd} without contact. Then, we benchmark the performance of the contact handling algorithm in Sec.~\ref{sec:contact}. We reiterate that just like in the standard PD framework, any resultant speedup from DiffPD over Newton's method is under the assumption that the material model has a quadratic energy function. We extend our discussion to general hyperelastic materials at the end of paper and leave it as future work.

\paragraph{Simulation without Contact}
We benchmark our method, Newton-PCG, and Newton-Cholesky using a cantilever with $32\times8\times8$ elements, 8019 DoFs, and 243 Dirichlet boundary constraints (``Cantilever'' in Fig.~\ref{fig:benchmark_vis} and Table~\ref{tab:setting}). The example runs for 25 frames with time steps of 10 milliseconds. We define the loss $L$ as a randomly generated weighted sum of the final nodal positions and velocities.

\begin{figure}[ht]
  \centering
  \includegraphics[width=\linewidth]{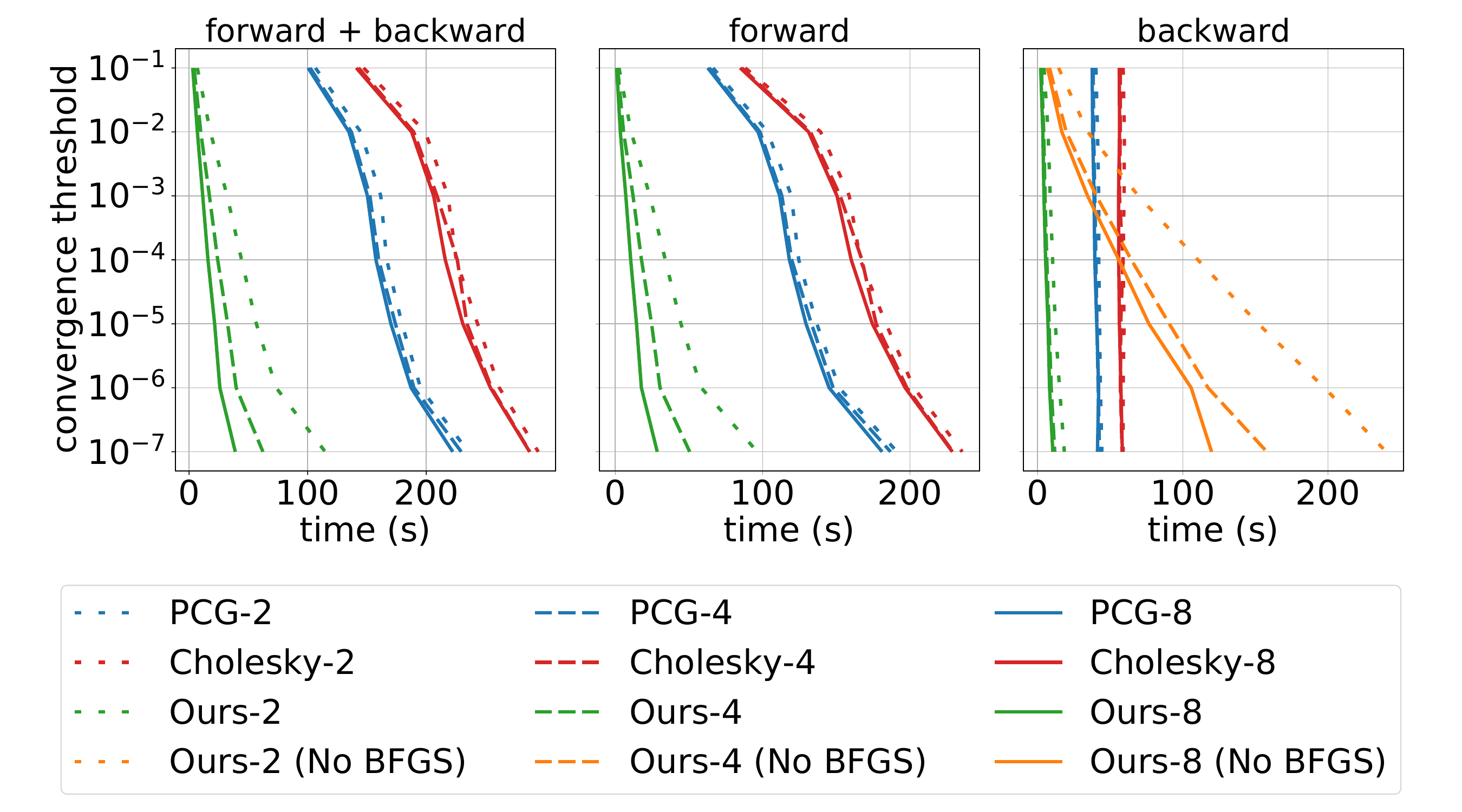}
  \includegraphics[width=0.9\linewidth]{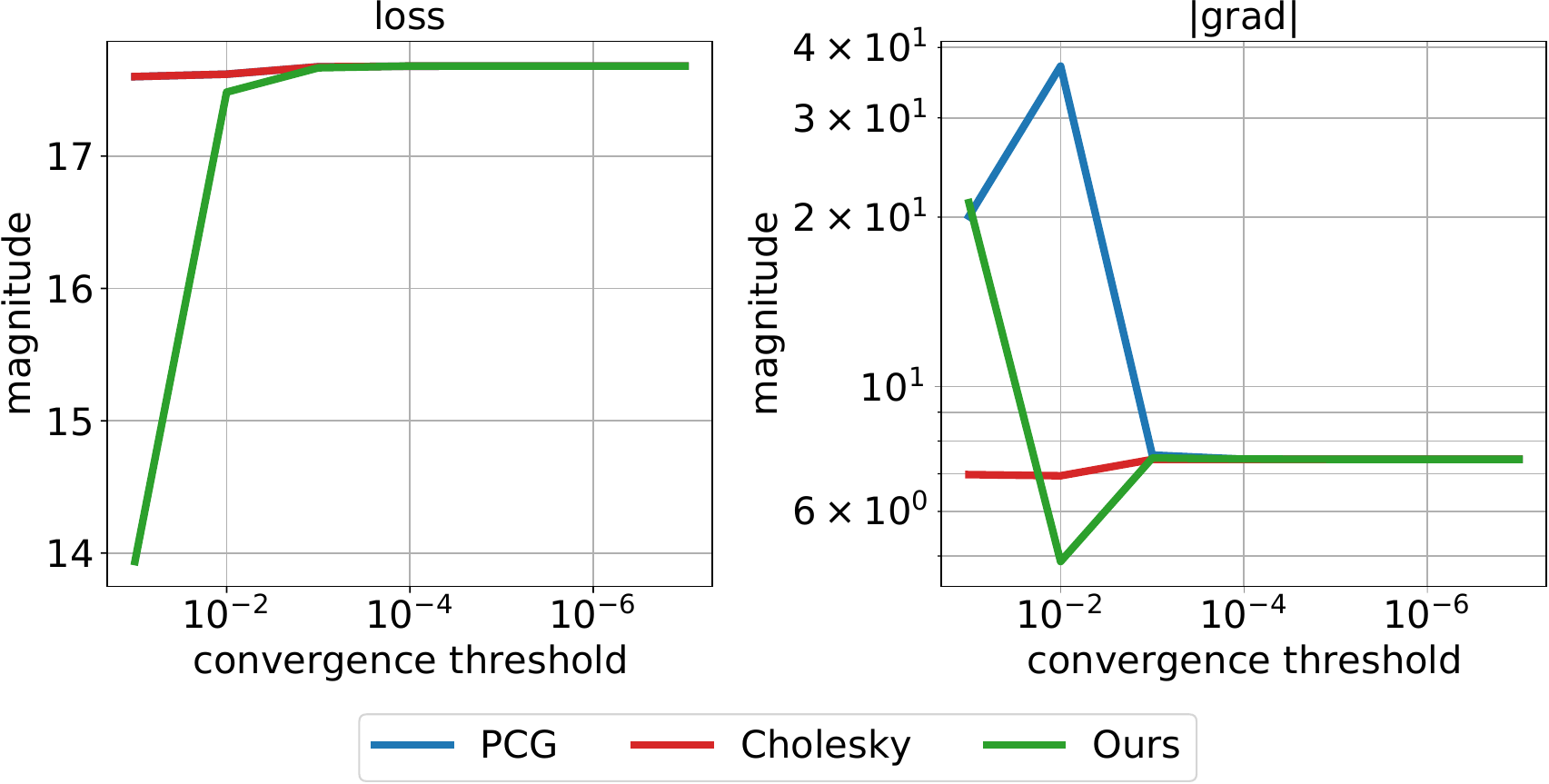}
  \caption{Top: the net wall-clock times (left), forward times (middle), and backpropagation times (right) for different convergence thresholds tested on the ``Cantilever'' example in Sec.~\ref{sec:evaluation:implicit}. The results are obtained from simulating this example using each of the three methods: Newton-PCG (PCG), Newton-Cholesky (Cholesky), and DiffPD (Ours). The number following the method name denotes the number of threads used. Also shown are the results from running our method without applying the quasi-Newton method (orange, right). Bottom: the loss (left) and magnitude of the gradient of the loss (right) for different convergence thresholds used to terminate iterations in Newton's method and our simulator.}
  \label{fig:benchmark}
  \Description{Time cost of each methods with multiple threads}
\end{figure}

In terms of the running-time comparison, we report results in Fig.~\ref{fig:benchmark} from running all three methods with 2, 4, and 8 threads and a range of convergence threshold (from 1e-1 to 1e-7) on the relative error in solving Eqn. (\ref{eq:time_integration}). The speedup from parallel computing is less evident in the Newton's method because the majority of their computation time is spent on matrix refactorization -- a process that cannot be trivially parallelized in Eigen. We conclude that our simulator has a clear advantage over Newton's method on the time cost of both forward simulation and backpropagation. For forward simulation, the speedup is well understood and discussed in many previous PD papers~\cite{bouaziz2014projective,liu2017quasi}. For moderate tolerances (1e-3 to 1e-5), we observe a speedup of 9-16 times in forward simulation with 8 threads and note that it becomes less significant as precision increases. Both of these observations agree with previous work on PD for forward simulation. In backpropagation, DiffPD method is 6-13 times faster than Newton's method for moderate tolerances due to the reuse of the Cholesky decomposition and the quasi-Newton update. Specifically, we point out that without the proposed acceleration technique with quasi-Newton methods in Sec.~\ref{sec:diff_pd}, PD backpropagation is faster than Newton's method only for very low precision (orange in Fig.~\ref{fig:benchmark} right), confirming the necessity of the quasi-Newton updates.

Since PD is an iterative method whose result is dependent on the convergence threshold, it is necessary to justify which threshold is the most proper. To analyze the influence of the choice of thresholds, we use results from Newton-Cholesky as the oracle because it is a direct solver whose solution is computed with the machine precision in Eigen. We then compare both our method and Newton-PCG with the oracle by computing the loss and gradients of the ``Cantilever'' example with varying convergence thresholds and analyze when the results from the three methods start to coincide. This comparison provides quantitative guidance on the choice of convergence threshold and reveal the range in which our method can be a reliable alternative to the Newton's method in optimization tasks. We report our findings in Fig.~\ref{fig:benchmark}. As Newton-PCG and our method are iterative methods, their accuracy improves when the convergence threshold becomes tighter. It can be seen from the figure that our method agrees with the Newton's method on the numerical losses and gradients when using a threshold as large as 1e-4. Therefore, we use 1e-4 as our default threshold in all applications to be discussed below unless otherwise specified. Referring back to Fig.~\ref{fig:benchmark}, using 8 threads and with a convergence threshold of 1e-4, our method achieves significant speedup (12-16 times faster in forward simulation and 6.5-9 times faster in backpropagation) compared with Newton-PCG and Newton-Cholesky.

\begin{figure}[b]
  \centering
  \includegraphics[width=\linewidth]{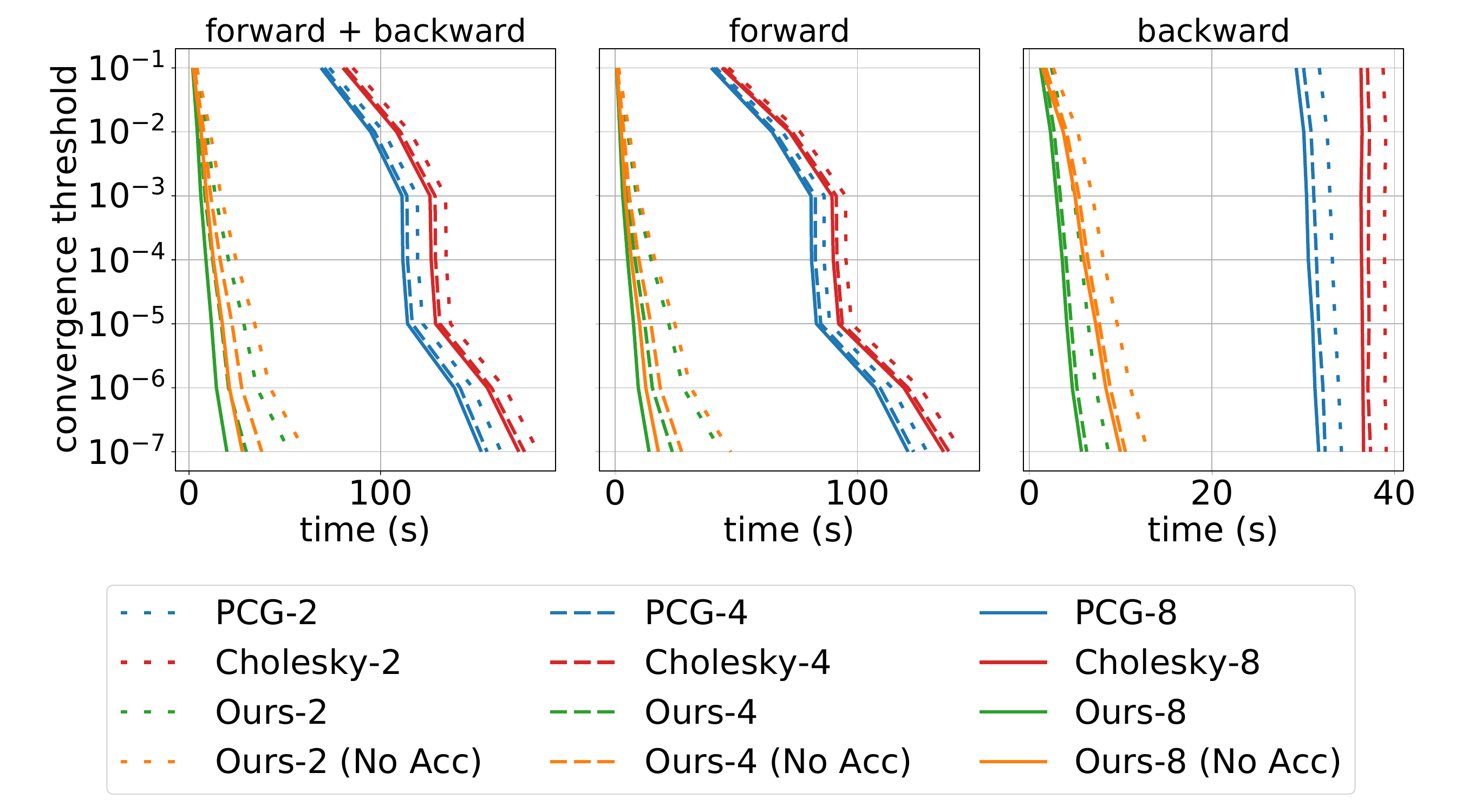}
  \includegraphics[width=0.9\linewidth]{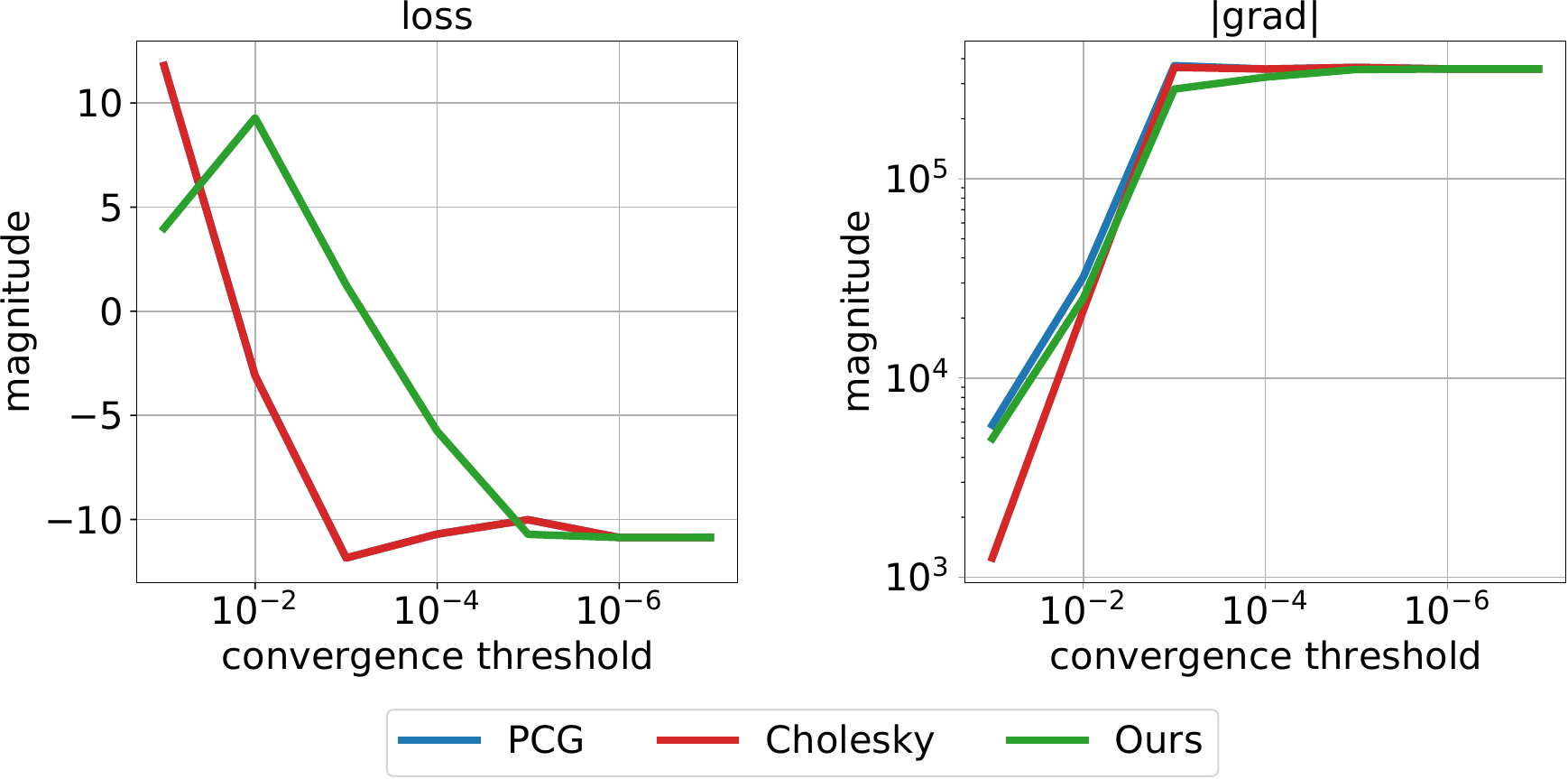}
  \caption{Top: the net wall-clock times (left), forward times (middle), and backpropagation times (right) for different convergence thresholds tested on the ``Rolling sphere'' example for contact handling. The results are obtained from three methods: Newton-PCG (PCG), Newton-Cholesky (Cholesky), and DiffPD (Ours). The number following the method name denotes the number of threads used. Also shown are the results from running our method without applying the further acceleration technique (Alg.~\ref{alg:linear}) in Sec.~\ref{sec:contact:compl} (orange). Bottom: the loss (left) and magnitude of the gradient of the loss (right) for different convergence thresholds used to terminate iterations in the Newton's method and our simulator.}
  \label{fig:rolling_jelly}
  \Description{Time cost of each methods with multiple threads for contact handling.}
\end{figure}

\paragraph{Simulation with Contact}
To create a benchmark scene that requires contact handling constantly, we roll a soft sphere on a horizontal collision plane for 100 frames with a time step of 5 milliseconds (``Rolling sphere'' in Fig.~\ref{fig:benchmark_vis} and Table~\ref{tab:setting}). The sphere is voxelized into 552 elements with 2469 DoFs, and the maximum possible contact set $\Cset$ we consider consists of 72 nodes (216 DoFs) on the surface of the sphere. Similar to the ``Cantilever'' example, we define the loss function $L$ as a randomly generated weighted average of the final nodal positions and velocities. We implement the contact handling algorithm in Sec.~\ref{sec:contact:compl} with Newton-PCG, Newton-Cholesky, and our method, and we report their time cost as well as their loss and gradients in Fig.~\ref{fig:rolling_jelly}. It can be seen from Fig.~\ref{fig:rolling_jelly} that the results from three methods start to converge when the convergence threshold reaches 1e-6, with which our method is 10 times faster than the Newton's method in both forward and backward mode (Fig.~\ref{fig:rolling_jelly}). Such speedup mainly comes from the low-rank update algorithm (Alg.~\ref{alg:linear}) which avoids the expensive matrix factorization from scratch. Additionally, by comparing the orange and green curves in Fig.~\ref{fig:rolling_jelly}, we conclude that the acceleration technique of caching $\A^{-1}\Imat_{\I\times\Cset}$ further speeds up DiffPD by $25\%$ in forward mode and $44\%$ in backward mode when measured with 8 threads and a convergence threshold of 1e-6.

\subsection{Ablation Study}
We end this section with an ablation study on multiple components in our algorithm. We start with an empirical analysis on the iterative solver and the line search algorithm in our backpropagation algorithm (Sec.~\ref{sec:diff_pd}), followed by an evaluation on the penalty-based and the complementarity-based contact models.

\paragraph{Spectral radius and line search} One key assumption we have made in our backpropagation solver is that the spectral radius of $\rho(\mathbf{A}^{-1}\Delta\mathbf{A})<1$, which is also one of the primary reasons why we have employed the line search algorithm as a safeguard when the assumption does not hold. Here, we use the ``Cantilever'' example check if this assumption holds empirically. We explicitly calculate $\rho(\A^{-1}\Delta\A)$ we experience in ``Cantilever'' and observe a maximum value of $0.996$, indicating that we can expect convergence in the iterative solver, which we further confirm by testing the iterative solver with 100 randomly generated, artificial right-hand side vector $\frac{\partial L}{\partial\mathbf{x}}$. We observe similar results about the convergence of the backpropagation solver in the ``Rolling sphere'' example as well as in our applications to be described in Sec.~\ref{sec:application}, indicating that it seems safe to expect the iterative solver to converge in practice despite the lack of a theoretical guarantee on it.

As employing line searches in our algorithm serves as a safeguard to cases when $\rho(\mathbf{A}^{-1}\Delta\mathbf{A})>1$, an implication from the observations on the spectral radius is that we rarely trigger line searches to reduce the step size in practice. In fact, in this ``Cantilever'' example, and in almost all applications below, we notice that the default step size ($1$ in Newton's and quasi-Newton methods) almost always allows us to skip the line search stage. Still, we precautionarily set the maximum number of line search iterations to be 10 for all examples.

\paragraph{Penalty-based contact} We implement the penalty-based contact and frictional forces from~\citet{Macklin2020PrimalDual} in DiffPD and analyze them in both forward simulation and backpropagation. First, we use a standard ``Slope'' test with varying frictional coefficients in the penalty-based model to understand the expressiveness of this contact model in forward simulation. Second, we use a ``Duck'' example which optimizes frictional coefficients using the gradients of this contact model in backpropagation.

\begin{figure}[htb]
    \centering
    \includegraphics[width=\columnwidth]{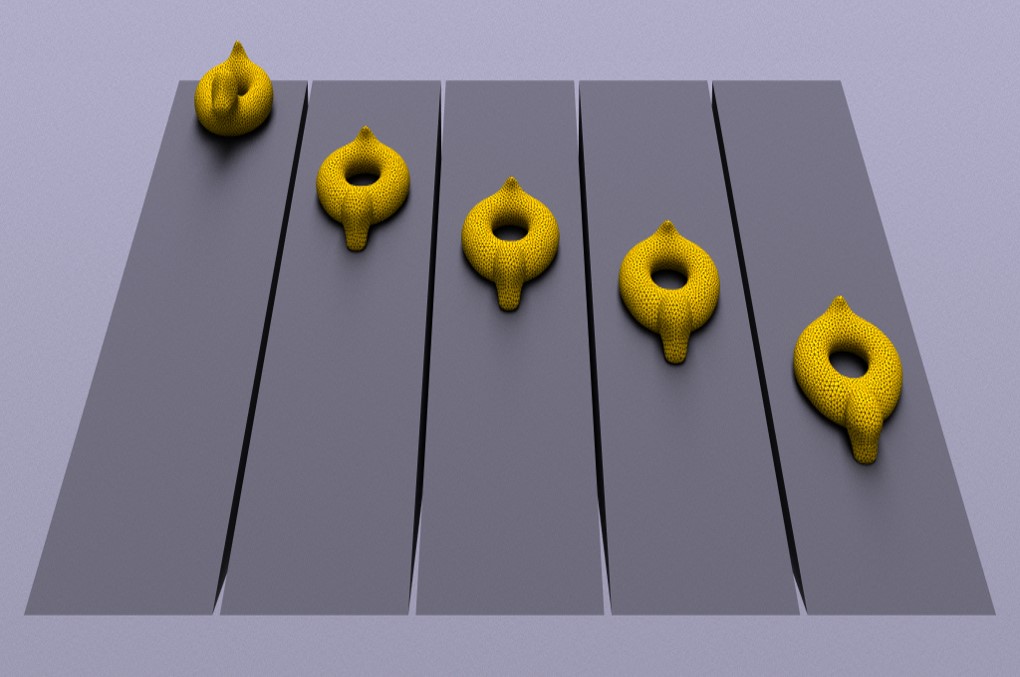}
    \caption{\textbf{Slope}. A rubber duck (16776 DoFs and 24875 tetrahedrons) slides on slopes with varying frictional coefficients implemented in DiffPD using a penalty-based contact and friction model~\cite{Macklin2020PrimalDual}. Left: the initial position of the duck. Middle left to right: the final positions of the duck after two seconds with a decreasing frictional coefficient.}
    \label{fig:duck_slope}
\end{figure}

To show the capacity of the penalty-based contact model in forward simulation, we consider the ``Slope'' test visualized in Fig.~\ref{fig:duck_slope}. We place a squishy rubber duck (16776 DoFs and 24875 tetrahedrons) on four slopes with varying frictional coefficients from the penalty form in~\citet{Macklin2020PrimalDual} and let it slide for two seconds under gravity. We can see from the figure that with decreasing sliding friction from the left slope to the right slope, the implementation of~\citet{Macklin2020PrimalDual} in DiffPD generates different sliding distances that match our expectation qualitatively.

\begin{figure}[htb]
    \centering
    \includegraphics[width=\columnwidth]{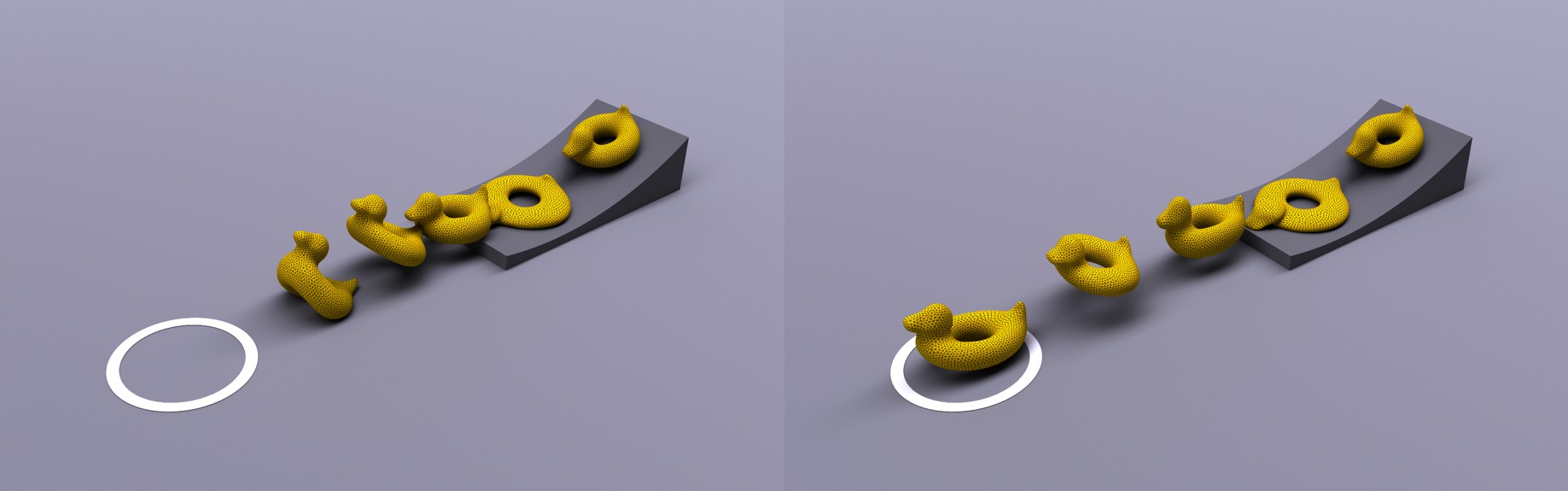}
    \caption{\textbf{Duck}. The same rubber duck in Fig.~\ref{fig:duck_slope} now slides on a curved surface with trainable frictional coefficients. The goal in this test is to optimize the frictional coefficients so that the duck's final position after one second of simulation reaches the center of the white circle as closely as possible. We overlay the intermediate positions of the rubber duck at 0s, 0.25s, 0.5s, 0.75s, and 1s in simulation with an initial guess of the frictional coefficients before optimization (left) and the final coefficients after gradient-based optimization (right).}
    \label{fig:duck_slide}
\end{figure}

Backpropagating a penalty-based contact model is straightforward because it only requires a procedural application of chain rules to differentiate the penalty energy. To show the penalty-based model is fully compatible with DiffPD's backpropagation and can be useful in optimization problems, we design a ``Duck'' example (Fig.~\ref{fig:duck_slide}) with the same rubber duck but on a curved slide with frictional coefficients to be optimized (3 DoFs in total). The duck slides off the curved surface and aims to land on a target location (indicated by the white circle). The frictional coefficients affect the stickiness of the curve surface and control the exiting velocity of the duck when it leaves the slide, which further determines its movement under gravity afterwards. From the two motion sequences in Fig.~\ref{fig:duck_slide} before and after gradient-based optimization, we observe a substantial improvement that eventually leads the duck to the target position. This confirms the usefulness of gradients computed in DiffPD using the penalty-based contact method.

\paragraph{Complementarity-based contact} For the contact model described in the complementarity form, our backpropagation algorithm assumes the contact set is a small subset of full DoFs. Specifically, Alg.~\ref{alg:linear} requires a relatively small size of $\Cset$ at each time step to gain substantial speedup over directly solving the modified linear system without leveraging the low-rank update. Given that $\Cset$ is a subset of surface vertices, whose number is much fewer than the number of interior vertices in a typical 3D volumetric deformable body, such an assumption can be easily satisfied in many applications. Indeed, in the next section we will present various 3D examples involving contact, none of which have more than $6\%$ active contact nodes throughout simulation.

\begin{figure}
    \centering
    \includegraphics[width=\columnwidth]{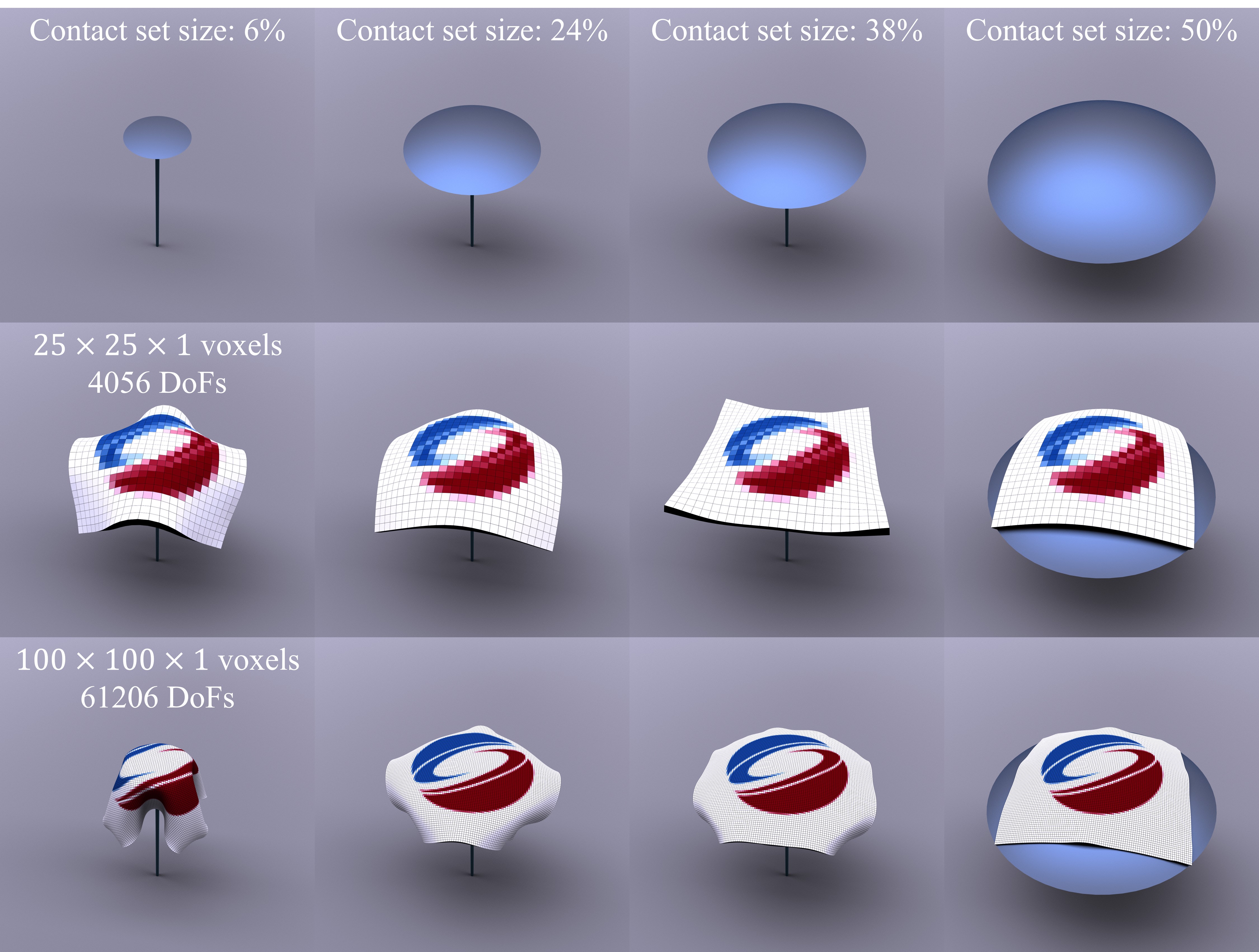}
    \caption{Ablation study on the relative size of the active contact set $|\Cset|$ and its influence on DiffPD's speed. We simulate a one-layer ``Napkin'' with resolutions from $25\times25\times1$ voxels (middle row) to $100\times100\times1$ voxels (bottom row) falling onto a spherical obstacle (blue, top row) with a varying solid angle. We report the relative size of $|\Cset|$ and the degrees of freedom of the napkin. Please refer to Table~\ref{tab:napkin} for the detailed running time and our video for the full motion of the falling napkin.}
    \label{fig:napkin}
\end{figure}

The assumption that $|\Cset|$ is relatively small is much more likely to be violated when we simulate a co-dimensional object, e.g., a one-dimensional rope or a piece of cloth in 3D, in which case it is entirely possible to have all nodes in $\Cset$ at some point. Although simulating co-dimensional objects is beyond the scope of this work, it can be a good test to reveal a critical ratio where the speedup from Alg.~\ref{alg:linear} starts to diminish. To mimic a co-dimensional object, we engineer a ``Napkin'' example consisting of one-layer voxels (Fig.~\ref{fig:napkin}) falling onto a spherical obstacle with an adjustable solid angle to control the size of $|\Cset|$. The relative size of $|\Cset|$ is capped by $50\%$ when all the bottom nodes are in contact with the spherical obstacle (Fig.~\ref{fig:napkin} right column). We vary the mesh resolution from $25\times25\times1$ voxels (4056 DoFs) to $100\times100\times1$ voxels (61206 DoFs) and report the running time of Newton's method and DiffPD in Table~\ref{tab:napkin} for each resolution and contact set size. We can use Table~\ref{tab:napkin} to decide between using our low-rank update method and directly solving the modified matrix in a downstream application. For example, for around 15k DoFs, Table~\ref{tab:napkin} suggests that the low-rank update method is faster until the relative size of $\Cset$ reaches around $40\%$.

\begin{table}[tb!]
\caption{Average running time per step in the napkin example with various mesh resolution (``Res.'') and relative contact set size from 6\% to 50\% (Fig.~\ref{fig:napkin} top row). The reported time is averaged over all steps when the napkin is in contact with the obstacle. All times are in seconds. For each mesh resolution and each relative contact set, we report the running time from both Newton's method and DiffPD with the shorter time in bold.}
\label{tab:napkin}
\begin{tabular}{c|c|cccc}
\toprule
Res. & Method & 6\% & 24\% & 38\% & 50\% \\
\midrule
$25\times25\times1$ & Newton-PCG & 0.8 & 1.6 & 1.6 & 1.5 \\
(4056 DoFs) & DiffPD & \textbf{0.1} & \textbf{0.6} & \textbf{1.2} & \textbf{1.4} \\
\midrule
$50\times50\times1$ & Newton-PCG & 5.4 & 8.4 & \textbf{10.2} & \textbf{8.5} \\
(15606 DoFs) & DiffPD & \textbf{1.2} & \textbf{6.0} & 11.3 & 10.5 \\
\midrule
$75\times75\times1$ & Newton-PCG & 14.7 & \textbf{25.1} & \textbf{25.2} & \textbf{22.5} \\
(34656 DoFs) & DiffPD & \textbf{7.1} & 29.0 & 42.8 & 48.1 \\
\midrule
$100\times100\times1$ & Newton-PCG & 44.6 & \textbf{65.0} & \textbf{47.3} & \textbf{48.5} \\
(61206 DoFs) & DiffPD & \textbf{32.0} & 169.8 & 158.7 & 163.4 \\
\bottomrule
\end{tabular}
\end{table}

\section{Applications} \label{sec:application}

In this section, we show various tasks that can benefit from DiffPD and classify them into five categories: system identification, inverse design, trajectory optimization, closed-loop control, and real-to-sim applications. Although prior efforts on differentiable simulators have demonstrated their capabilities on almost all these examples, we highlight that DiffPD is able to achieve comparable results but reduce the time cost by almost an order of magnitude. We provide a summary of each example in Table~\ref{tab:setting}. For examples with actuators, we implement the contractile fiber model as discussed in~\citet{min2019softcon}. Regarding the optimization algorithm, we use L-BFGS in our examples by default unless otherwise specified. We report the time cost and the final loss after optimization in Table~\ref{tab:performance}. For fair comparisons, we use the same initial guess and termination conditions when running L-BFGS with different simulation methods.  When reporting the loss in Table~\ref{tab:performance}, we linearly normalize it so that a loss of 1 represents the average performance from 16 randomly sampled solutions and a loss of 0 maps to a desired solution. For examples using a bounded loss, we map zero loss to an oracle solution that achieves the lower bound of the loss (typically 0). For unbounded losses used in the walking and swimming robots (Sec.~\ref{sec:app:plan} and~\ref{sec:app:control}), we map zero loss to the performance of solutions obtained from DiffPD. The full details about each experiment can be found in our supplemental material and our source code.

\begin{table*}[ht!]
\caption{The setup of all examples in the paper. The right five columns report whether the example has gravity as an external force, imposes Dirichlet boundary conditions on nodal positions, handles contact, requires hydrodynamical forces, and has actuators.}
\label{tab:setting}
\begin{tabular}{c|l|rrrr|ccccc}
\toprule
Sec. & Task name & \# of elements & \# of DoFs & $h$ (ms) & \# of steps & Gravity & Dirichlet & Contact & Hydrodynamics & Actuation \\ \midrule
\multirow{2}{*}{\ref{sec:evaluation:implicit}}       & Cantilever         & 2048  & 8019  & 10    & 25    & $\checkmark$  & $\checkmark$  &               &             & $\checkmark$ \\
                                            & Rolling sphere    & 552   & 2469  & 5     & 100   & $\checkmark$  &               & $\checkmark$  &             & \\
\midrule
\multirow{2}{*}{\ref{sec:app:parameter}}    & Plant        & 3863   & 29763  & 10    & 200   &  & $\checkmark$  &               &             & \\
                                            & Bouncing ball     & 1288   & 9132  & 4     & 125    & $\checkmark$  &               & $\checkmark$  &             & \\
\midrule
\multirow{2}{*}{\ref{sec:app:initial}}      & Bunny             & 1601  & 7062  & 1     & 100   & $\checkmark$  &               & $\checkmark$  &             & \\
                                            & Routing tendon    & 512   & 2475  & 10    & 100   & $\checkmark$  & $\checkmark$  &               &             & $\checkmark$ \\
\midrule
\multirow{3}{*}{\ref{sec:app:plan}}         & Torus             & 568   & 3204  & 4     & 400   & $\checkmark$  &               & $\checkmark$  &             & $\checkmark$ \\
                                            & Quadruped         & 648   & 3180  & 10    & 100   & $\checkmark$  &               & $\checkmark$  &             & $\checkmark$ \\
                                            & Cow               & 475   & 2488  & 1     & 600   & $\checkmark$  &               & $\checkmark$  &             & $\checkmark$ \\
\midrule
\multirow{2}{*}{\ref{sec:app:control}}      %
                                            & Starfish     & 1492   & 7026  & 33.3  & 200   &               &               &               & $\checkmark$ & $\checkmark$ \\ 
                                            & Shark             & 2256  & 9921  & 33.3  & 200   &               &               &               & $\checkmark$ & $\checkmark$ \\
                                            \midrule
\multirow{1}{*}{\ref{sec:app:sim2real}}      & Tennis balls     & 640  & 978  & 5.6  & 150   & $\checkmark$              &               &  $\checkmark$             &  &  \\ 
                                            \bottomrule
\end{tabular}
\end{table*}

\begin{table*}[ht!]
\caption{The performance of DiffPD on all examples. For each method and each example, we report the time cost of evaluating the loss function once (Fwd.) and its gradients once (Back.) in the unit of seconds. Also shown are the number of evaluations of the loss and its gradients (Eval.) in BFGS optimization. The ``Loss'' column reported the normalized final loss after optimization (lower is better) with the best one in bold. Finally, we report the speedup computed as the ratio between the forward plus backward time of the Newton's method and of DiffPD, i.e., ratio between the sum of ``Fwd.'' and ``Back.'' columns. Note that no speedup is gained from DiffPD in the real-to-sim example ``Tennis balls'' because of its too small number of DoFs (Table~\ref{tab:setting}).}
\label{tab:performance}
\begin{tabular}{l|l|rrrr|rrrr|rrrrr}
\toprule
\multirow{2}{*}{Sec.} & \multirow{2}{*}{Task name} & \multicolumn{4}{c|}{Newton-PCG}   & \multicolumn{4}{c|}{Newton-Cholesky} & \multicolumn{5}{c}{DiffPD (Ours)} \\
                      &                            & Fwd. & Back. & Eval. & Loss & Fwd. & Back. & Eval. & Loss & Fwd. & Back. & Eval. & Loss & Speedup \\
\midrule
\multirow{2}{*}{\ref{sec:evaluation:implicit}}& Cantilever & 118.2 & 39.4 & - & - & 160.1 & 55.9 & - & - & 10.5 & 5.5 & - & - & $10\times$ \\
& Rolling sphere & 107.3 & 31.3 & - & - & 135.6 & 36.6 & - & - & 14.0 & 5.7 & - & - & $8\times$ \\
\midrule
\multirow{2}{*}{\ref{sec:app:parameter}}& Plant        & 1089.5    & 530.5    & 10    & 1.9e-3 & 929.6 & 525.2 & 10 & 1.9e-3  & 71.6 & 94.7 & 28 & \textbf{5.9e-7} & $9\times$\\
                                        & Bouncing ball     & 269.3    & 90.9     & 43    & \textbf{7.9e-2} & 262.6 & 102.5 & 22 & 8.4e-2 & 15.8 & 14.2 & 12 & 9.6e-2 & $12\times$ \\
\midrule
\multirow{2}{*}{\ref{sec:app:initial}}  & Bunny             & 277.7   & 88.0    & 21    & 7.0e-3 & 358.2 & 126.9 & 29 & \textbf{5.1e-3} & 24.0 & 17.3 & 11 & 2.3e-2 & $9\times$ \\
                                        & Routing tendon    & 108.2   & 56.7    & 36    & 6.0e-4 & 107.3 & 58.7  & 38 & \textbf{4.9e-4} & 8.3  & 9.9  & 30 & 9.6e-4 & $9\times$ \\
\midrule
\multirow{3}{*}{\ref{sec:app:plan}}     & Torus             & 751.9 & 210.3 & 47 & -2.3e-3 & 719.9 & 212.4 & 43 & \textbf{-2.4e-2} & 84.3 & 81.9 & 27 & 0 & $6\times$ \\
                                        & Quadruped         & 289.2 & 51.5  & 69 & \textbf{-1.8e0} & 246.3 & 47.8 & 54 & -1.1e0 & 50.2 & 15.8 & 30 & 0 & $4\times$ \\
                                        & Cow               & 771.7 & 141.7 & 14 & 9.7e-1 & 620.1 & 140.2 & 20 & 9.8e-1 & 105.3 & 43.7 & 31 & \textbf{0} & $5\times$ \\
\midrule
\multirow{2}{*}{\ref{sec:app:control}}  & Starfish          & 217.7   & 105.1   & 100   & 4.8e-1 & 244.0 & 129.4 & 100 & 1.4e-1 & 5.7 & 10.8 &100 & \textbf{0} & $19\times$ \\
                                        & Shark             & 260.7   & 159.3   & 100   & 9.8e-1 & 599.4 & 241.8 & 100 & \textbf{-9.0e-3} & 35.5 & 15.3 &100 & 0 & $8\times$ \\
                                        \midrule
\multirow{1}{*}{\ref{sec:app:sim2real}}  & Tennis balls    &  54.6  & 6.4 & 14   & 7.2e-2 & 26.8 & 5.8 & 12 & 7.2e-2 & 24.1 & 15.9 & 41 & \textbf{6.9e-2}  & $0.8\times$ \\
\bottomrule
\end{tabular}
\end{table*}

\begin{figure}[ht]
    \centering
    \newcommand{\figheight}{0.235\linewidth}
    \rotatebox{90}{\small\hspace*{0.9em}Init guess}\hfill%
    \includegraphics[trim=140 120 150 70,clip,width=\figheight]{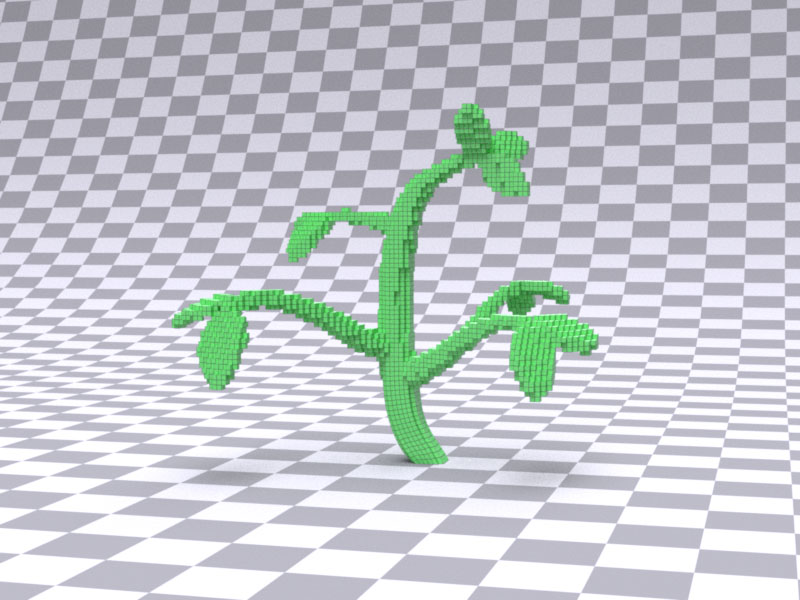}
    \plantimgwithbox{\figheight}{50,150,200}{220,150,50}{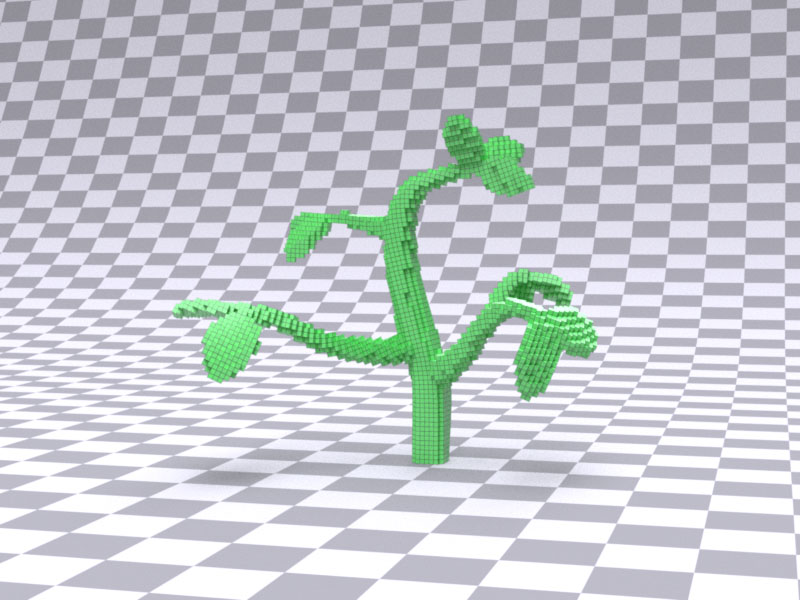}\hspace{-0.25em}
    \plantimgwithbox{\figheight}{50,150,200}{220,150,50}{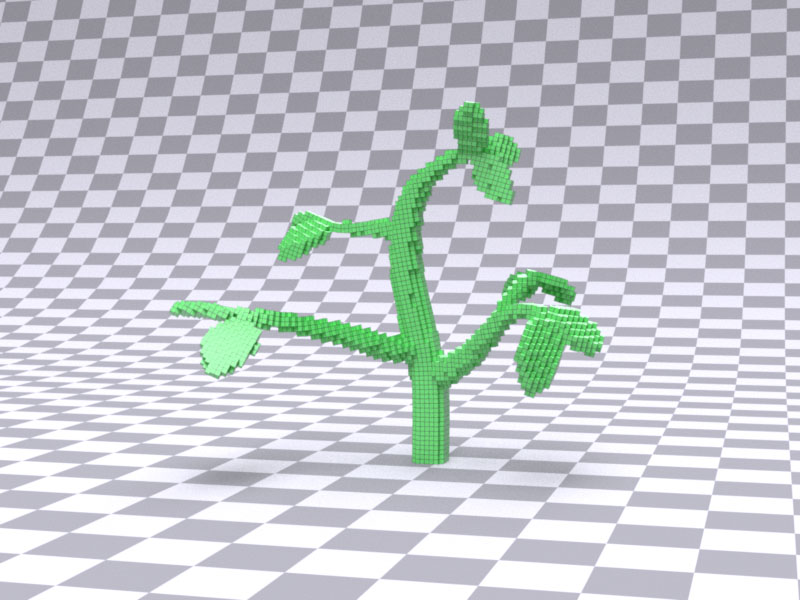}\hspace{-0.25em}
    \plantimgwithbox{\figheight}{50,150,200}{220,150,50}{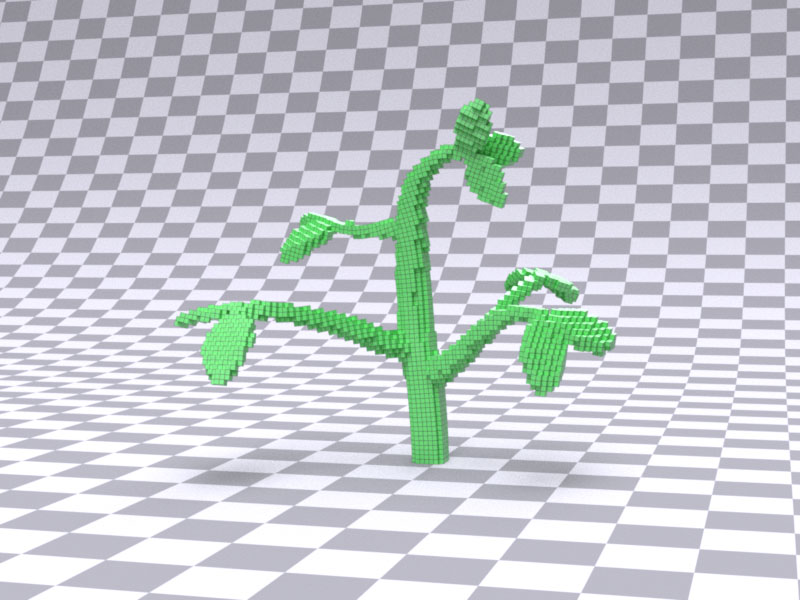} \\
    \rotatebox{90}{\small\hspace*{0.9em}Optimized}\hfill%
    \includegraphics[trim=140 120 150 70,clip,width=\figheight]{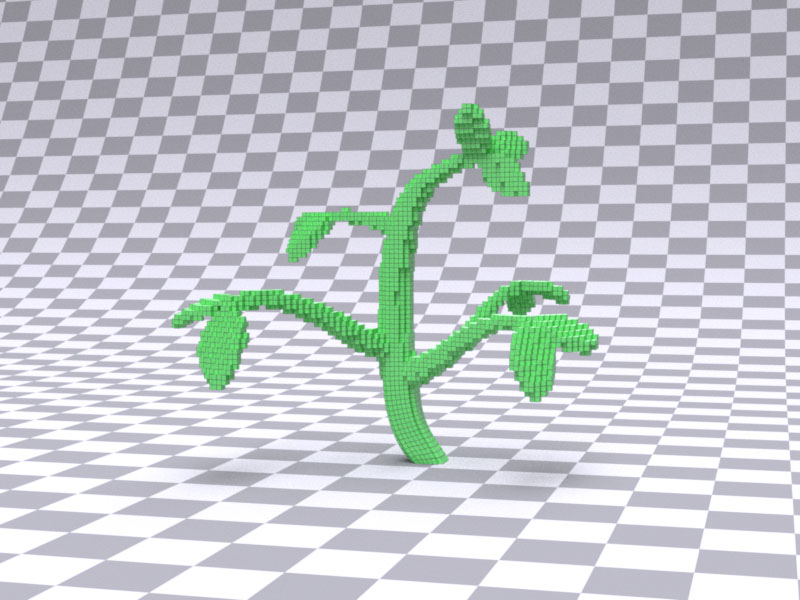}
    \plantimgwithbox{\figheight}{50,150,200}{220,150,50}{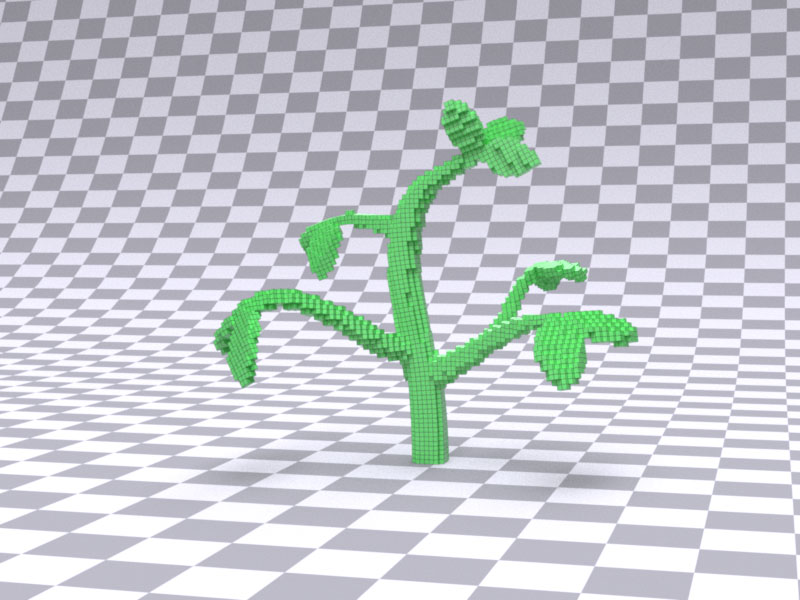}\hspace{-0.25em}
    \plantimgwithbox{\figheight}{50,150,200}{220,150,50}{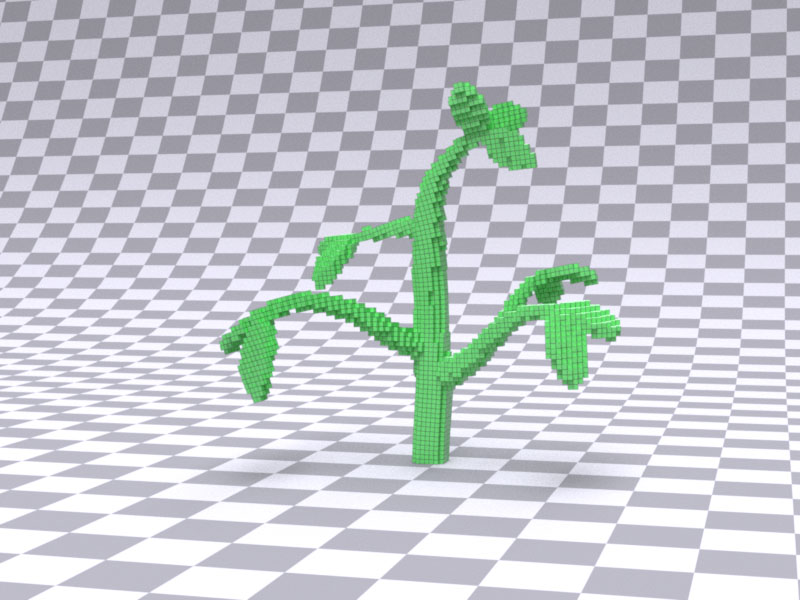}\hspace{-0.25em}
    \plantimgwithbox{\figheight}{50,150,200}{220,150,50}{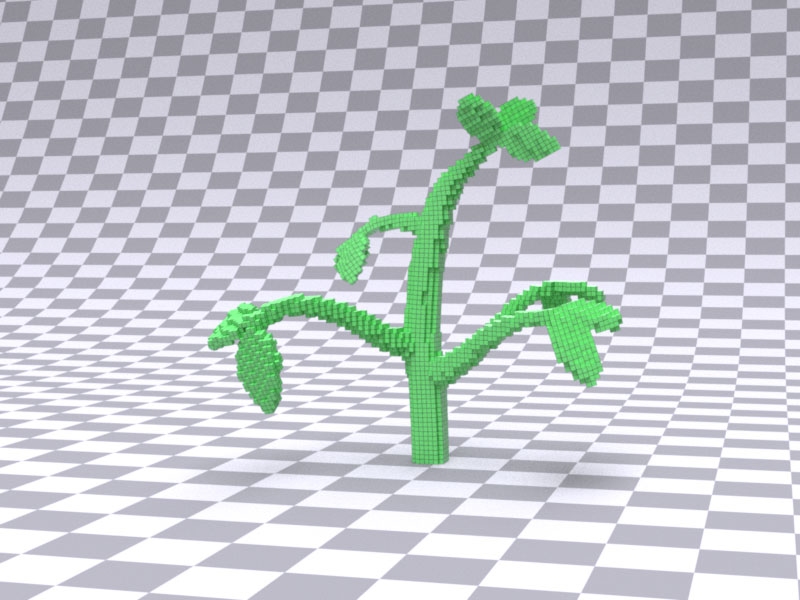} \\
    \rotatebox{90}{\small\hspace*{0em}Ground truth}\hfill%
    \includegraphics[trim=140 120 150 70,clip,width=\figheight]{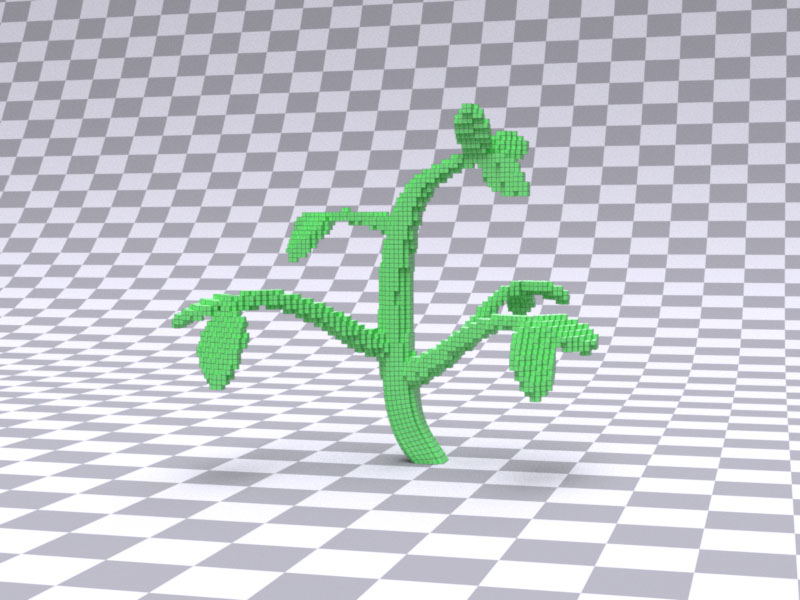}
    \plantimgwithbox{\figheight}{50,150,200}{220,150,50}{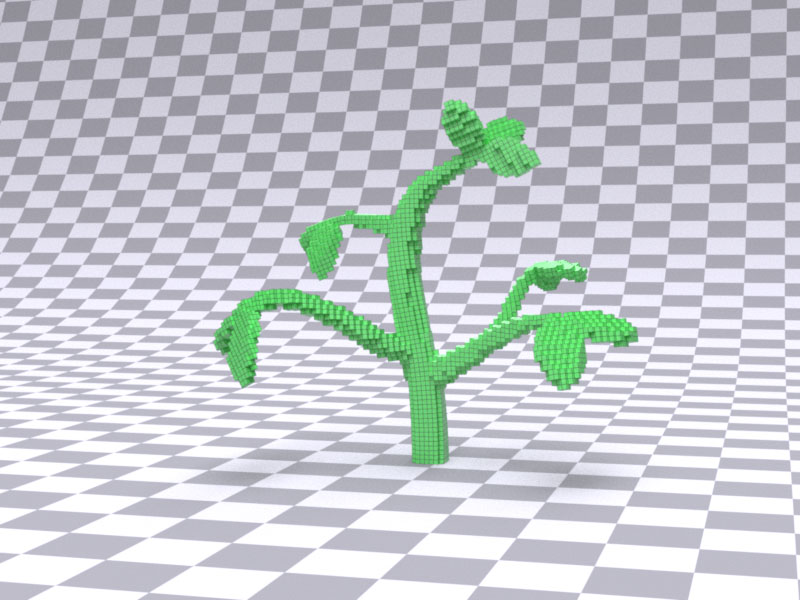}\hspace{-0.25em}
    \plantimgwithbox{\figheight}{50,150,200}{220,150,50}{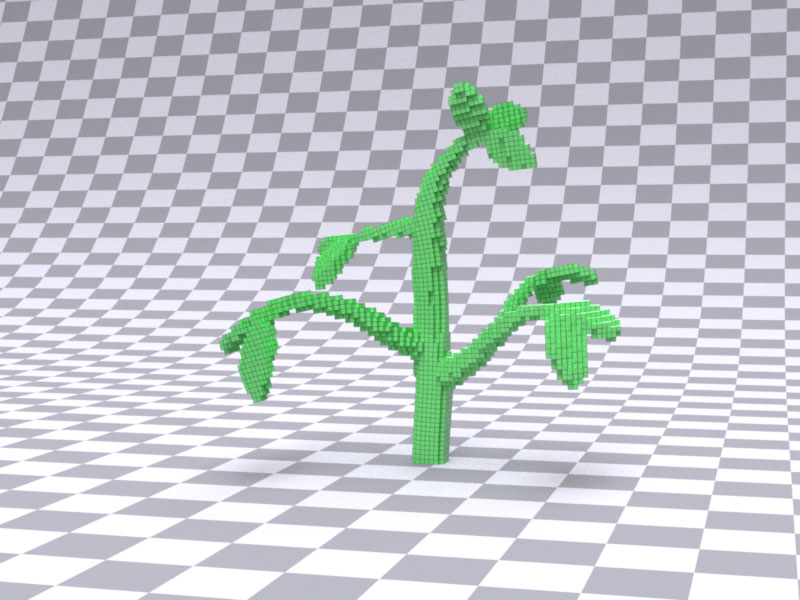}\hspace{-0.25em}
    \plantimgwithbox{\figheight}{50,150,200}{220,150,50}{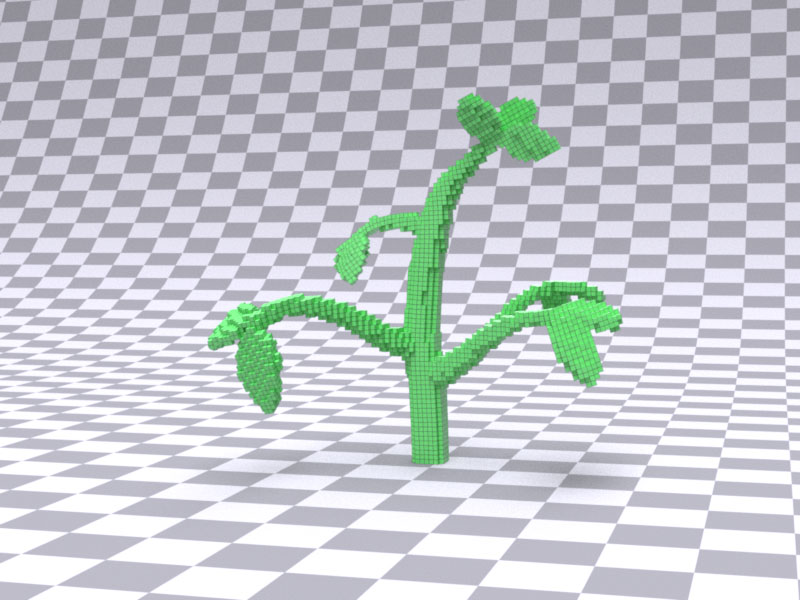}\\    
    \vspace{-1.2em}  
  \caption{\textbf{System identification} Motion sequences of the ``Plant'' example sampled at the 1st, 100th, 150th, and 200th (final) frames (left to right). We generated three motion sequences with a random initial guess of the material parameters (top row), optimized material parameters (middle row), and the ground truth (bottom row). The colored boxes highlight the motion differences before and after optimization. The goal is to optimize the material parameters so that the motion of the plant matches the ground truth.}
  \label{fig:plant}
  \Description{Plant.}
\end{figure}

\subsection{System Identification}\label{sec:app:parameter}

In this section, we discuss two examples that aim to estimate the material parameters (Young's modulus and Poisson's ratio) from dynamic motions of soft bodies: the ``Plant'' example estimates material parameters from its vibrations, and the ``Bouncing ball'' example predicts its parameters from its interaction with the ground. We generate the ground truth using our forward PD simulator with a set of predefined material parameters.

\paragraph{Plant} We first initialize an elastic, 3D house plant model with $3863$ hexahedral elements and $29763$ DoFs (Fig.~\ref{fig:plant}). We impose Dirichlet boundary conditions at the root of the plant such that it is fixed to the ground. We apply an initial horizontal force at the start of simulation, causing the plant to oscillate. Starting from an initial guess using randomly picked material parameters, we deform a new plant in the same manner as the ground truth and optimize the logarithm of the Young's modulus and Poisson's ratio of the new plant to match that of the ground truth plant.  The loss at each time step is determined as the squared sum of the element-wise difference in positions between the new plant and the reference plant.

After optimization, DiffPD, Newton-PCG, and Newton-Cholesky converge to local minima with a final Young's modulus of 1.00 MPa, 0.96 MPa, and 0.96 MPa respectively. Regarding the Poisson's ratio, DiffPD converge to 0.4 while both Newton's methods converged to 0.44. The reference plant is initialized with a Young's modulus of 1 MPa and a Poisson's ratio of 0.4. While the three methods all reach solutions that are similar to the ground truth, the optimization process is highly expedited by a factor of 9 for loss and gradient evaluation using our method (Table~\ref{tab:performance}). We observe that DiffPD converged to a solution closer to the ground truth but used more function evaluations due to the numerical difference between DiffPD and the Newton's method. However, if DiffPD terminated after the same number of function evaluations (10) as the Newton's method, the optimized Young's modulus and Poisson's ratio would be almost identical to results from the Newton's method (0.97 MPa for Young's modulus and 0.44 for Poisson's ratio), implying the $9\times$ speedup indeed comes from DiffPD's improvements over the Newton's method on the simulation side.

\begin{figure}[ht]
  \centering
    \rotatebox{90}{\small\hspace*{1.5em}Init guess}\hfill%
    \includegraphics[trim=150 150 370 250,clip,width=0.31\linewidth]{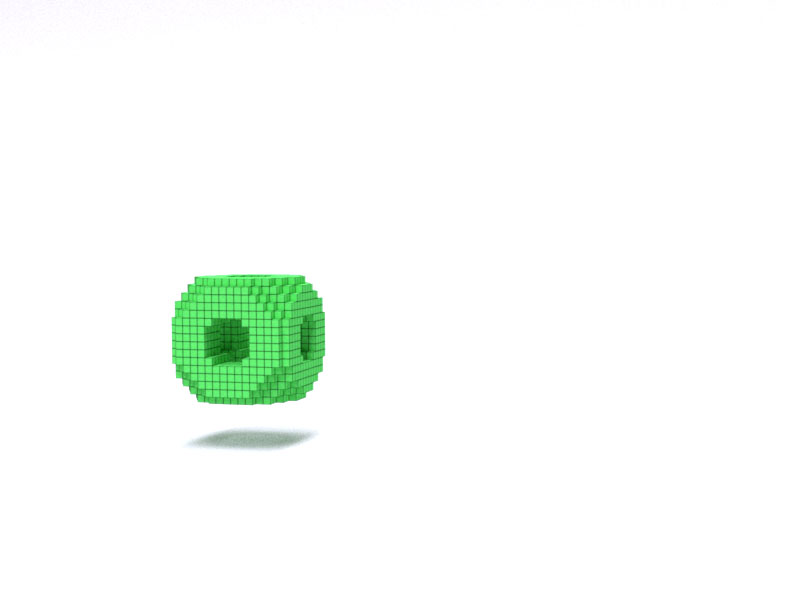}
    \includegraphics[trim=220 150 300 250,clip,width=0.31\linewidth]{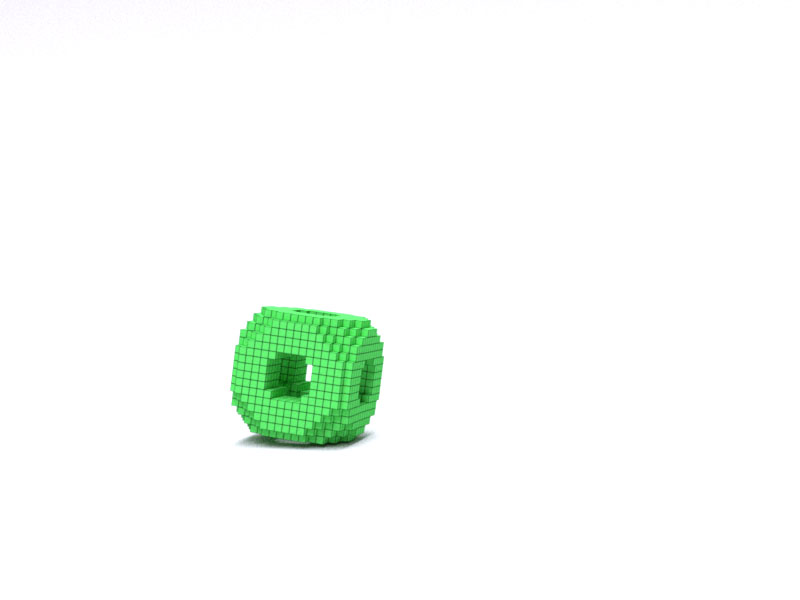}
    \includegraphics[trim=290 150 230 250,clip,width=0.31\linewidth]{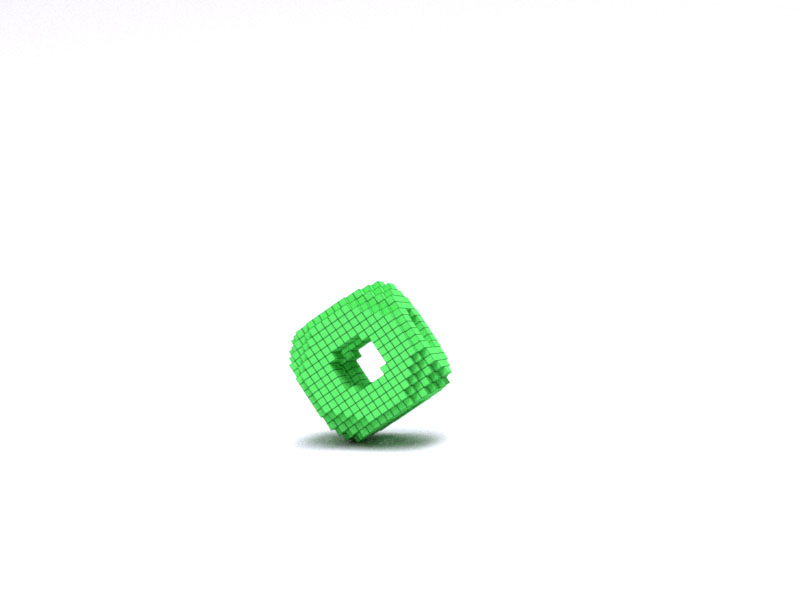} \\
    \rotatebox{90}{\small\hspace*{1.2em}Optimized}\hfill%
    \includegraphics[trim=150 150 370 250,clip,width=0.31\linewidth]{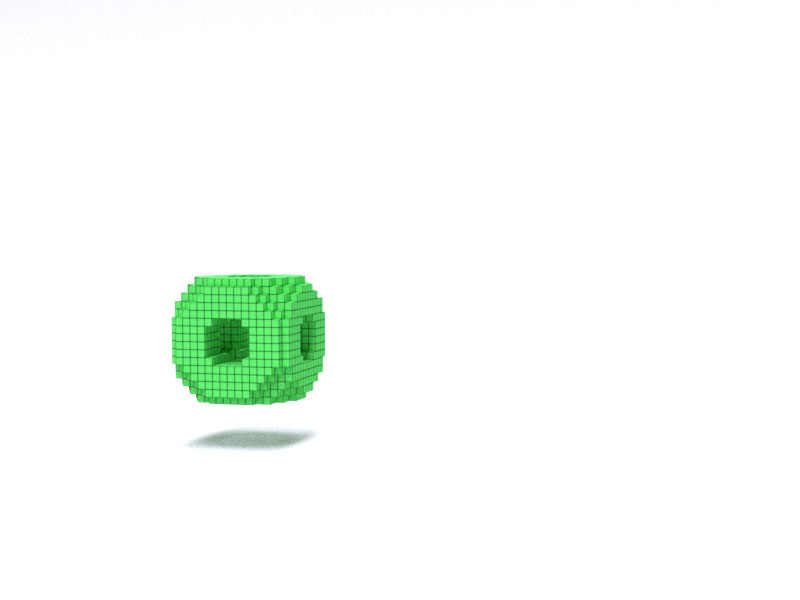}
    \includegraphics[trim=220 150 300 250,clip,width=0.31\linewidth]{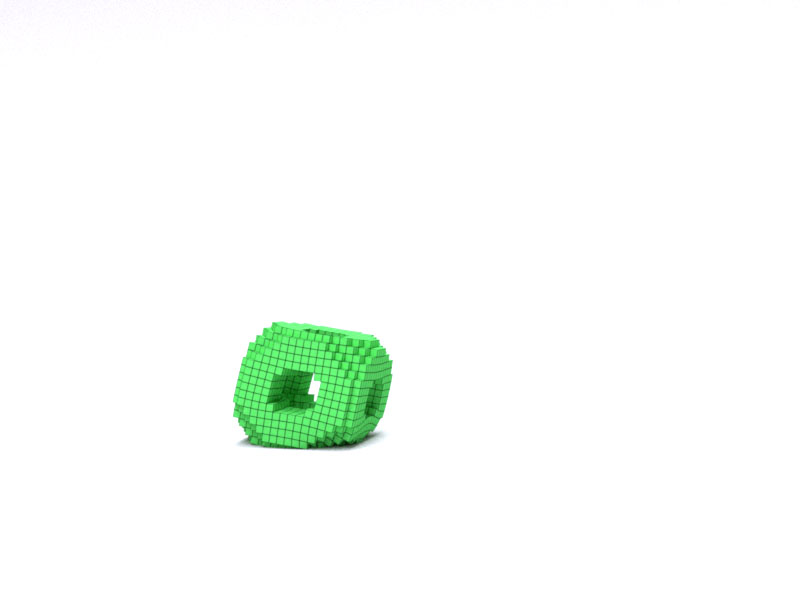}
    \includegraphics[trim=290 150 230 250,clip,width=0.31\linewidth]{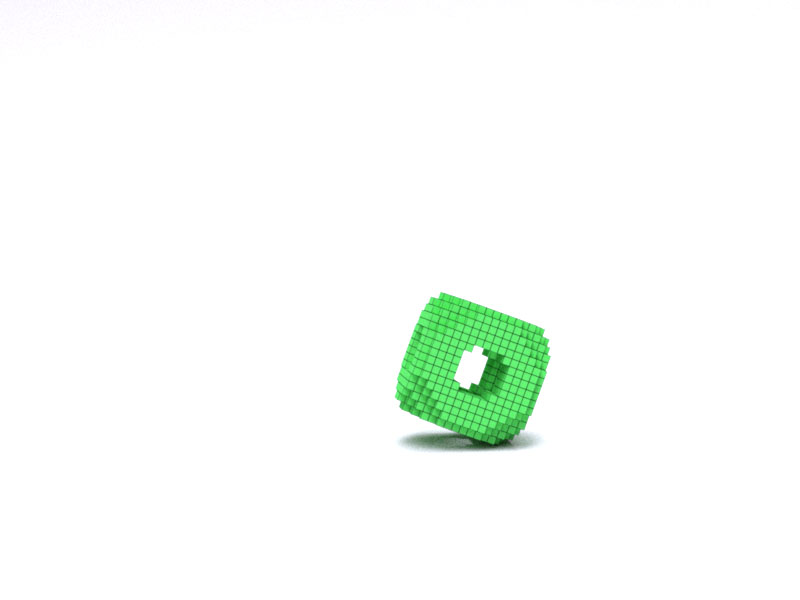} \\
    \rotatebox{90}{\small\hspace*{0em}Ground truth}\hfill%
    \includegraphics[trim=150 150 370 250,clip,width=0.31\linewidth]{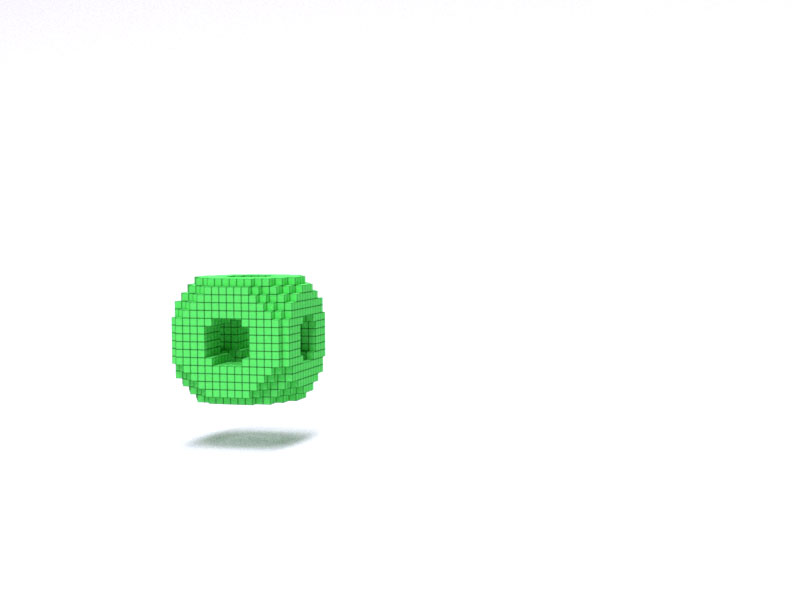}
    \includegraphics[trim=220 150 300 250,clip,width=0.31\linewidth]{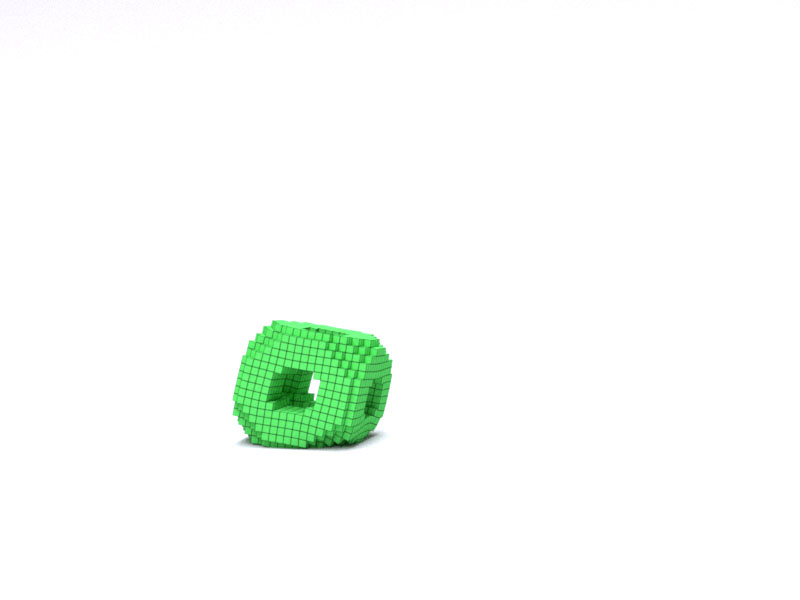}
    \includegraphics[trim=290 150 230 250,clip,width=0.31\linewidth]{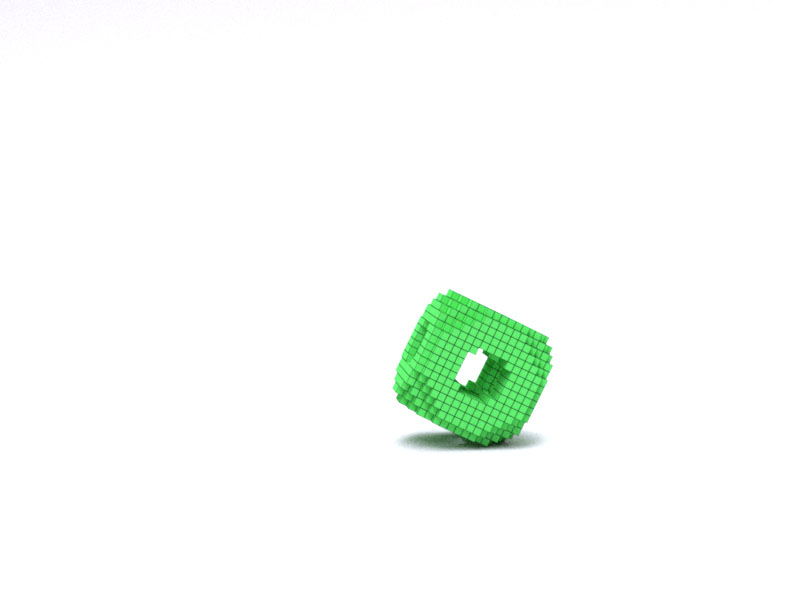}
    \vspace{-0.8em}  
  \caption{\textbf{System identification} Motion sequences of the ``Bouncing ball'' example sampled at the 1st frame (left), the 19th frame when collision occurs (middle), and the 125th (final) frame (right). We generated three motion sequences with a random initial guess of the material parameters (top row), optimized material parameters (middle row), and the ground truth (bottom row). The goal is to optimize the material parameters so that the motion of the ball matches the ground truth.}
  \label{fig:bouncing_ball}
  \Description{Bouncing ball.}
\end{figure}

\paragraph{Bouncing ball}  In this example, we consider a ball with 1288 hexahedral elements and 9132 DoFs thrown at the ground from a known initial position (Fig.~\ref{fig:bouncing_ball}). The ball has three cylindrical holes extruded through the faces in order to produce more complex deformation behavior than a fully solid ball.  This example uses the complementarity-based contact model in Sec.~\ref{sec:contact:compl}. We can estimate the material parameters of a bouncing ball by observing its behavior after it collides the ground. The loss definition is the same as in the parameter estimation of the ``Plant'' example. Regarding the optimization process, all three methods converge to a Young's modulus of 1.78 MPa and Poisson's ratio of 0.2. The ground truth values for the Young's modulus and Poisson's ratio are 2 MPa and 0.4, respectively. While the optimized material parameters are significantly different from the ground truth values, the motion sequences are very similar as reflected by the final loss in Table~\ref{tab:performance} and Fig.~\ref{fig:bouncing_ball}. Since the loss function is defined on the motion only, there could exist many material parameters that result in close-to-zero loss. As in the ``Plant'' example, our method enjoys a substantial speedup (12$\times$) in computation time even with collisions in simulation.

\subsection{Initial State Optimization}\label{sec:app:initial}

We present two examples demonstrating the power of using gradient information to optimize the initial configuration of a soft-body task. In the ``Bunny'' example, we optimize the initial position and velocity of a soft Stanford bunny so that its bounce trajectory ends at a target position. In the ``Routing tendon'' example, we optimize a constant actuation signals applied to each muscle in a soft cuboid with one face sticky on the ground so that the corner at the opposite face reaches a target point at the end of simulation.

\begin{figure}[htb]
  \centering
  \includegraphics[width=\linewidth]{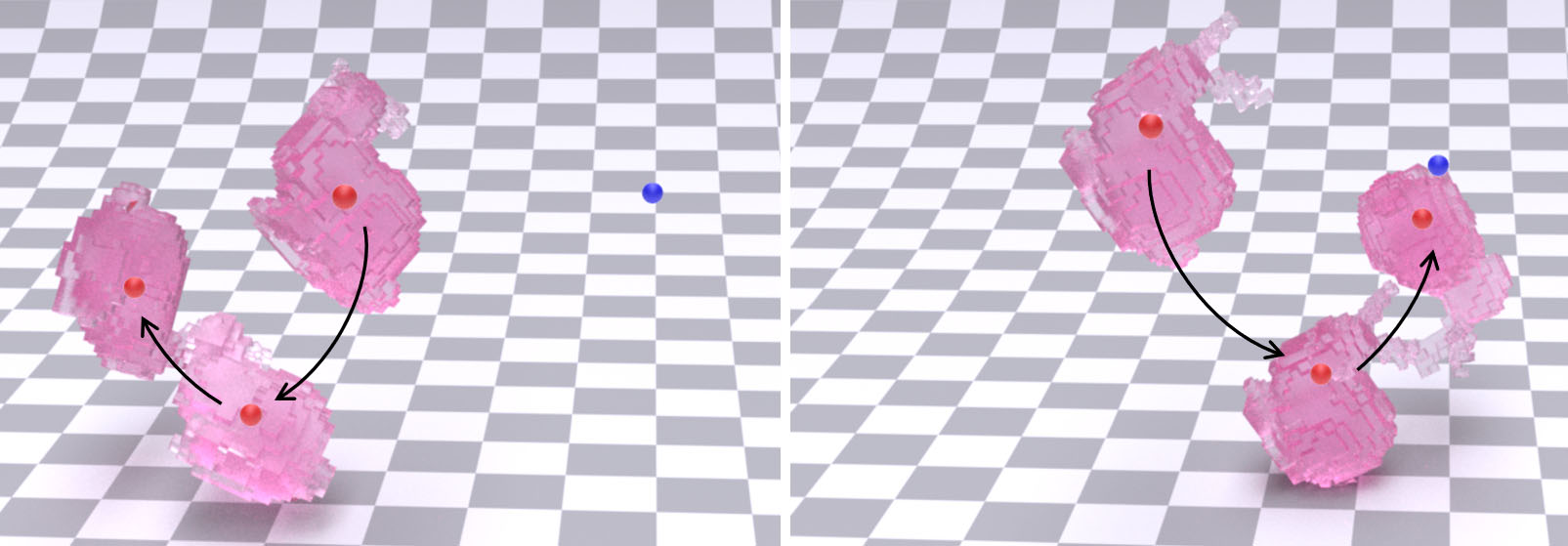}
  \vspace{-2em}
  \caption{\textbf{Inverse design} Initial (left) and optimized (right) trajectories of the ``Bunny'' example. The red dots indicate the location of the center of mass of the bunny, and the blue dot is the target location. The goal is to adjust the initial position, velocity, and orientation of the bunny so that its final center of mass can reach the target location.}
  \label{fig:bunny}
  \Description{Bunny.}
\end{figure}

\paragraph{Bunny} For this example, we optimize the initial pose and velocity of a Stanford bunny (1601 elements and 7062 DoFs) so that its center of mass (red dots in Fig.~\ref{fig:bunny}) reaches a target position (blue dot in Fig.~\ref{fig:bunny}) when the simulation finishes. This example uses the complementarity-based contact model, and we add 251 surface vertices (753 DoFs) to the set of possible contact nodes -- approximately 10.7\% of the 2354 vertices. Fig.~\ref{fig:bunny} illustrates the trajectory of the bunny before and after optimization: the initial guess generates a trajectory almost to the opposite direction of the target, and the optimized trajectory ends much closer to the target. Note that none of the three methods solve this task perfectly: the trajectory does not reach the target even after optimization. This is because the target is chosen arbitrarily rather than generated from simulating a ground truth bounce trajectory, so it is not guaranteed to be reachable. Table~\ref{tab:performance} shows that the final loss from DiffPD is larger than from the Newton's method, but the increase in performance makes up for it. Using 8 threads, our method achieves a speedup of 9 times overall with a large set of potential contact points.

\begin{figure}[htb]
  \centering
    \includegraphics[trim=280 80 290 155,clip,width=0.22\linewidth]{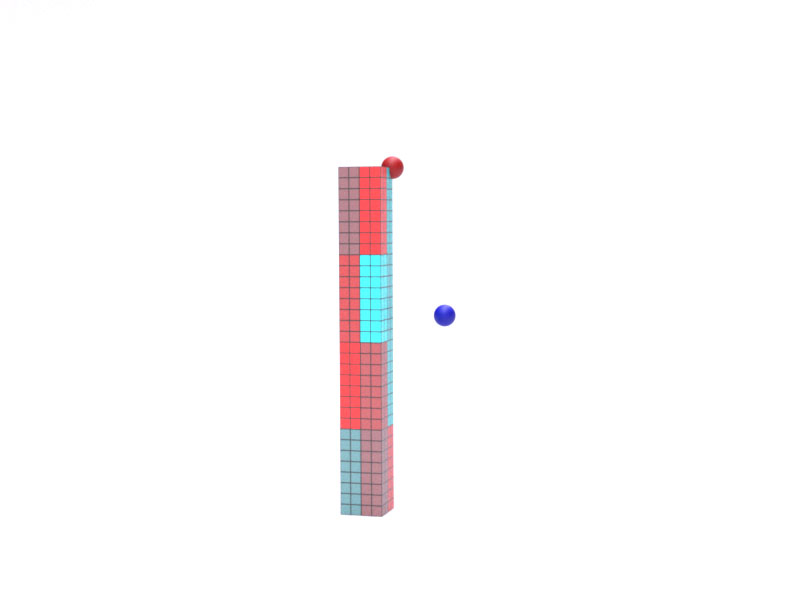}
    \includegraphics[trim=280 80 290 155,clip,width=0.22\linewidth]{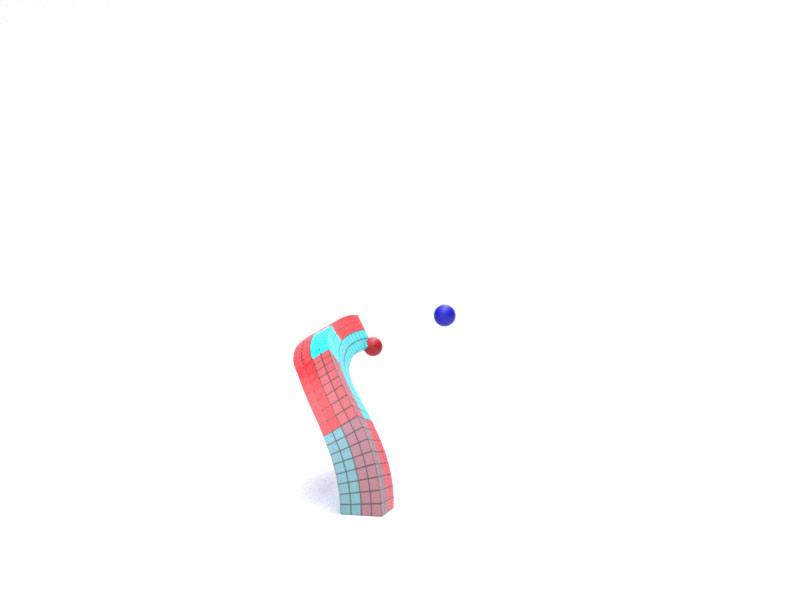}\hfill
    \includegraphics[trim=280 80 290 155,clip,width=0.22\linewidth]{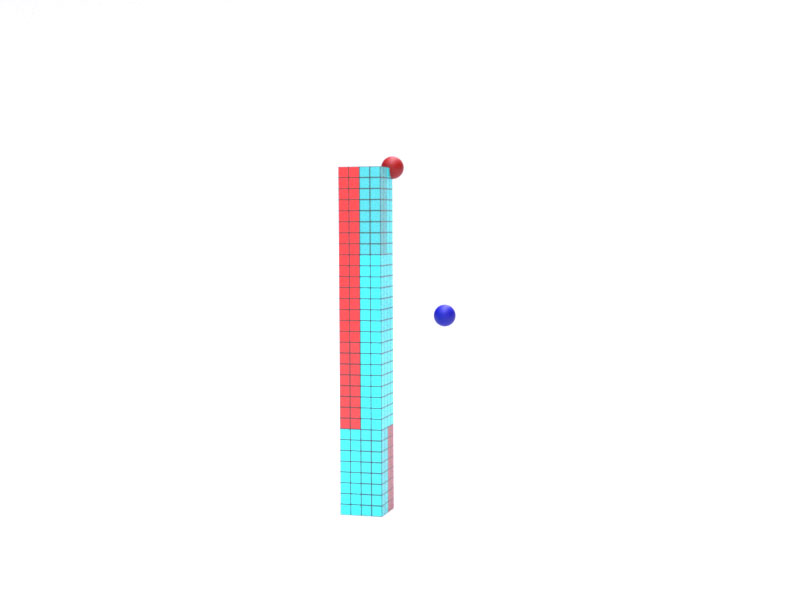}
    \includegraphics[trim=280 80 290 155,clip,width=0.22\linewidth]{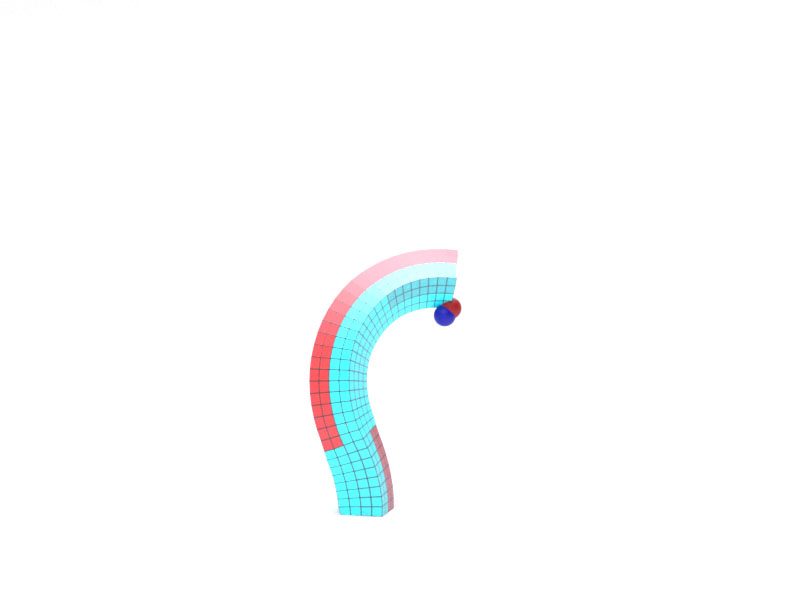}\\
    \vspace{-9.7em}
    \hspace{20em}
    \begin{minipage}[t]{0.1\textwidth}
    \includegraphics[width=\textwidth]{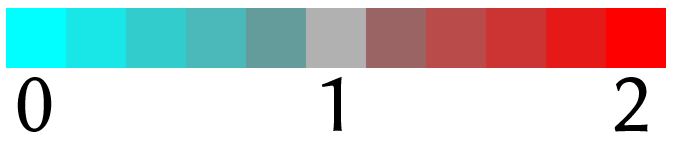}
    \end{minipage}
    \vspace{9em}
    \vspace{-1em}
  \caption{\textbf{Inverse design} The initial and final states of the ``Routing tendon'' example before and after optimization. The goal is to let the end effector (red dot) hit a target position (blue dot) at the end of simulation. Left: the initial configuration of the tendon with randomly generated actuation signals. The red (muscle expansion) and cyan (muscle contraction) colors indicate the magnitude of the action. Middle left: final state of the tendon with random actuation. Middle right: initial configuration of the tendon with optimized actuation. Right: final state with optimized actuation.}
  \label{fig:tendon}
  \Description{Tendon.}
\end{figure}

\paragraph{Routing tendon} We initialize a soft cuboid with 512 elements and 2475 DoFs and impose Dirichlet boundary conditions such that its bottom face is stuck to the ground. We also add actuators to each element and group them into 16 muscle groups. The level of actuator activation is a scalar between $0$ to $1$, indicating muscle contraction and expansion, respectively. The elements within a specific actuation group all share the same, time-invariant actuation signal to be optimized in order to manipulate the endpoint (red dot in Fig.~\ref{fig:tendon}) of the soft body to reach a target point (blue dot in Fig.~\ref{fig:tendon}). The normalized losses at the final iteration for each of the methods (Table~\ref{tab:performance}) are all close to zero, indicating that the task is solved almost perfectly. Using DiffPD, we observe a $9\times$ speedup over the Newton's methods.

\subsection{Trajectory Optimization}\label{sec:app:plan}

\begin{figure*}[htb]
  \centering
    \rotatebox{90}{\small\hspace*{2em}Random}~%
    \includegraphics[trim=105 60 175 310,clip,width=0.24\linewidth]{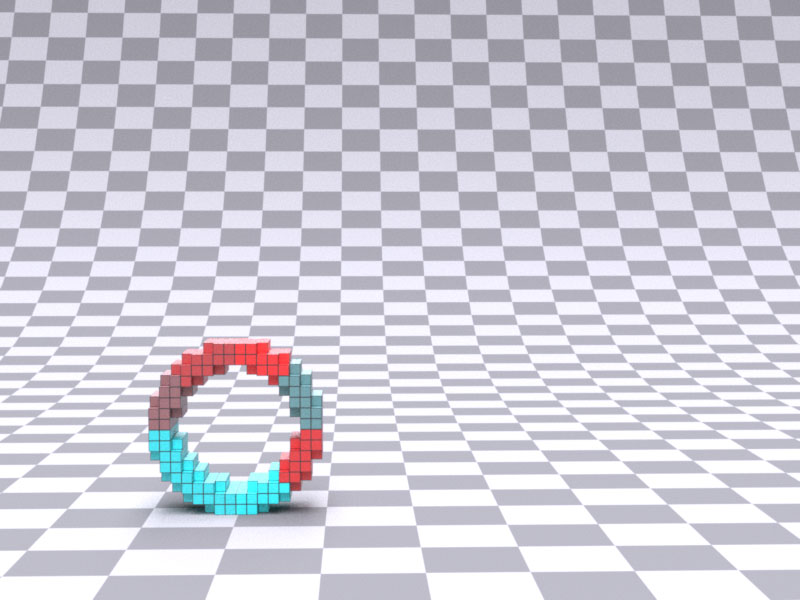}
    \includegraphics[trim=105 60 175 310,clip,width=0.24\linewidth]{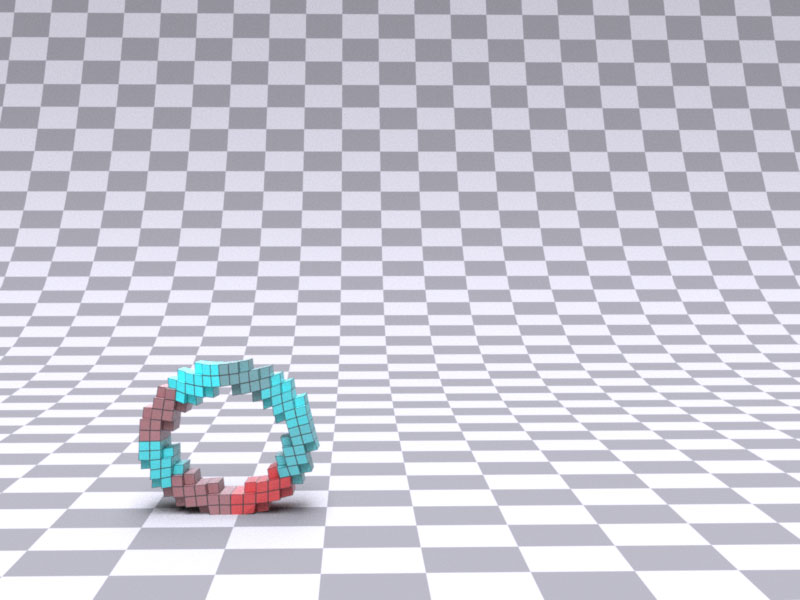}
    \includegraphics[trim=105 60 175 310,clip,width=0.24\linewidth]{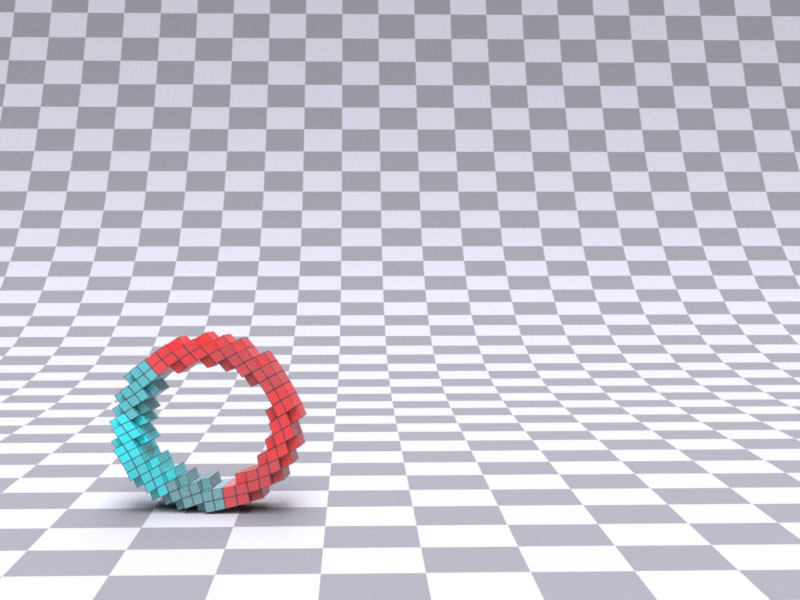}
    \includegraphics[trim=105 60 175 310,clip,width=0.24\linewidth]{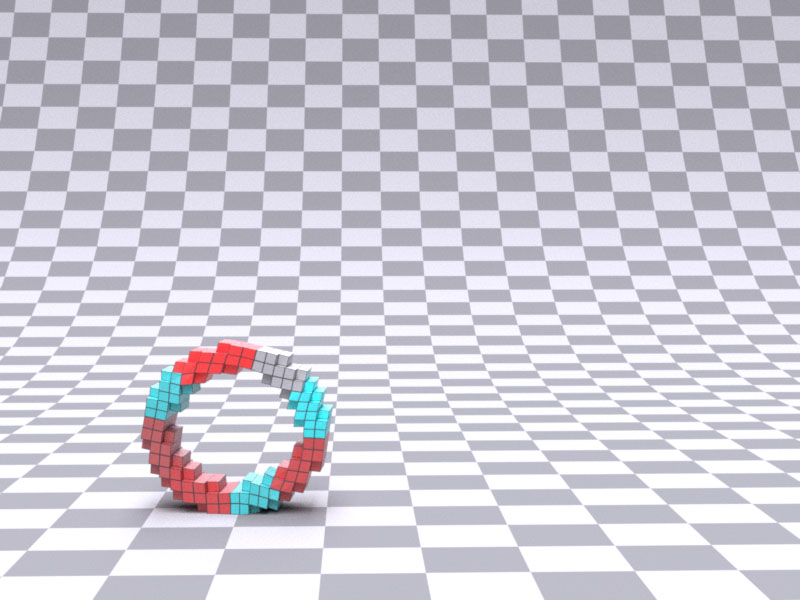} \\
    \rotatebox{90}{\small\hspace*{1.3em}Optimized}~%
    \includegraphics[trim=105 60 175 310,clip,width=0.24\linewidth]{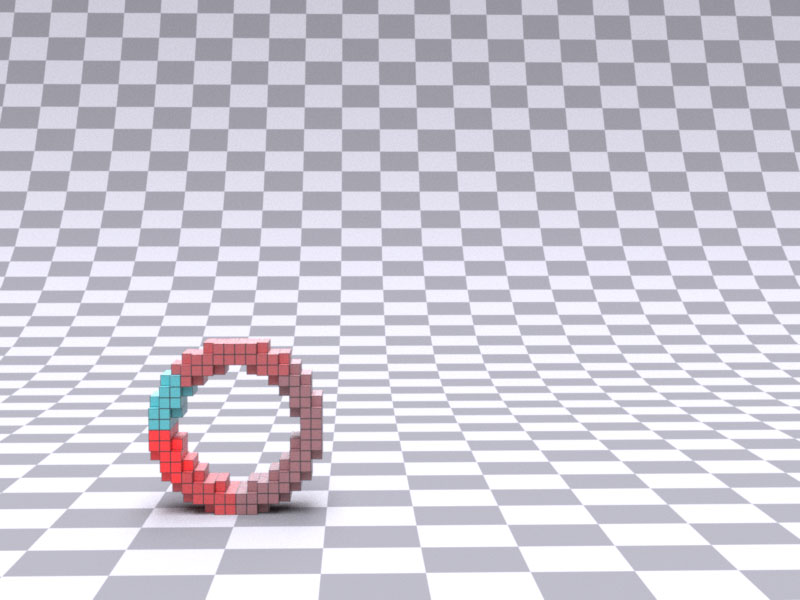}
    \includegraphics[trim=105 60 175 310,clip,width=0.24\linewidth]{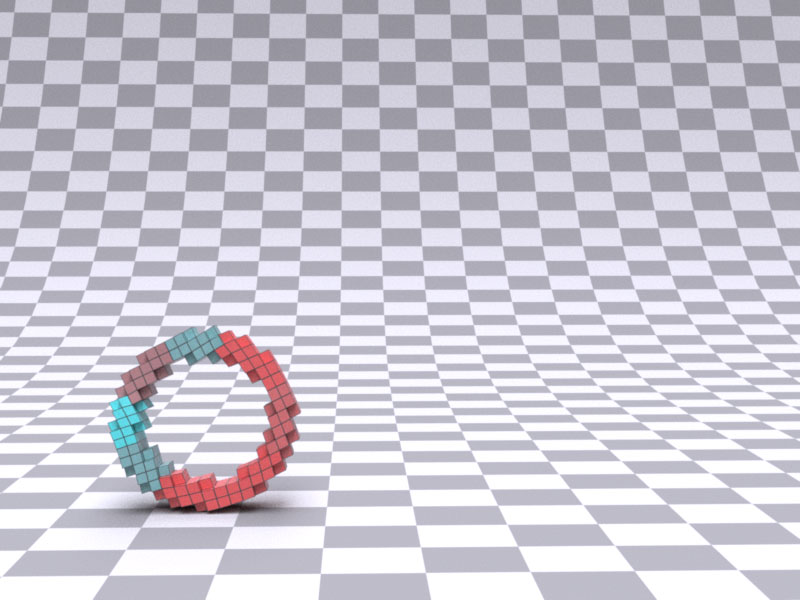}
    \includegraphics[trim=105 60 175 310,clip,width=0.24\linewidth]{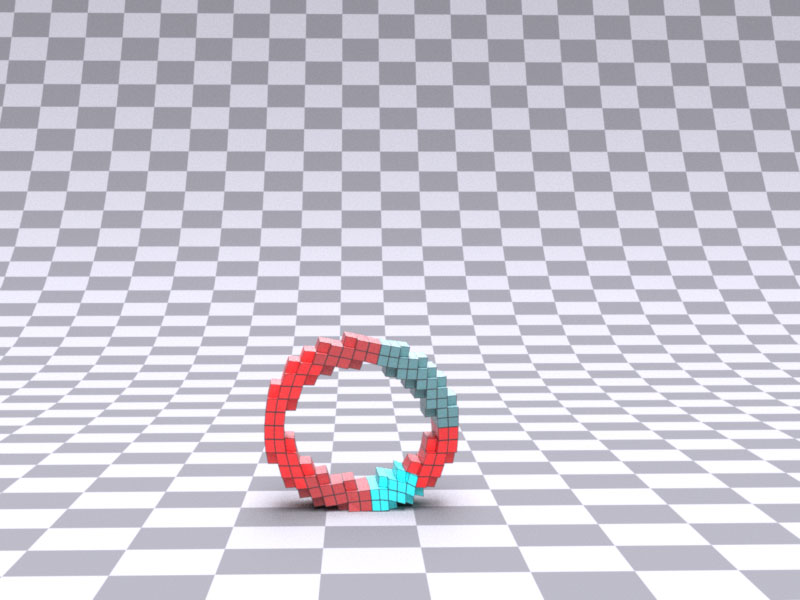}
    \includegraphics[trim=105 60 175 310,clip,width=0.24\linewidth]{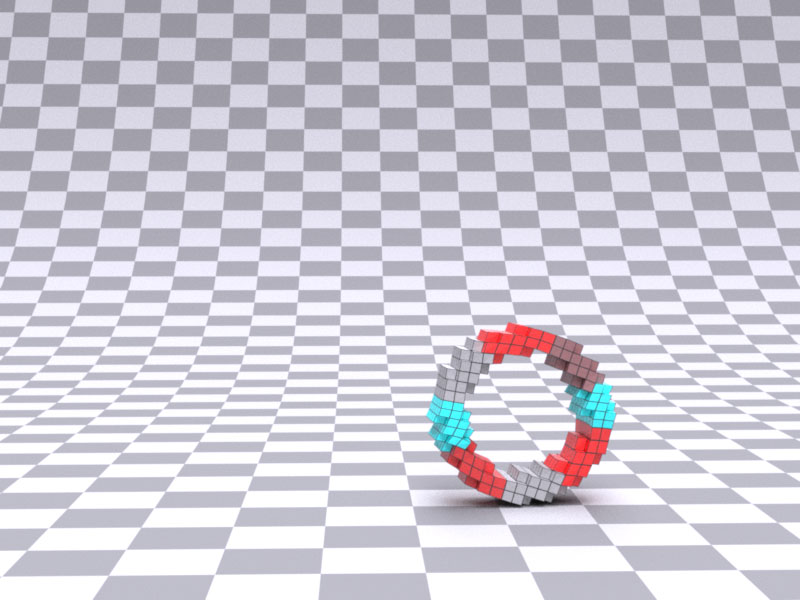}\\
        \vspace{-12.2em}
    \hspace{50em}
    \begin{minipage}[t]{0.1\textwidth}
    \includegraphics[width=\textwidth]{figure/colormap.JPG}
    \end{minipage}
    \vspace{10.em}
  \caption{\textbf{Trajectory optimization} The motion sequence of the ``Torus'' example with random actions (top) and after optimizing the action sequences (160 parameters to be optimized) with DiffPD (bottom). The goal is to maximize the rolling distance of the torus while maintaining its balance. The red and cyan color indicates the magnitude of the action signal. In particular, the expansion and contraction pattern (middle left and middle right) allows the torus to roll forward.}
  \label{fig:torus}
  \Description{Torus.}
\end{figure*}

\begin{figure*}[htb]
  \centering
\rotatebox{90}{\small\hspace*{1em}Random}~%
    \includegraphics[trim=120 50 30 320,clip,width=0.24\linewidth]{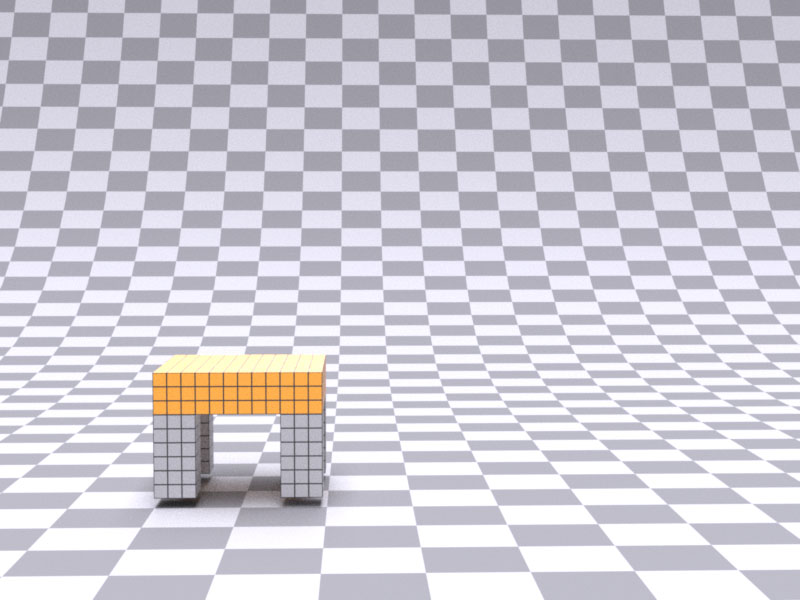}
    \includegraphics[trim=120 50 30 320,clip,width=0.24\linewidth]{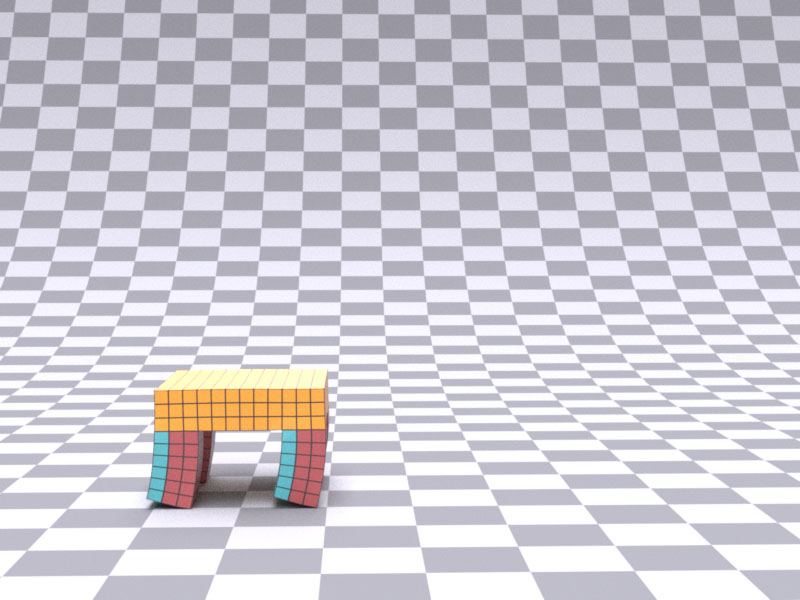}
    \includegraphics[trim=120 50 30 320,clip,width=0.24\linewidth]{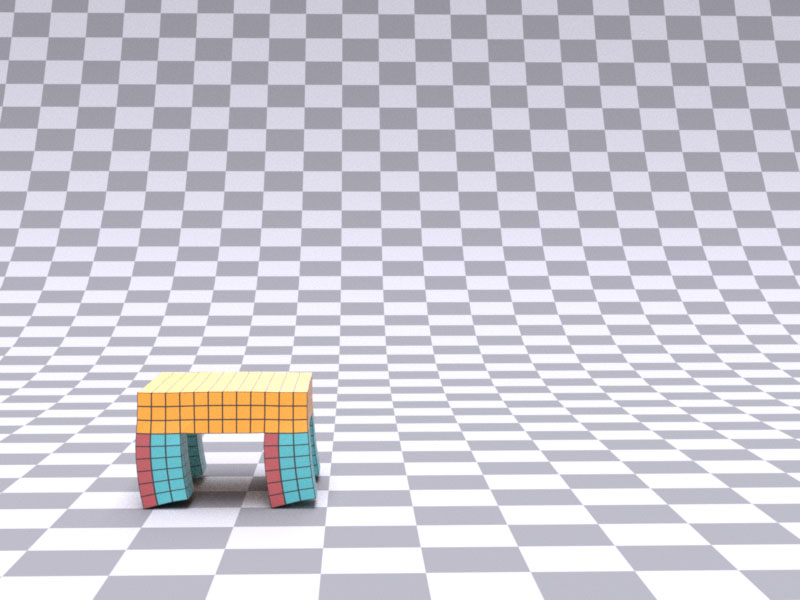}
    \includegraphics[trim=120 50 30 320,clip,width=0.24\linewidth]{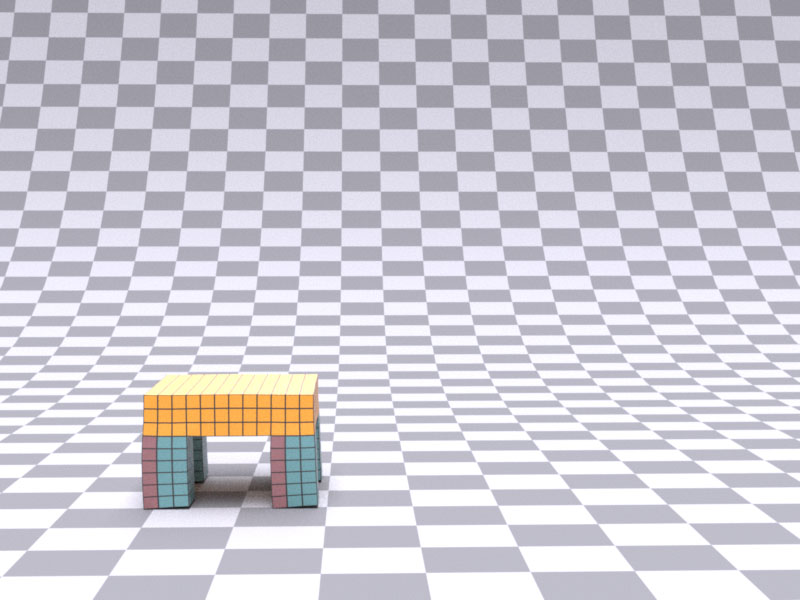} \\
    \rotatebox{90}{\small\hspace*{0.8em}Optimized}~%
    \includegraphics[trim=120 50 30 320,clip,width=0.24\linewidth]{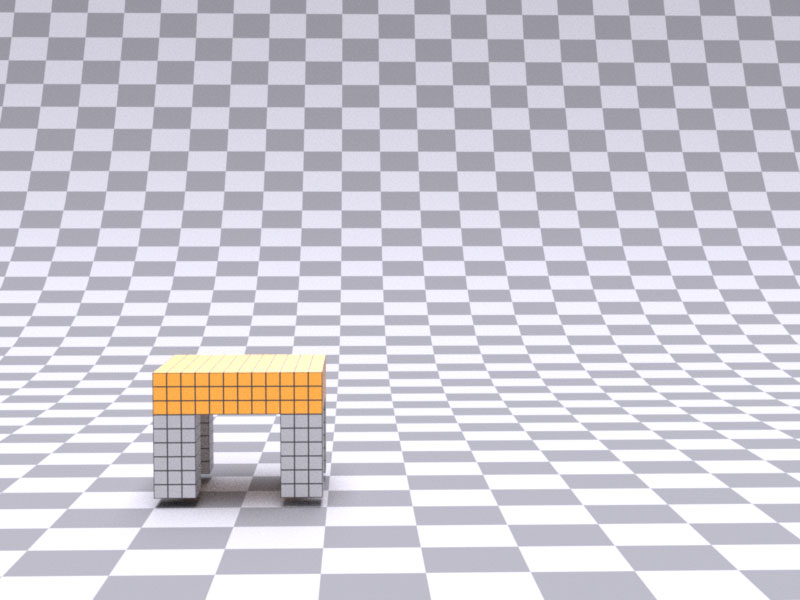}
    \includegraphics[trim=120 50 30 320,clip,width=0.24\linewidth]{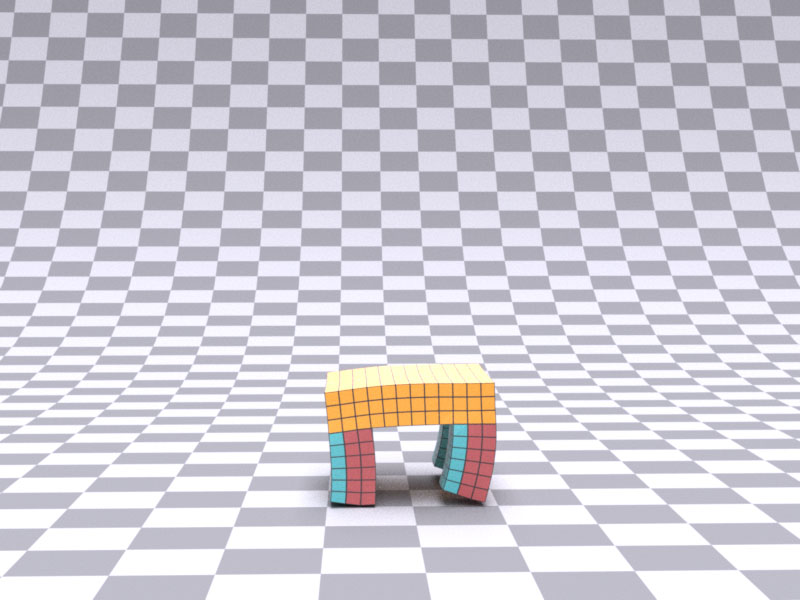}
    \includegraphics[trim=120 50 30 320,clip,width=0.24\linewidth]{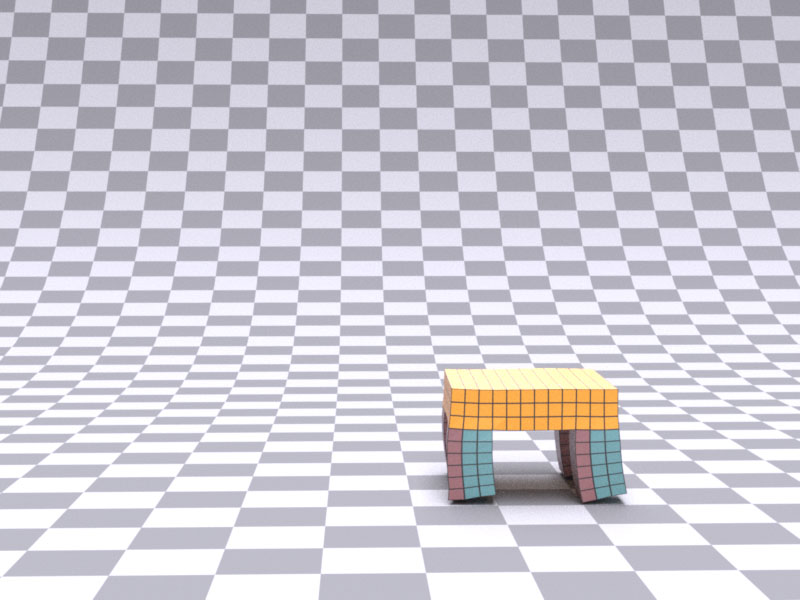}
    \includegraphics[trim=120 50 30 320,clip,width=0.24\linewidth]{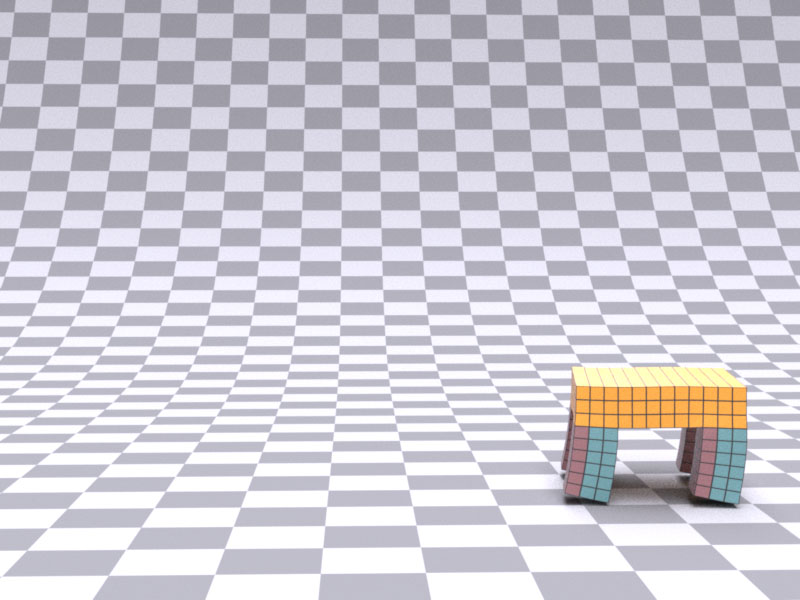} \\
    \vspace{-9.7em}
    \hspace{50em}
    \begin{minipage}[t]{0.1\textwidth}
    \includegraphics[width=\textwidth]{figure/colormap.JPG}
    \end{minipage}
    \vspace{7.5em}
  \caption{\textbf{Trajectory optimization} The motion sequences of the ``Quadruped'' example with sinusoidal control signals whose 3 parameters are to be optimized. The goal is to maximize the walking distance of the quadruped. Top: the motion sequence with a random sinusoidal wave of actions. Bottom: the motion sequence after optimization with DiffPD.}
  \label{fig:quadruped}
  \Description{Quadruped.}
\end{figure*}

\begin{figure*}[htb]
  \centering
\rotatebox{90}{\small\hspace*{1em}Random}~%
    \includegraphics[trim=120 80 100 290,clip,width=0.24\linewidth]{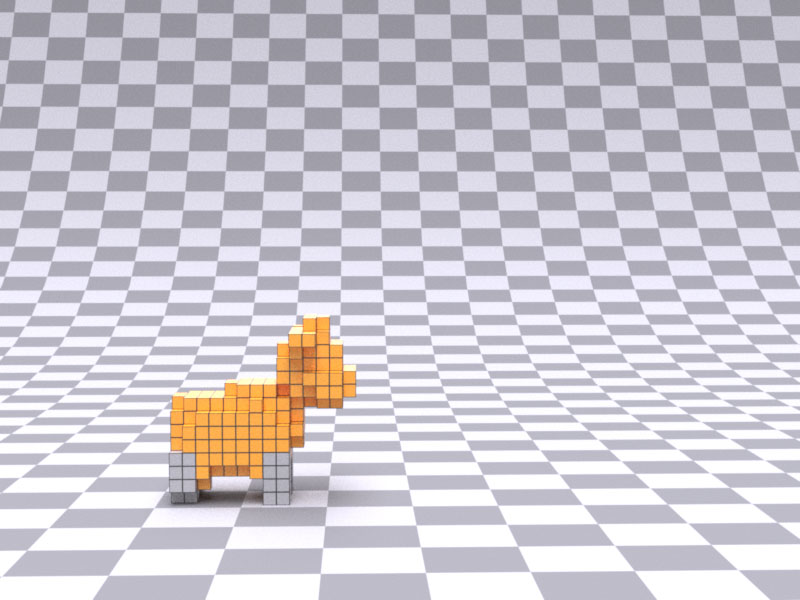}
    \includegraphics[trim=120 80 100 290,clip,width=0.24\linewidth]{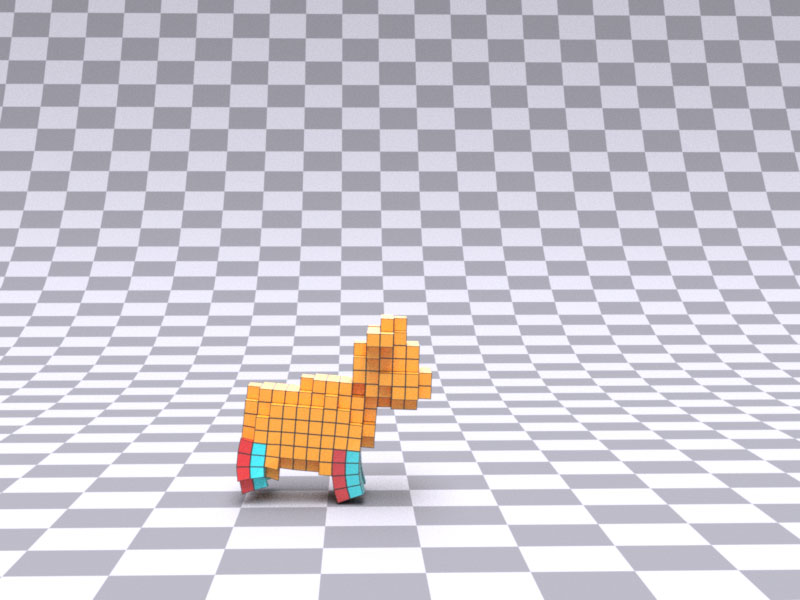}
    \includegraphics[trim=120 80 100 290,clip,width=0.24\linewidth]{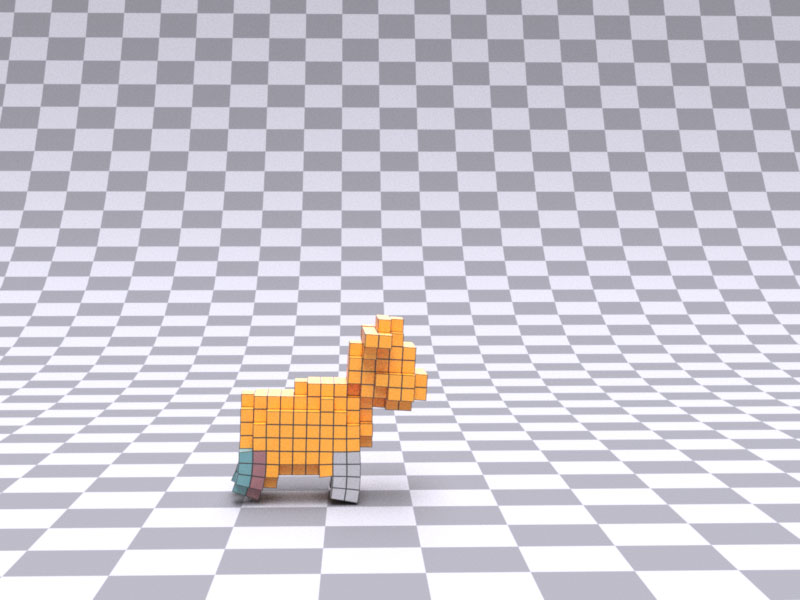}
    \includegraphics[trim=120 80 100 290,clip,width=0.24\linewidth]{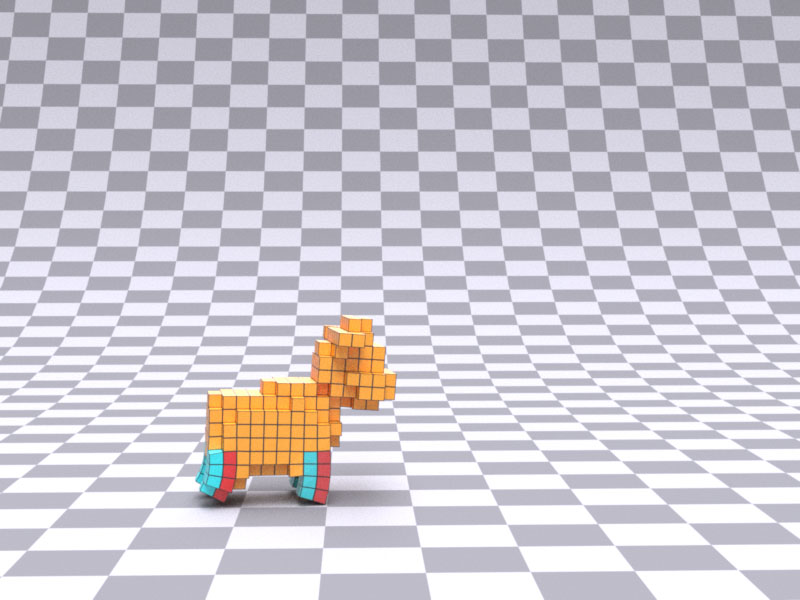} \\
    \rotatebox{90}{\small\hspace*{0.8em}Optimized}~%
    \includegraphics[trim=120 80 100 290,clip,width=0.24\linewidth]{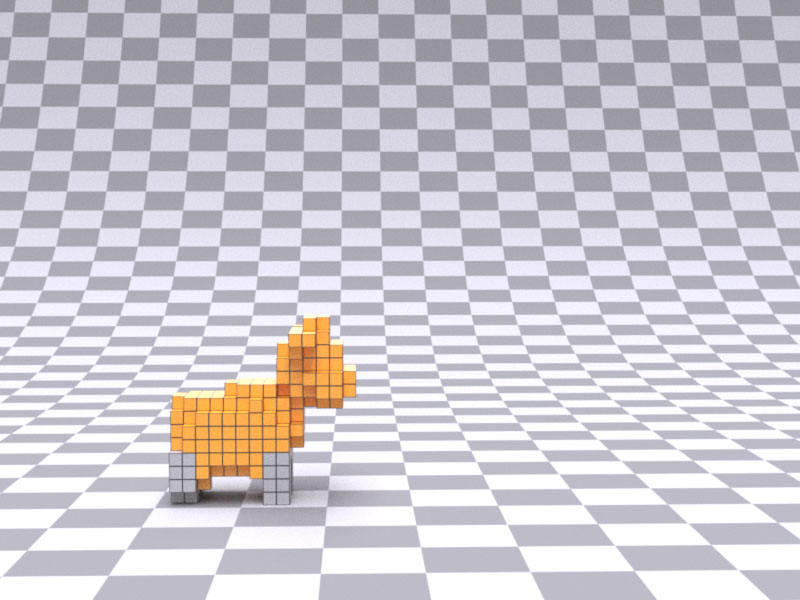}
    \includegraphics[trim=120 80 100 290,clip,width=0.24\linewidth]{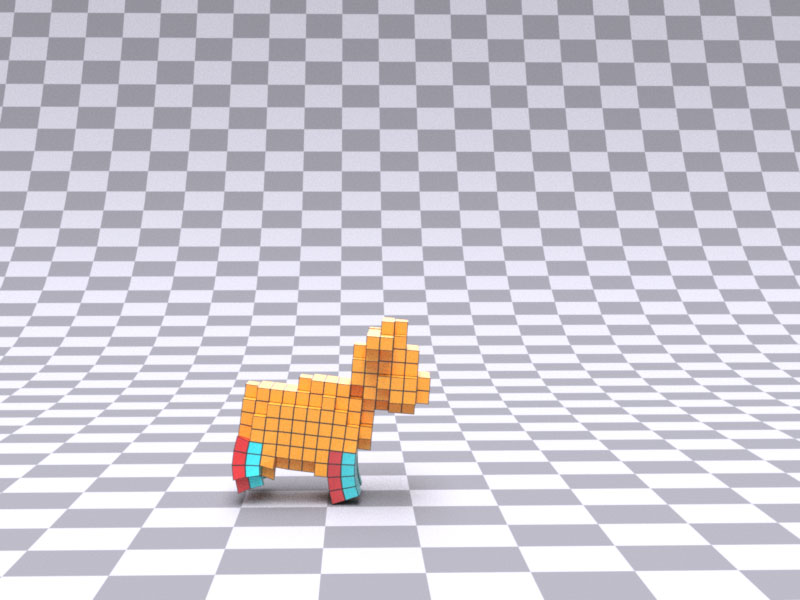}
    \includegraphics[trim=120 80 100 290,clip,width=0.24\linewidth]{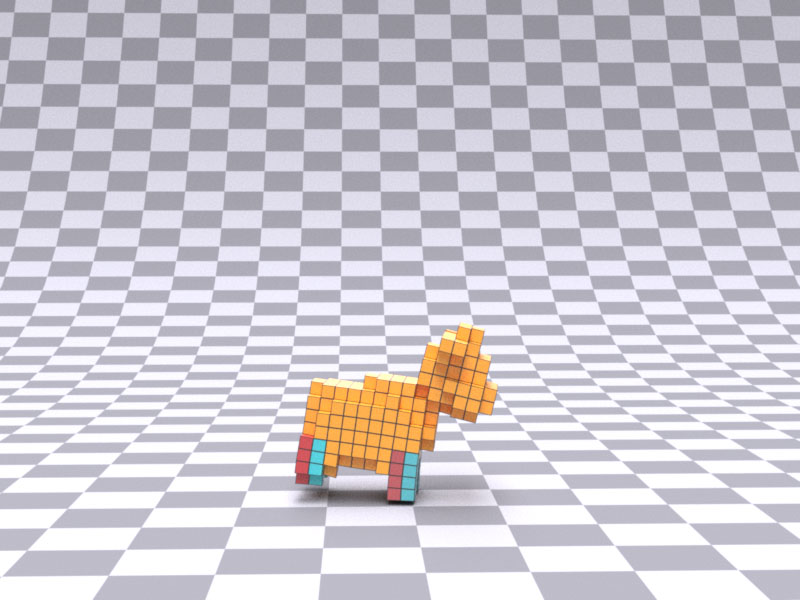}
    \includegraphics[trim=120 80 100 290,clip,width=0.24\linewidth]{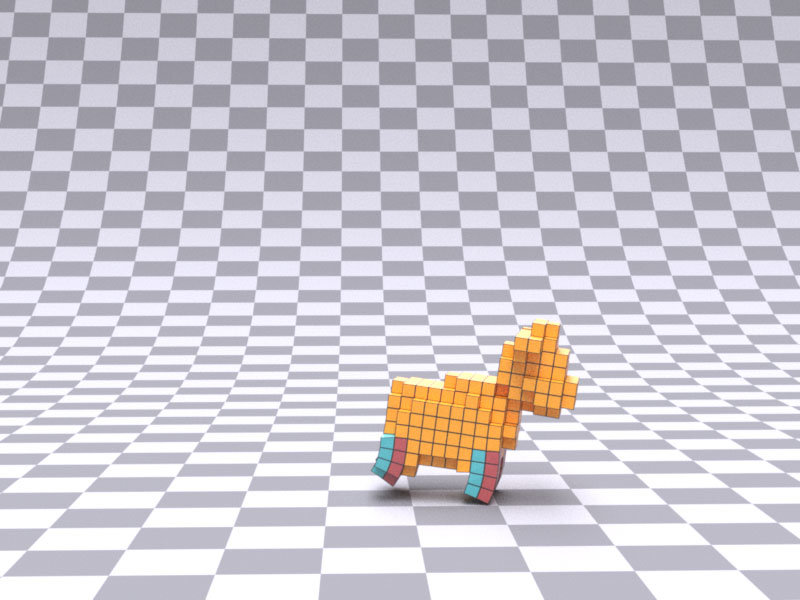}\\
    \vspace{-10.7em}
    \hspace{50em}
    \begin{minipage}[t]{0.1\textwidth}
    \includegraphics[width=\textwidth]{figure/colormap.JPG}
    \end{minipage}
    \vspace{8.5em}
  \caption{\textbf{Trajectory optimization} The motion sequences of the ``Cow'' example with sinusoidal control signals whose 3 parameters are to be optimized. The goal is to maximize the walking distance of the cow. Top: the motion sequence with a random sinusoidal wave of actions. Bottom: the motion sequence after optimization with DiffPD.}
  \label{fig:cow}
  \Description{Cow.}
\end{figure*}

In order to demonstrate the applicability of our system's differentiability to solving complex trajectory optimization tasks, we apply our simulator to three locomotion tasks: a ``Torus'', a ``Quadruped'', and a ``Cow''. All three robots are equipped with muscle fibers whose sequences of actions are to be optimized, and the goal for all three robots is to walk forward without losing balance or drifting sideways. All examples use the complementarity-based contact model in Sec.~\ref{sec:contact:compl}.

\paragraph{Torus}
In our first trajectory optimization example, a torus is tasked with rolling forward as far as possible in $1.6$ seconds, simulated as $400$ steps of $4$ milliseconds in length (Fig.~\ref{fig:torus}).  To achieve this, we set the objective to be the negation of the robot's center of mass at the final step of its trajectory.  Eight muscle tendons are routed circumferentially along the center of the torus, combined, creating a circle that can be actuated along any of eight segments.  The optimization variables are the actuation of the each muscle at each of $20$ linearly spaced knot points, and the actual action sequences are generated by linearly interpolating variables at these knot points.  Since there are $8$ muscles, this results in $160$ decision variables overall. We use a convergence threshold of 1e-6 as indicated by the evaluation experiment in Sec.~\ref{sec:evaluation:implicit}.

The major challenge in optimizing the sequence of actions of this rolling torus lies in the fact that it constantly breaks and reestablishes contact with the ground. When running L-BFGS on this example, we noticed more local minima than previous examples and L-BFGS often terminated prematurely without making significant progress. To alleviate this issue, we randomly sampled 16 initial solutions and selected the best among them to initialize L-BFGS optimization, which eventually learned a peristaltic contraction pattern that allows it to start rolling forward and make considerable forward progress (Fig.~\ref{fig:torus}); further, DiffPD provides a $6\times$ speedup using $8$ threads compared to the Newton's method.

\begin{figure*}[htb]
  \centering
  \includegraphics[width=\linewidth]{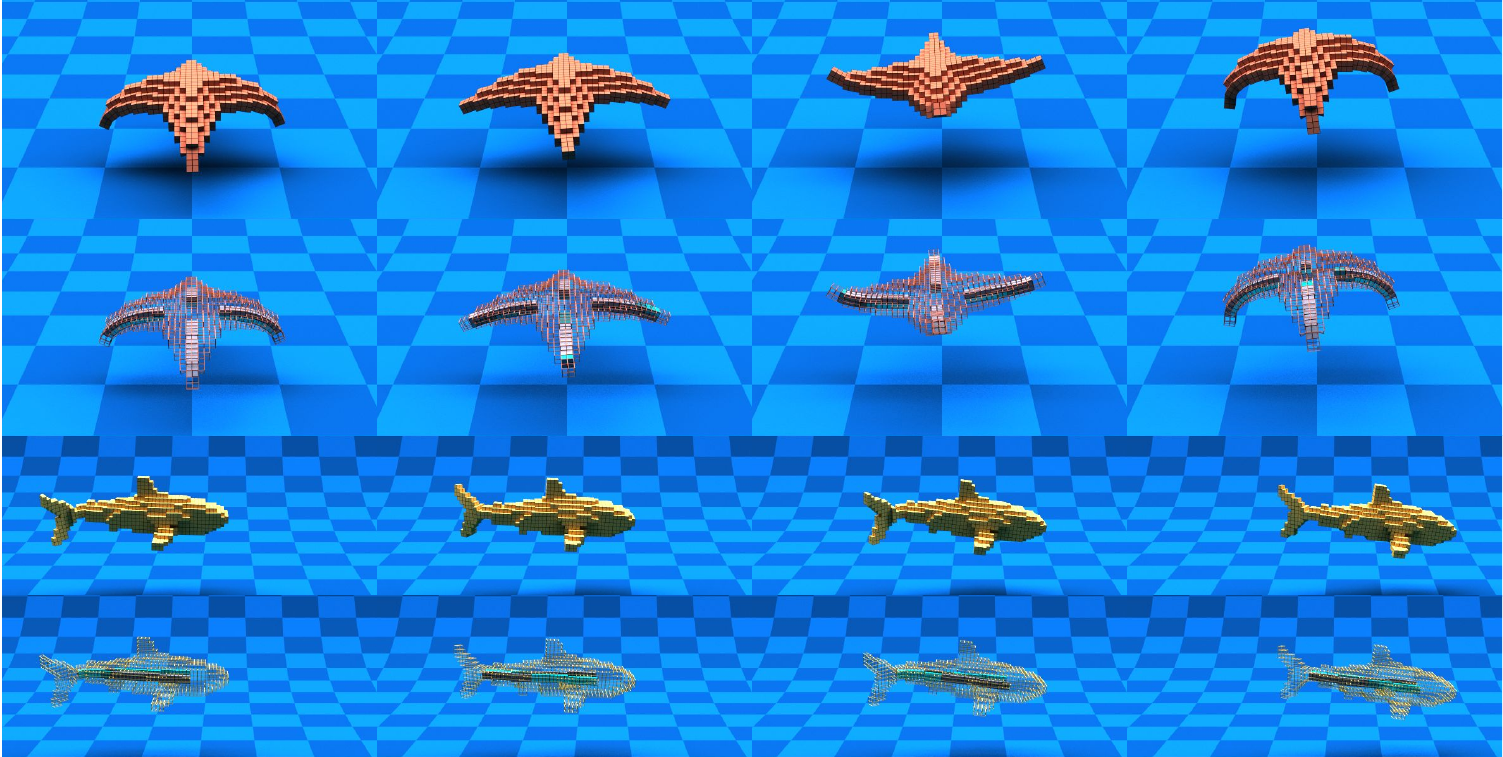}
  \vspace{-2em}
  \caption{\textbf{Control optimization} Motion sequences from the optimized closed-loop, neural network controllers of ``Starfish'' (first row) and ``Shark'' (third row), with the corresponding muscle fibers plotted in the second and fourth rows. The goal is to optimize a controller so that the marine creatures can rise (``Starfish'') or swim forward (``Shark''). The gray and cyan colors on the surface of the muscle fibers indicate the magnitude of the actuation, with gray being zero actuation and blue the maximum contraction.}
  \label{fig:fish}
  \Description{Fish.}
\end{figure*}

\paragraph{Quadruped}
For our second trajectory optimization example, a rectangular quadruped is tasked with moving forward as far as possible in $1$ second.  The same performance objective is applied here as in the ``Torus'' example, however in this example a simpler control scheme is implemented. This robot has eight muscles, routed vertically along the front and back face of each leg, allowing the legs to bend forward or backward. For each leg, the front and back muscle groups are paired antagonistically, however they are allowed a different maximum actuation strength -- a parameter to be optimized. Finally, the entire quadruped is provided a single sinusoidal control signal, whose frequency is to be optimized, that actuates each leg synchronously. These front and back actuation strengths, combined with the frequency of the input  signal, provide $3$ parameters to be optimized.  After optimization, the quadruped was able to walk forward several body lengths (Fig.~\ref{fig:quadruped}). In terms of the speedup, DiffPD accelerates loss and gradient evaluation by a factor of $4$ compared with the Newton's method.

\paragraph{Cow}
For our third and final trajectory optimization example, a cow quadruped based off Spot~\cite{spot} is tasked with walking forward as far as possible in $0.6$ seconds (Fig.~\ref{fig:cow}).  This is a particularly difficult task, as Spot's oversized head makes it front-heavy, and prone to falling forward.  In order to compensate, we regularized the objective to promote a more upright gait, adding an additional $-0.3$ times the center of mass in $z$ to regularize the forward objective.  Spot uses the same controller and muscle arrangement as the ``Quadruped'' example, and a convergence threshold of 1e-6 is used during optimization. Similar to previous examples, the cow optimizes to walk forward and DiffPD provides a 5 times speedup compared to the Newton's method.

\paragraph{Discussion}  Locomotion tasks generally involve significant contact, which limits the speedups (4-6 times in the examples above) DiffPD is able to achieve compared with contact-free problems. Given the complexity of planning the motion of walking robots with contacts (optimizing each of the three examples above took hours to converge with the Newton's method), a 4-6 times speedup is still favorable. It is also worth noting that for the ``Quadruped'' and ``Cow'' examples, optimization with the Newton's method led to solutions significantly different from DiffPD, as indicated by the final loss reported in Table~\ref{tab:performance}. We believe this is mostly due to the algorithmic difference between the Newton's method and DiffPD: as discussed in~\citet{liu2017quasi} and Sec.~\ref{sec:diff_pd}, DiffPD is essentially running the quasi-Newton method (as opposed to the Newton's method) to minimize the objective in Eqn. (\ref{eq:opt}) which is typically not convex. Therefore, multiple critical points may exist especially when contacts are involved. For the three locomotion tasks in this section, it is possible that DiffPD and two Newton's methods each explored different critical points individually and led to different solutions.

\subsection{Closed-Loop Control}\label{sec:app:control}
Inspired by~\citet{min2019softcon}, we consider designing a closed-loop neural network controller for two marine creatures: ``Starfish'' and ``Shark'' (Fig.~\ref{fig:fish}). For each example, we specify muscle fibers as internal actuators similar to~\citet{min2019softcon} in the arms of the starfish and the caudal fin of the shark. We manually place velocity sensors on the body of each example serving as the input to the neural network controller. The goal of these examples is to optimize a swimming controller so that each fish can advance without drifting sideways. To achieve this, we define the loss function as a weighted sum of forward velocities and linear velocities at each time step. In terms of the neural network design, we choose a 3-layer multilayer perceptron network with 64 neurons in each layer (30788 parameters in ``Starfish'' and 22529 parameters in ``Shark''). We use the hyperbolic tangent function as the activation function in the neural network. Unlike prior examples for which L-BFGS is used for optimization, we follow the common practice of using gradient descent with Adam~\cite{kingma2015adam} to optimize the neural network parameters. During optimization, we use a convergence threshold of 1e-3 in DiffPD and the Newton's method. Table~\ref{tab:performance} provides the final loss after optimization, and we observe a speedup of 8-19 times for the two examples respectively.

\paragraph{Comparisons with reinforcement learning} We compare our gradient-based optimization method to PPO \cite{schulman2017proximal}, a state-of-the-art reinforcement learning algorithm. In particular, we use the forward simulation of DiffPD as the simulation environment for PPO. For a fair comparison, we construct and initialize the network for both DiffPD and PPO with the same random seed. We also implement code-level optimization techniques as suggested in~\citet{engstrom2019implementation} and tune PPO hyperparameters towards its best performance. Please refer to our supplemental material for more implementation details.

\begin{figure}
    \centering
    \includegraphics[width=0.9\linewidth]{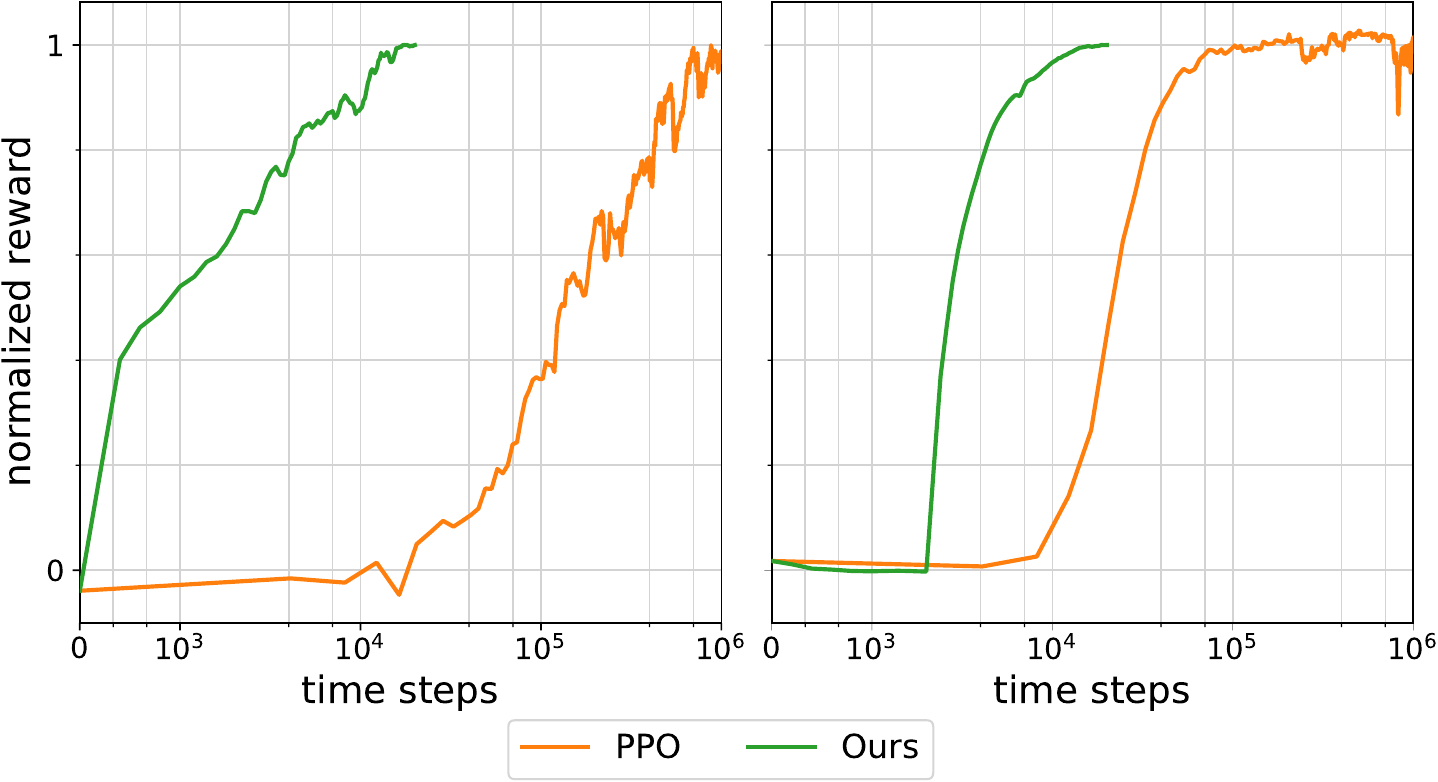}
    \caption{The optimization progress of Adam plus DiffPD (green) and PPO (orange) for ``Starfish'' (left) and ``Shark'' (right). Note that the axis of time steps spent during training or optimization is plotted on a logarithmic scale.}
    \label{fig:marine_drl}
\end{figure}

When comparing the performance of a gradient-free algorithm like PPO to gradient-based algorithms like Adam or L-BFGS, we expect gradient-based optimization to be more sampling efficiency than PPO as gradients expose more information about the soft body dynamics that are not accessible to PPO. Note that this does not ensure gradient-based methods are always faster than PPO when measured by their wall-clock time, because each sample in a gradient-based method requires additional gradient computation time. Furthermore, the sampling scheme in PPO is massively parallelizable. We report the optimization progress of PPO and our method in Fig.~\ref{fig:marine_drl}. Note that unlike other examples, we follow the convention in reinforcement learning of maximizing a reward as opposed to minimizing a loss. In particular, a zero reward indicates the average performance of randomly selected unoptimized neural networks, and a unit reward is the result from DiffPD after optimization. We conclude from Fig.~\ref{fig:marine_drl} that Adam and DiffPD achieves comparable results to PPO but is more sampling efficient by one or two orders of magnitude. Regarding the wall-clock time, we observe a speedup of 9-11 times for both examples respectively, although each sample in DiffPD is more expensive due to the gradient computation.

\subsection{A Real-to-Sim Experiment}\label{sec:app:sim2real}
We end this section with a real-to-sim experiment ``Tennis balls''. In this example, we capture a video clip of two colliding tennis balls on a flat terrain, from which we aim to estimate the camera information, the initial position and velocity of each ball, and the parameters in the contact model. We model each ball in simulation using a mesh of sphere with 320 tetrahedrons and 489 DoFs (Table~\ref{tab:setting}). We model the ball-ground contact with the complementarity-based method and use the following penalty-based model to compute the ball-ball contact: when the two balls are in contact, we add a restitution force computed as the product of a stiffness parameter to be optimized and the difference between the ball diameter and the actual distance between the two ball centers. Essentially, the restitution force can be treated as a spring model with a rest length equal to the ball diameter. Additionally, we add a frictional force whose direction is orthogonal to the restitution force and whose magnitude is controlled by a frictional coefficient to be optimized.

To define a loss function that measures the discrepancy between the simulated and actual motions of the two balls, we first extract the pixel location of two balls' centers in each frame of the video clip. Next, we compute in simulation the position of each ball and project them to the same image space through a pinhole camera model. We define the loss function as the difference between the pixel locations of the balls in simulation and from the video clip. By minimizing this loss, we get our estimation of the camera information, the initial state of each ball, and the parameters in the contact model.

\begin{figure*}[ht]
  \centering
    \includegraphics[trim=180 150 200 80,clip,width=0.33\linewidth]{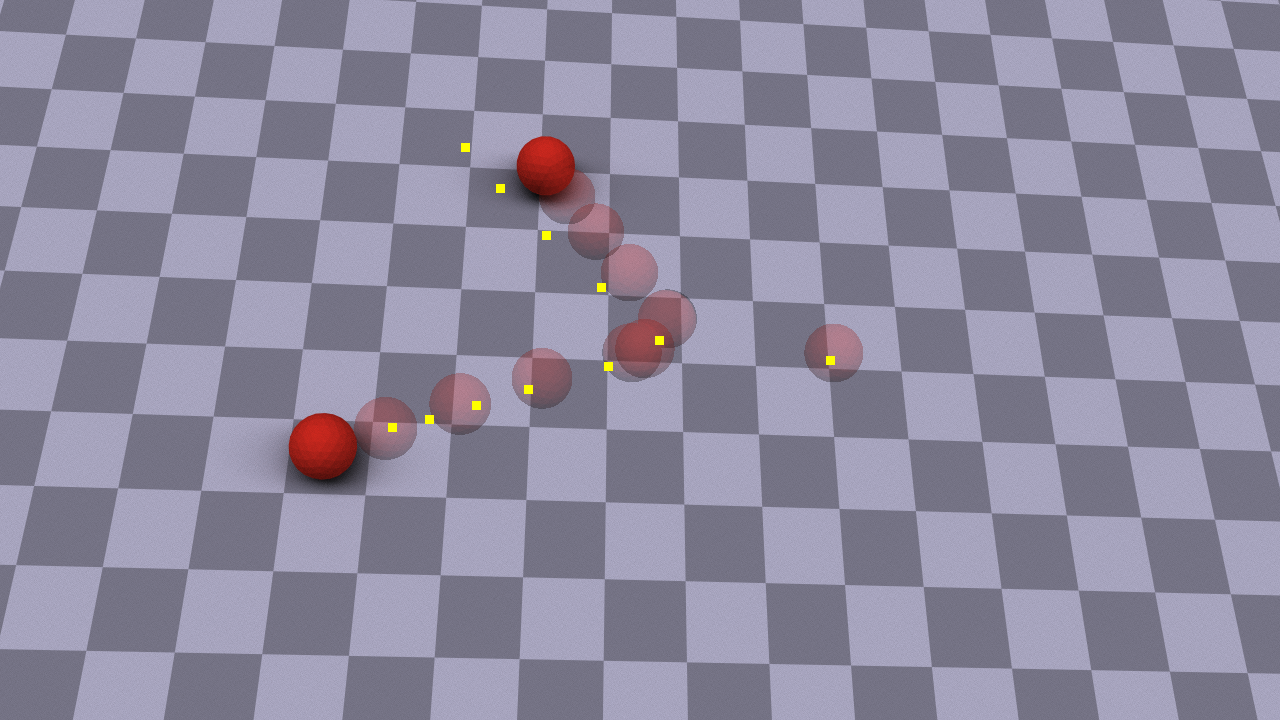}
    \includegraphics[trim=180 150 200 80,clip,width=0.33\linewidth]{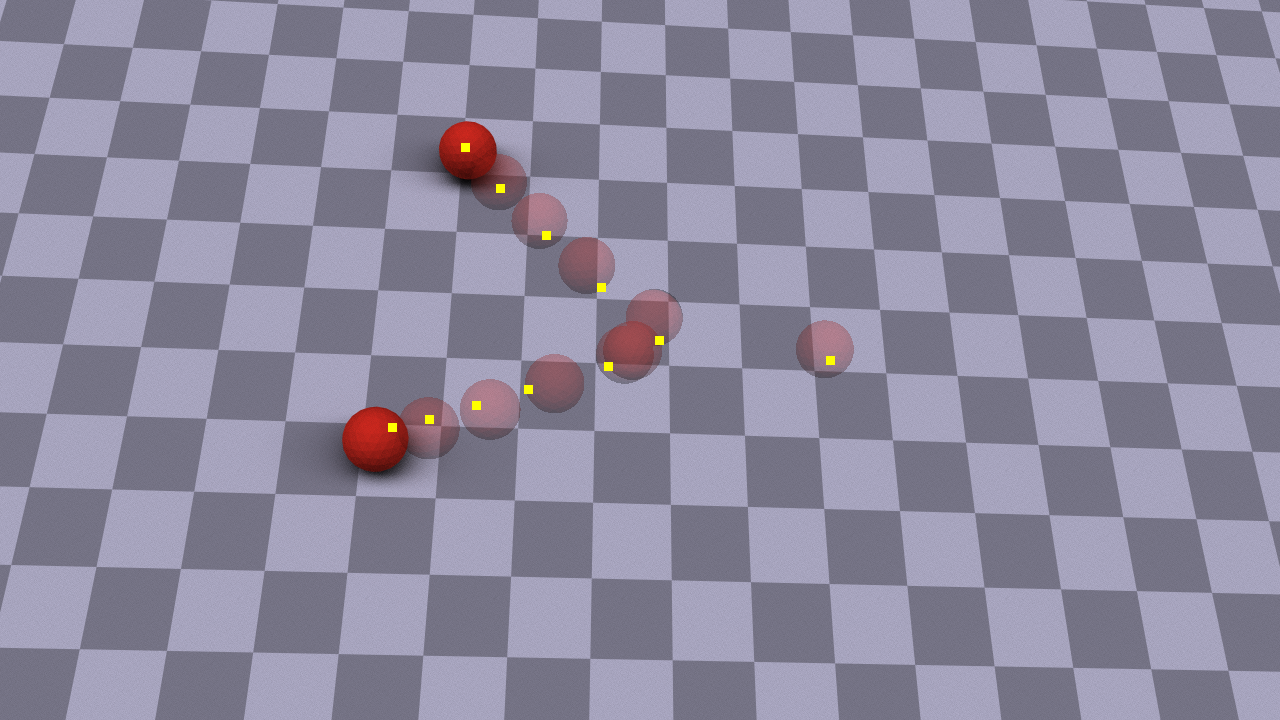}
    \includegraphics[trim=180 150 200 80,clip,width=0.33\linewidth]{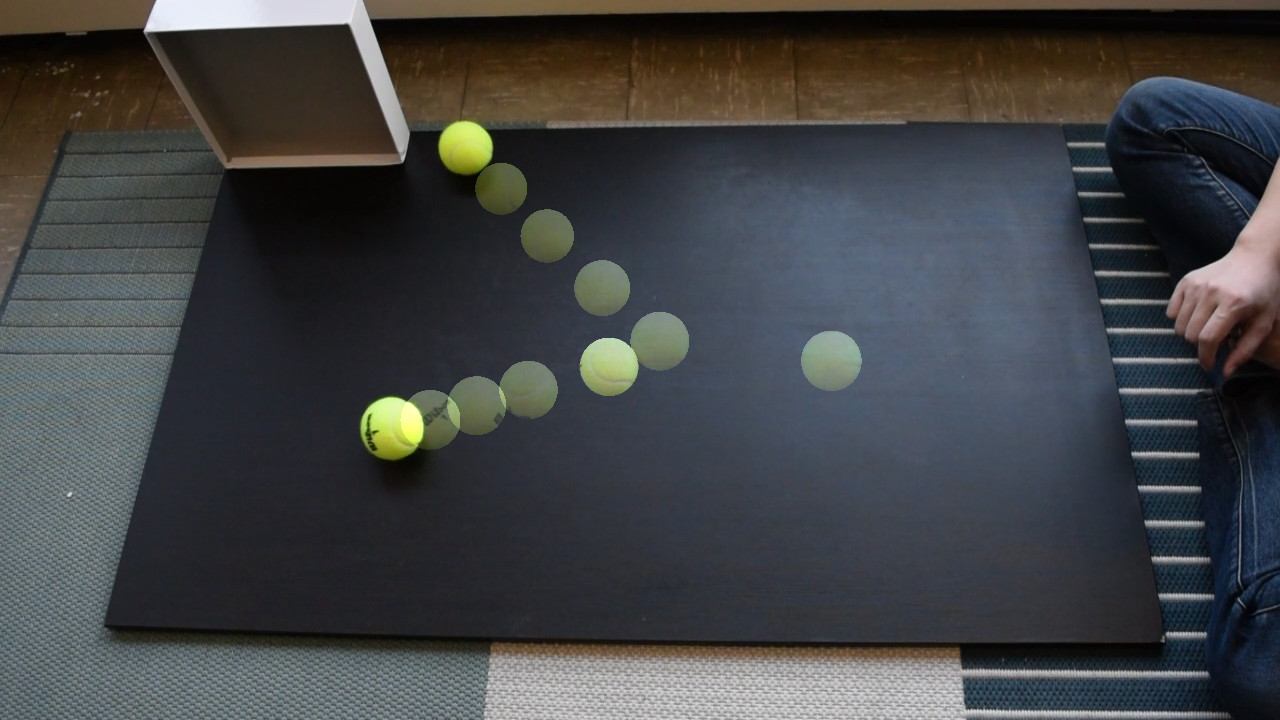}
    \vspace{-1.8em}  
  \caption{\textbf{A Real-to-Sim Experiment} Motion sequences of the ``Tennis balls'' example before (left) and after optimization (middle). The corresponding video clip is shown on the right. Transparent balls indicate the balls' intermediate locations. To visualize the difference between motion sequences in simulation and reality, we use yellow squares in the rendered images (left and middle) to denote the corresponding pixel locations of the balls' centers from the video clip.}
  \label{fig:sim2real}
  \Description{sim2real}
\end{figure*}

We summarize our optimization results in Table~\ref{tab:performance} and Fig.~\ref{fig:sim2real}. We randomly sample multiple sets of parameters and pick those with the smallest loss as the initial guess to our optimization (Fig.~\ref{fig:sim2real} left), which shows motion sequences similar to those in the video clip but still with substantial visual differences. The optimization process refines our estimation of the parameters and manages to further suppress the loss and mimics the motions in the video more closely (Fig.~\ref{fig:sim2real} middle). The results can be further improved if we take into account camera lens distortion or replace the penalty-based collision model between two balls with a more accurate one, which we leave as future work.

\section{Limitations and Future Work}\label{sec:limitations}

\begin{figure}
    \centering
    \includegraphics[width=\columnwidth]{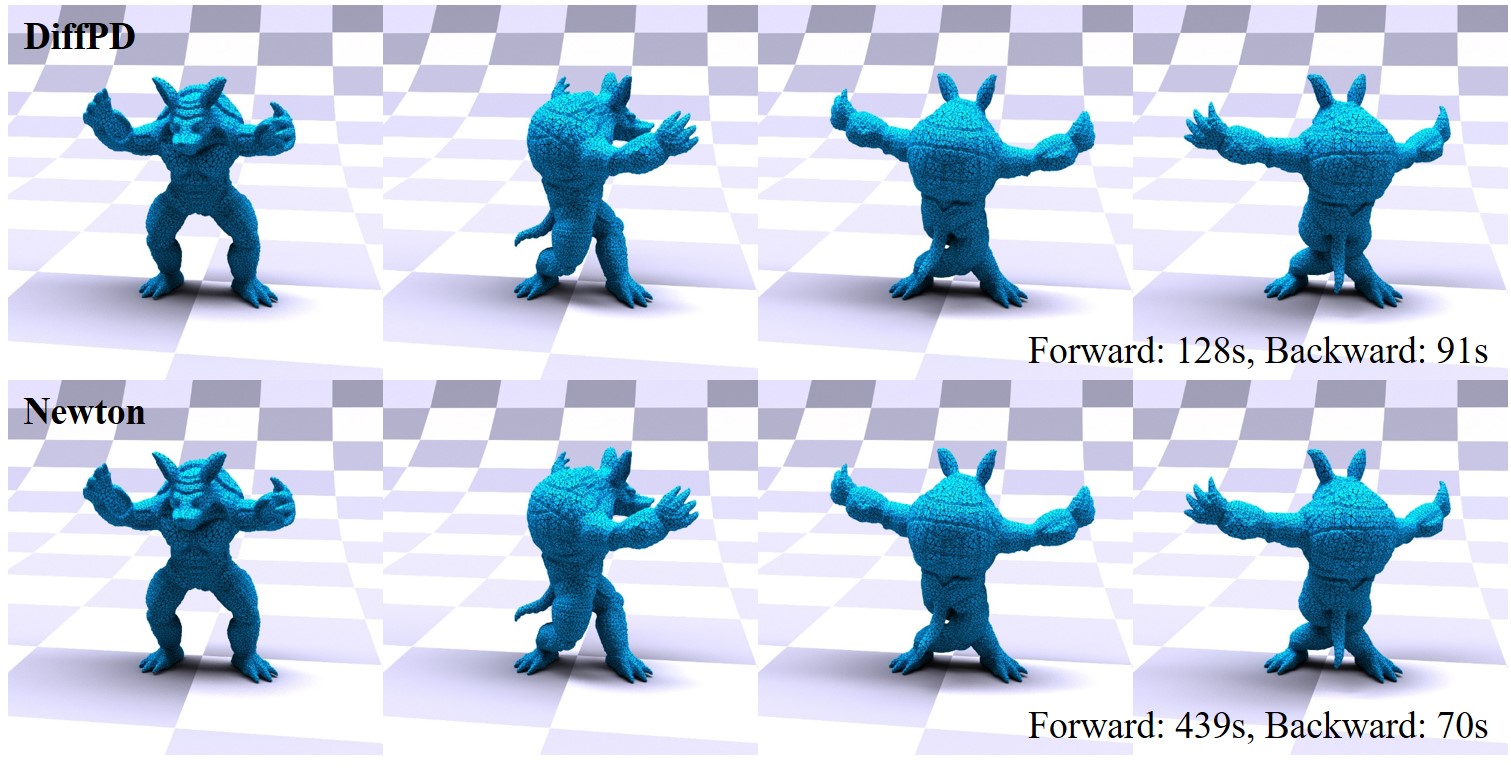}
    \caption{Twisting Armadillo (a 44337-DoF tetrahedral mesh) with the Neohookean material model for 1 second at 30 frames per second. We use DiffPD (top) and Newton's method (bottom) to compute the forward simulation and backpropagation. Left to right: the intermediate state of Armadillo at 0, 10, 20, and 30 frames. The visually identical motions from the top and bottom rows confirm the correctness of DiffPD's implementation of Neohookean material model. We report the time cost of DiffPD and Newton's method at the lower right corner of the images and no longer witness a speedup from DiffPD in backpropagation over a direct solver in Newton's method.}
    \label{fig:armadillo}
\end{figure}

Differentiable soft-body simulation with proper contact handling is a challenging problem due to its large number of DoFs and complexity in resolving contact forces. We believe one ambitious direction along this line of research is to provide a physically realistic simulator that can facilitate the design and control optimization of \emph{real} soft robots. To close the sim-to-real gap, some nontrivial but rewarding enhancements need to be integrated into our current implementation. First and foremost, similar to other PD papers, a major limitation in DiffPD is its assumption on the energy function of the material model. Even though technical solutions to supporting general hyperelastic materials in PD exist~\cite{liu2017quasi}, it turns out that supporting such materials in DiffPD is not straightforward. This is because the derivation in Sec.~\ref{sec:diff_pd} starts to fall apart from Eqns. (\ref{eq:energy_grad}) and (\ref{eq:energy_hessian}) when hyperelastic material models are used, forcing DiffPD to reassemble the Hessian matrix $\nabla^2 E$ in each time step during backpropagation. Although we can still apply the iterative solver from Sec.~\ref{sec:diff_pd} in this case, we no longer observe a speedup over a direct solver (Fig.~\ref{fig:armadillo}). Therefore, we switch to the direct solver for backpropagation in DiffPD when hyperelastic materials are used and leave speeding it up as future work.

Second, our contact methods do not fully resolve differentiable, complementarity-based contact and friction. Due to the focus of this paper, our choice of the contact model was intentionally biased towards ensuring differentiability and compatibility with PD. It would be more accurate and useful to upgrade the contact models for both static and sliding frictional forces~\cite{ly2020projective} or to apply a more realistic contact model~\cite{Li2020IPC} while maintaining its efficiency and differentiability in PD.

A third direction that is worth exploring is to improve the scalability of our algorithm. Currently, the largest example in this paper contains thousands of elements and tens of thousands of DoFs. It would be desirable to scale problems up by at least one or two orders of magnitude in order to explore the effects of more complex geometry. 
This would obviously be computationally expensive; it would therefore be interesting to explore possible GPU implementations of DiffPD method to unlock large-scale applications.

Fourth, although DiffPD is substantially faster than standard Newton's method when assumptions in PD hold, the speedup is less significant for locomotion tasks (4-6 times in our examples). We suspect it is the inclusion of contact that slows down DiffPD both in forward simulation and in backpropagation. Therefore, a more comprehensive analysis on the assumption of sparse contact in Sec.~\ref{sec:contact} would possibly reveal the source of inefficiency. Specifically, removing such an assumption would be much desired to unlock contact-rich applications, e.g., cloth simulation or manipulation.

Finally, there is room for improving the optimization strategies that can better leverage the benefits of gradients. In all our examples, we couple gradient information with gradient-based continuous optimization methods. Being inherently local, such methods inevitably suffer from terminating at local minima prematurely especially when the loss function has a non-convex landscape. It is worth exploring the field of global optimization methods or even combining ideas from gradient-free strategies, e.g., genetic algorithms or reinforcement learning, to present a more robust global optimization algorithm specialized for differentiable simulation.

\begin{acks}
We thank Desai Chen, David I.W. Levin, Bo Zhu, and Eftychios Sifakis for their feedback and suggestions on this paper. The duck and cow mesh models in Figs. \ref{fig:duck_slope}, \ref{fig:duck_slide}, and \ref{fig:cow} are obtained from Keenan Crane's 3D model repository under the CC0 1.0 Universal license. This work is sponsored by Defense Advanced Research Projects Agency (DARPA) under grant No. FA8750-20-C-0075, Intelligence Advanced Research Projects Activity (IARPA) under grant 2019-19020100001, and National Science Foundation (NSF) Award 2106962: Computational Design of Complex Fluidic Systems.
\end{acks}

\bibliographystyle{ACM-Reference-Format}
\bibliography{reference}

\end{document}

\title{DiffPD: Differentiable Projective Dynamics (Supplemental Material)}

\author{Tao Du}
\email{taodu@csail.mit.edu}
\orcid{0000-0001-7337-7667}

\author{Kui Wu}
\email{kuiwu@csail.mit.edu}

\author{Pingchuan Ma}
\email{pcma@csail.mit.edu}

\author{Sebastien Wah}
\email{sebwah@mit.edu}

\author{Andrew Spielberg}
\email{aespielberg@csail.mit.edu}

\affiliation{%
  \institution{MIT CSAIL}
}

\author{Daniela Rus}
\email{rus@csail.mit.edu}

\author{Wojciech Matusik}
\email{wojciech@csail.mit.edu}
\affiliation{%
    \institution{MIT CSAIL}
}

\renewcommand{\shortauthors}{Du et al.}

\begin{CCSXML}
<ccs2012>
   <concept>
       <concept_id>10010147.10010371.10010352.10010379</concept_id>
       <concept_desc>Computing methodologies~Physical simulation</concept_desc>
       <concept_significance>500</concept_significance>
       </concept>
 </ccs2012>
\end{CCSXML}

\ccsdesc[500]{Computing methodologies~Physical simulation}

\keywords{Projective Dynamics, differentiable simulation}

\maketitle

\section{Energy Definitions}

We now give the definitions of all quadratic energy used in our examples. For each energy definition, we briefly comment on how their gradients can be computed during backpropagation.

\paragraph{Background elasticity} We use the same elastic energy as defined in~\citet{min2019softcon}, which consists of a corotated term and a volume-preserving term. The energy function of the corotated term is defined as follows:
\begin{align}
    E(\mathbf{x}) = & \; \frac{w}{2}\|\mathbf{F}-\mathbf{R}\|_F^2.
\end{align}
Here, $\mathbf{F}$ is the deformation gradient evaluated at $\mathbf{x}$ and $\mathbf{R}$ the rotation matrix closest to $\mathbf{F}$. Comparing it to Eqn. (16) in the main paper, we can see $\mathbf{G}$ is defined as the linear operator that computes the deformation gradient and $\mathcal{M}$ is $SO(3)$, the set of all 3D rotation matrices.

It is worth mentioning that we deliberately skipped the details of discretization scheme for brevity. In our implementation, it is the energy density function $\Psi$ that takes the form $\|\mathbf{F}-\mathbf{R}\|_F^2$, and $E$ is defined as an integral over each element, which, after numerical discretization, becomes the sum of energy density functions evaluated at Gaussian quadratures. Interested readers can refer to~\citet{sifakis2012fem} for more details on discretization.

Regarding backpropagation, it is only necessarily to show how $\frac{\partial\mathbf{R}}{\partial\mathbf{F}}$ is computed. Recall that $\mathbf{R}$ is computed from the polar decomposition of $\mathbf{F}$~\cite{min2019softcon}, deriving $\frac{\partial\mathbf{R}}{\partial\mathbf{F}}$ is nothing more than deriving the gradients of the rotation matrix in the polar decomposition, which can be found in~\citet{hu2019chainqueen}.

Similarly to the corotated energy, the volume-preserving energy is defined as follows:
\begin{align}
    E(\mathbf{x})=\frac{w}{2}\|\mathbf{F}-\mathbf{D}\|_F^2,
\end{align}
where $\mathbf{D}$ is the projection of $\mathbf{F}$ onto the manifold $\mathcal{M}=\{|\mathbf{D}|=1|\mathbf{D}\in\mathcal{R}^{3\times3}\}$. During forward simulation, $\mathbf{D}$ can be obtained from solving an optimization problem~\cite{bouaziz2014projective}, which we restate below for the sake of explaining its gradient derivation later: let $\mathbf{F}=\mathbf{U}\mathbf{\Sigma}\mathbf{V}^\top$ be the SVD decomposition of $\mathbf{F}$ and consider the following problem:
\begin{align}
    \min_{\mathbf{d}\in\mathcal{R}^3} & \; \|\mathbf{d}\|_2^2 \\
    \text{s.t.} & \; \Pi_i (\mathbf{d}_{i}+\mathbf{\Sigma}_{ii})=1.
\end{align}
Once we obtain the optimal $\mathbf{d}^*$ from the optimization problem above, the optimal $\mathbf{D}$ is computed as $\mathbf{D}=\mathbf{U}\mathbf{\Sigma}_{\mathbf{D}}\mathbf{V}^\top$ where $\mathbf{\Sigma}_{\mathbf{D}}$ is a diagonal matrix whose $i$-th diagonal entry is $\mathbf{d}_i+\mathbf{\Sigma}_{ii}$.

To derive its gradients during backpropagation, it is sufficient to show how to compute $\frac{\partial\mathbf{d}^*}{\partial\mathbf{\Sigma}_{ii}}$, i.e., if we perturbed the singular values of $\mathbf{F}$, how would the optimal $\mathbf{d}^*$ change? Recall that $\mathbf{d}^*$ and $\mathbf{\Sigma}_{ii}$ must satisfy the following Karush–Kuhn–Tucker (KKT) condition:
\begin{align}
    \mathbf{d}^* + \lambda^* \nabla c(\mathbf{d}^*) = & \; 0, \\
    c(\mathbf{d}^*) = & \; 0,
\end{align}
where $\lambda^*\in\mathcal{R}$ is the Lagrangian multiplier and $c(\mathbf{d}^*)$ the equality condition:
\begin{align}
    c(\mathbf{d}^*)=\Pi_i(\mathbf{d}^*_i+\mathbf{\Sigma}_{ii})-1.
\end{align}
Differentiating the KKT condition reveals the follow equation that $\mathbf{\delta d}^*$ and $\mathbf{\delta \Sigma}$ needs to satisfy:
\begin{equation}\label{eq:vol_kkt}
    \begin{pmatrix}
    \mathbf{I} + \lambda^* \nabla^2 c(\mathbf{d}^*) & \nabla c(\mathbf{d}^*) \\
    [\nabla c(\mathbf{d}^*)]^\top & 0
    \end{pmatrix}
    \begin{pmatrix}
    \mathbf{\delta d^*} \\
    \delta \lambda
    \end{pmatrix} =
    \begin{pmatrix}
    -\lambda^*\nabla^2c(\mathbf{d}^*)\mathbf{\delta\Sigma} \\
    -[\nabla c(\mathbf{d}^*)]^\top\mathbf{\delta \Sigma}
    \end{pmatrix}.
\end{equation}
With such an equation at hand, we can solve $\frac{\partial \mathbf{d}^*}{\partial \mathbf{\Sigma}_{ii}}$ by letting $\mathbf{\delta\Sigma}$ be a one-hot vector and solving $\mathbf{\delta d}^*$ from Eqn. (\ref{eq:vol_kkt}).

\paragraph{Muscle fibers} We again use the same muscle model described in~\citet{min2019softcon}:
\begin{align}
    E(\mathbf{x})=\frac{w}{2}\|(1-r)\mathbf{F}{\mathbf{m}}\|_2^2,
\end{align}
where $\mathbf{m}$ is a predefined muscle fiber direction per element and $r\in\mathcal{R}_{+}$ the actuation signal. From the viewpoint of PD, this energy projects $\mathbf{Fm}$ onto a spherical manifold $\mathcal{M}=\{\mathbf{p}\in\mathcal{R}^3|\|\mathbf{p}\|_2=r\}$, and its gradients with respect to both $\mathbf{x}$ and $r$ can be easily derived by applying the chain rule.
\section{Experiment Details}

In this section, we provide more information about the experiments presented in Sec. 7 of the main paper. For each example, we present a detailed explanation of the experimental setup as well as the progress of optimization.

\subsection{System Identification}

\begin{figure}[htb]
    \centering
    \includegraphics[width=\linewidth]{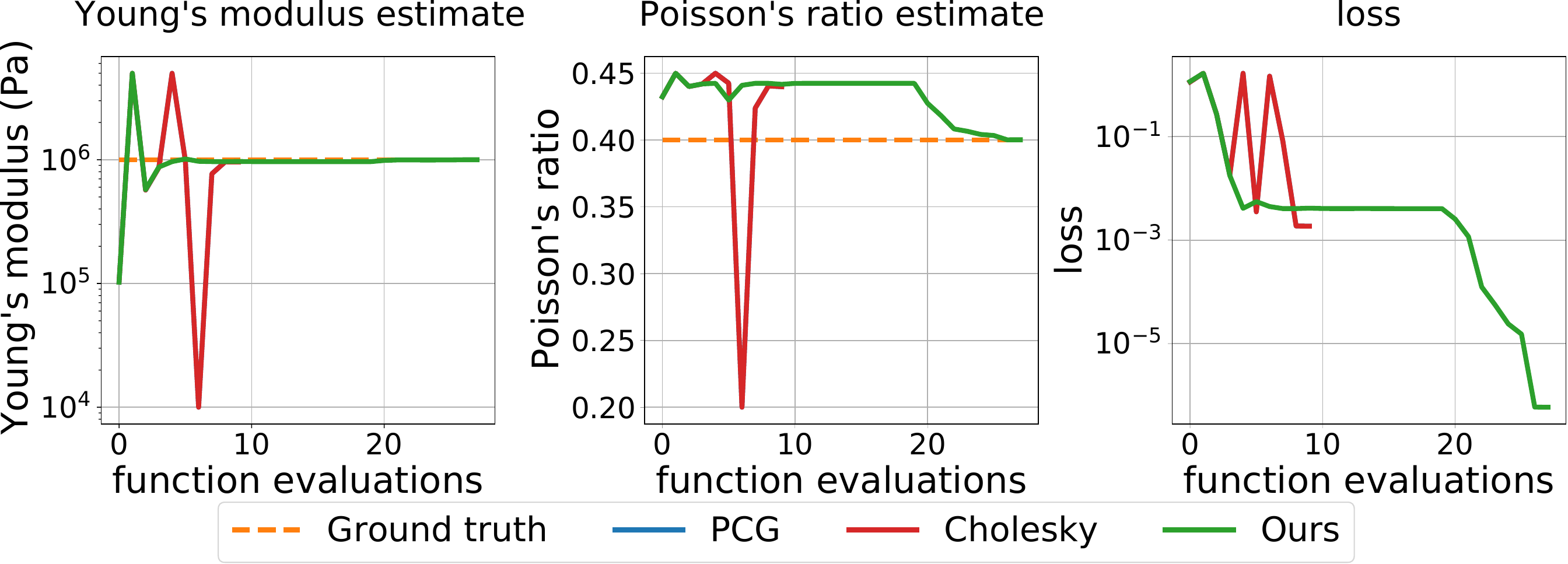}
    \caption{The progress of running L-BFGS optimization to solve the ``Plant'' example. The horizontal axis shows the number of function evaluations used, and the three vertical axes plot the intermediate Young's modulus (left), Poisson's ratio (middle), and loss (right).}
    \label{fig:supp:plant}
\end{figure}

\paragraph{Plant} We first deform the plant by pulling it horizontally with a constant force on the $xy$ plane, generating the first frame for simulation. During simulation, the external constant force is removed and the plant swings due to its deformation. This example has two decision variables: Young's modulus (range: 1e4 to 5e6) and Poisson's ratio (range: 0.2 to 0.45) with the ground truth values being 1e6 and 0.4, respectively. To make the optimization less ill-conditioned, we optimize the logarithm of the decision variables. Fig.~\ref{fig:supp:plant} displays the optimization progress. Note that we choose to use the number of function evaluations as the horizontal axis: compared with showing major L-BFGS optimization iterations, showing function evaluations flattens out the line search stage inside each optimization iteration and reflects the wall-clock time cost more faithfully.

\begin{figure}[htb]
    \centering
    \includegraphics[width=\linewidth]{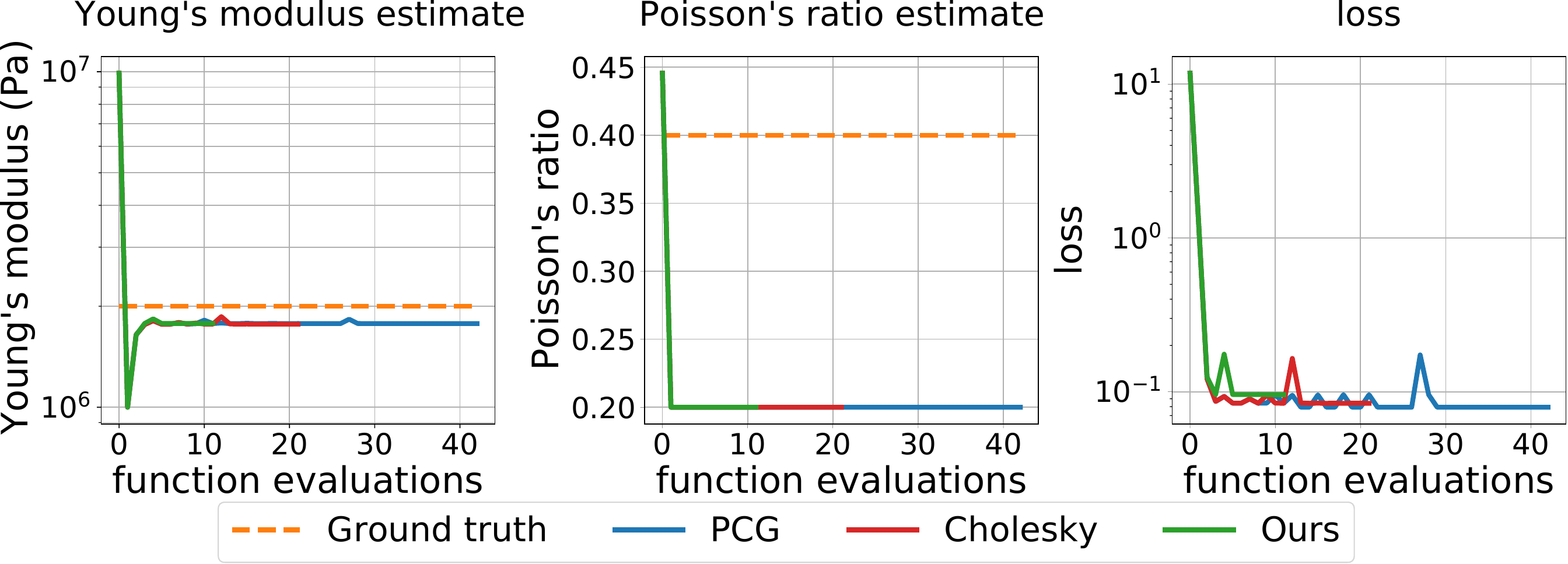}
    \caption{The progress of running L-BFGS optimization to solve the ``Bouncing ball'' example. The horizontal axis shows the number of function evaluations used, and the three vertical axes plot the intermediate Young's modulus (left), Poisson's ratio (middle), and loss (right).}
    \label{fig:supp:bouncing_ball}
\end{figure}

\paragraph{Bouncing ball} In this example, we initially put the ball 7 centimeters above the ground and set its initial velocity to be 1.5, 0.5, and -1 meters per second along the $x$, $y$, and $z$ axis, respectively. Gravity is along the negative $z$ direction. Similar to the ``Plant'' example, the decision variables are the logarithm of Young's modulus and Poisson's ratio. The lower and upper bounds are 1e6 and 1e7 for Young's modulus and 0.2 and 0.45 for Poisson's ratio, and the ground truth values are 2e6 and 0.4, respectively. Fig.~\ref{fig:supp:bouncing_ball} shows the progress of optimization. It is worth noting that while the optimized Poisson's ratio is almost half that of the ground truth value, the resulting motion of the ball is very similar to the ground truth. Since the loss function is defined on the motion and not the material parameters, such an observation reveals that there could exist multiple combinations of material parameters that can result in very similar motion sequences.

\subsection{Initial Condition Optimization}

\paragraph{Bunny} In this example, we optimize 9 decision variables that characterize the initial orientation (represented as Euler angles), position, and velocity of the bunny. We constrain the bounds of the initial position so that it can only be perturbed mildly. This is to avoid a trivial solution that directly puts the initial position at the target position without the need of bouncing it off the ground. Similarly, the bounds of the initial velocity are designed so that it always points downwards, ensuring at least one bounce of the bunny.

\begin{figure}[htb]
    \centering
    \includegraphics[width=\linewidth]{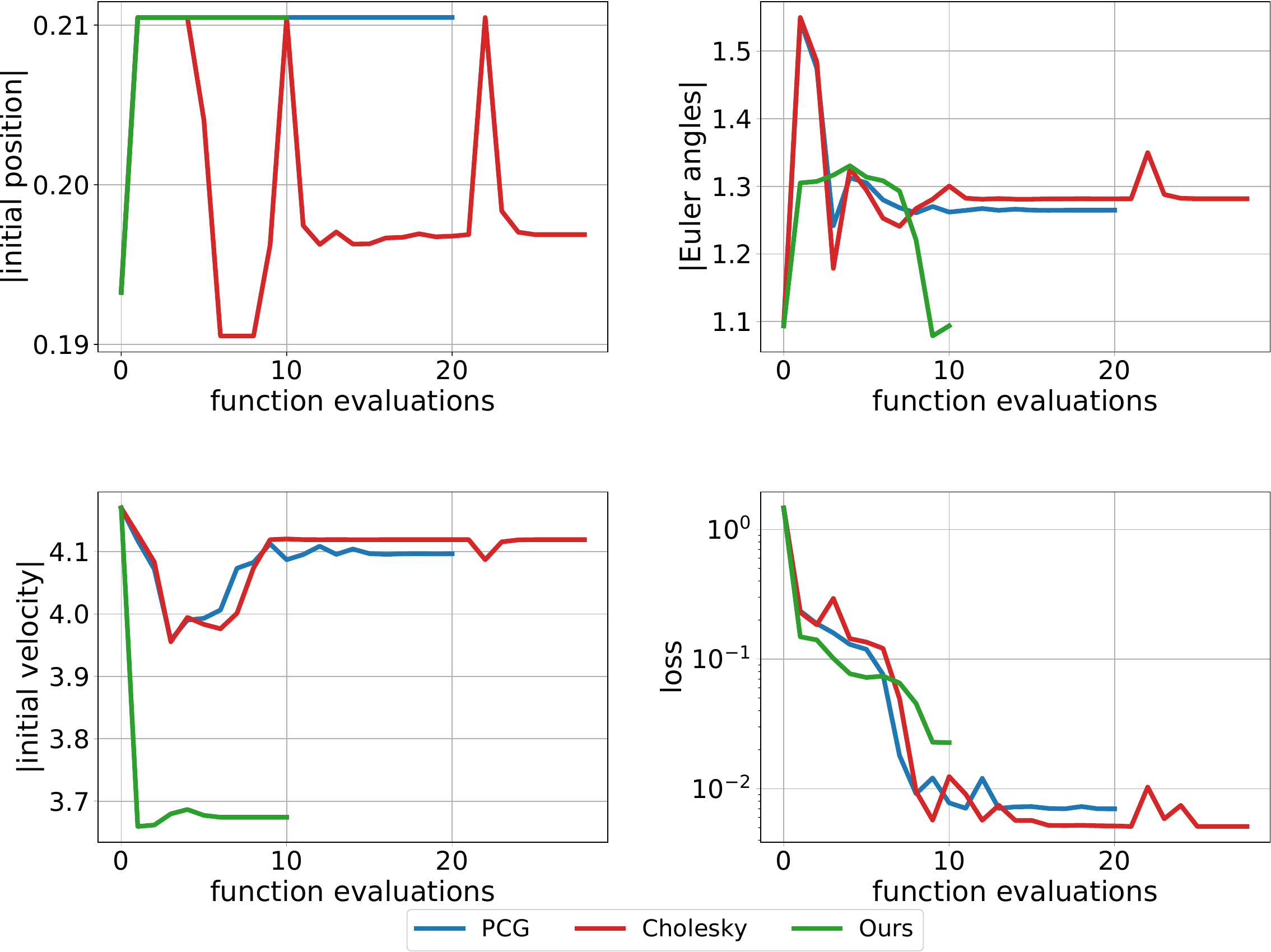}
    \caption{The progress of running L-BFGS optimization to solve the ``Bunny'' example. The horizontal axis shows the number of function evaluations used, and the vertical axes plot the lengths of the initial position (top left), initial Euler angles (top right), initial velocity (bottom left), and loss (bottom right).}
    \label{fig:supp:bunny}
\end{figure}

Fig.~\ref{fig:supp:bunny} shows the evolution of decision variables during optimization. From the viewpoint of optimal performance, we can see that Newton's methods achieve a lower final loss than DiffPD. None of them solved this task perfectly, i.e., ending the trajectory with the center of mass of the bunny stopping at the exact target. It is also worth noting that all three methods ended up providing quite different solutions to this task, which is not surprising as the end goal requires only the center of mass be aligned with the target without further constraints.

\begin{figure}[htb]
    \centering
    \includegraphics[width=0.85\linewidth]{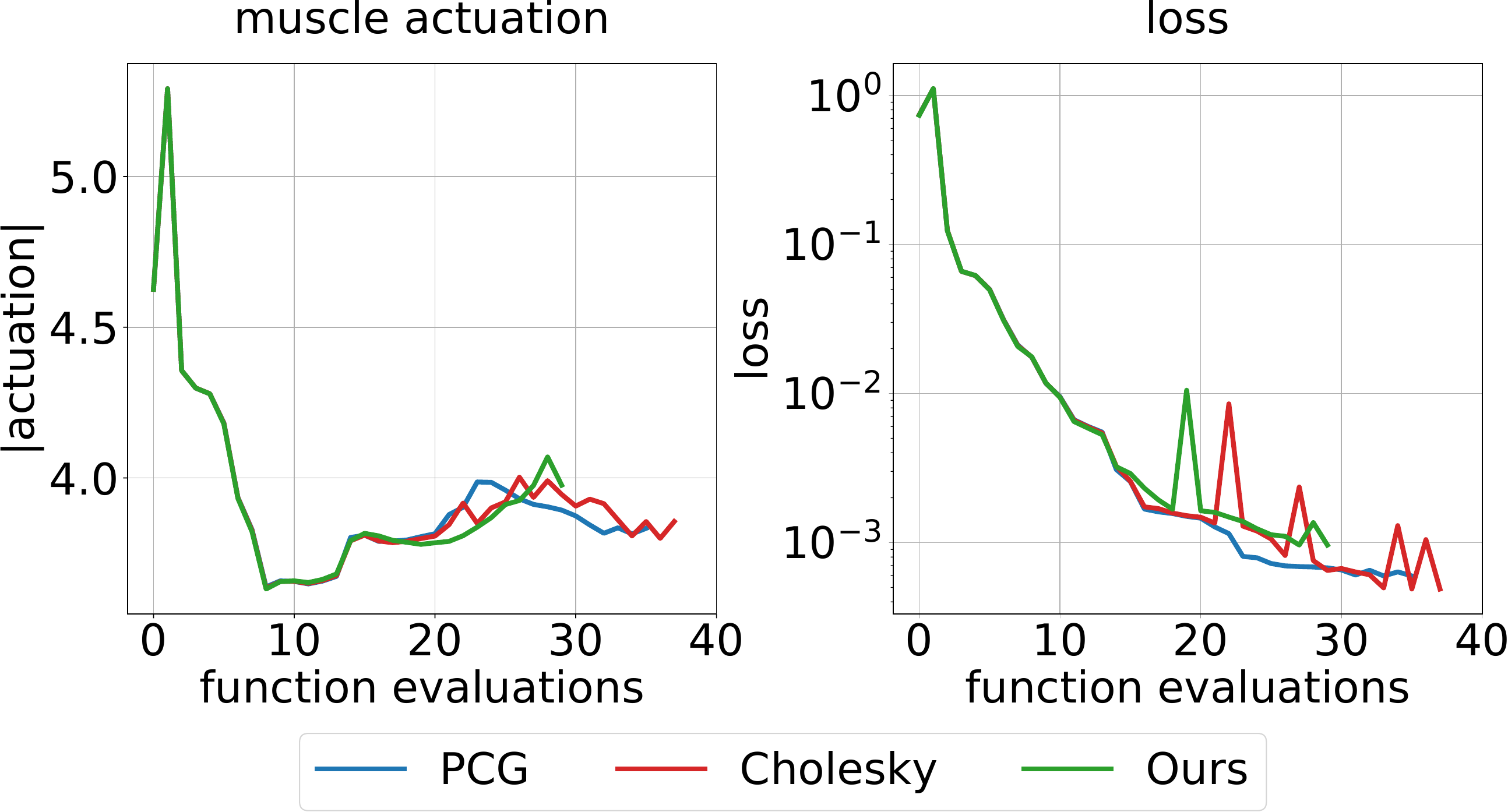}
    \caption{The progress of running L-BFGS optimization to solve the ``Routing tendon'' example. The horizontal axis shows the number of function evaluations used, and the vertical axes plot the magnitude of the actuation (left) and loss (right).}
    \label{fig:supp:tendon}
\end{figure}

\paragraph{Routing tendon} For this example, we report the optimization progress in Fig.~\ref{fig:supp:tendon}. This is an easy task to which all three methods provide satisfying solutions. In particular, all three methods converge to muscle actuation signals with very similar magnitudes.

\subsection{Motion Planning}

\paragraph{Torus} In this example, the torus is given a small horizontal velocity at the beginning of simulation to initiate the rolling motion. Without further optimization, however, the initial velocity will be damped quickly due to the infinite friction from our contact model, and the torus will stop near its original position. To roll itself forward, the torus needs to use its muscles to redistribute its mass in order to change the location of its center of mass. It can be seen from the video that the torus also needs to swing back and forth in order to gain enough momentum for forward motion.

\begin{figure}[htb]
    \centering
    \includegraphics[width=0.85\linewidth]{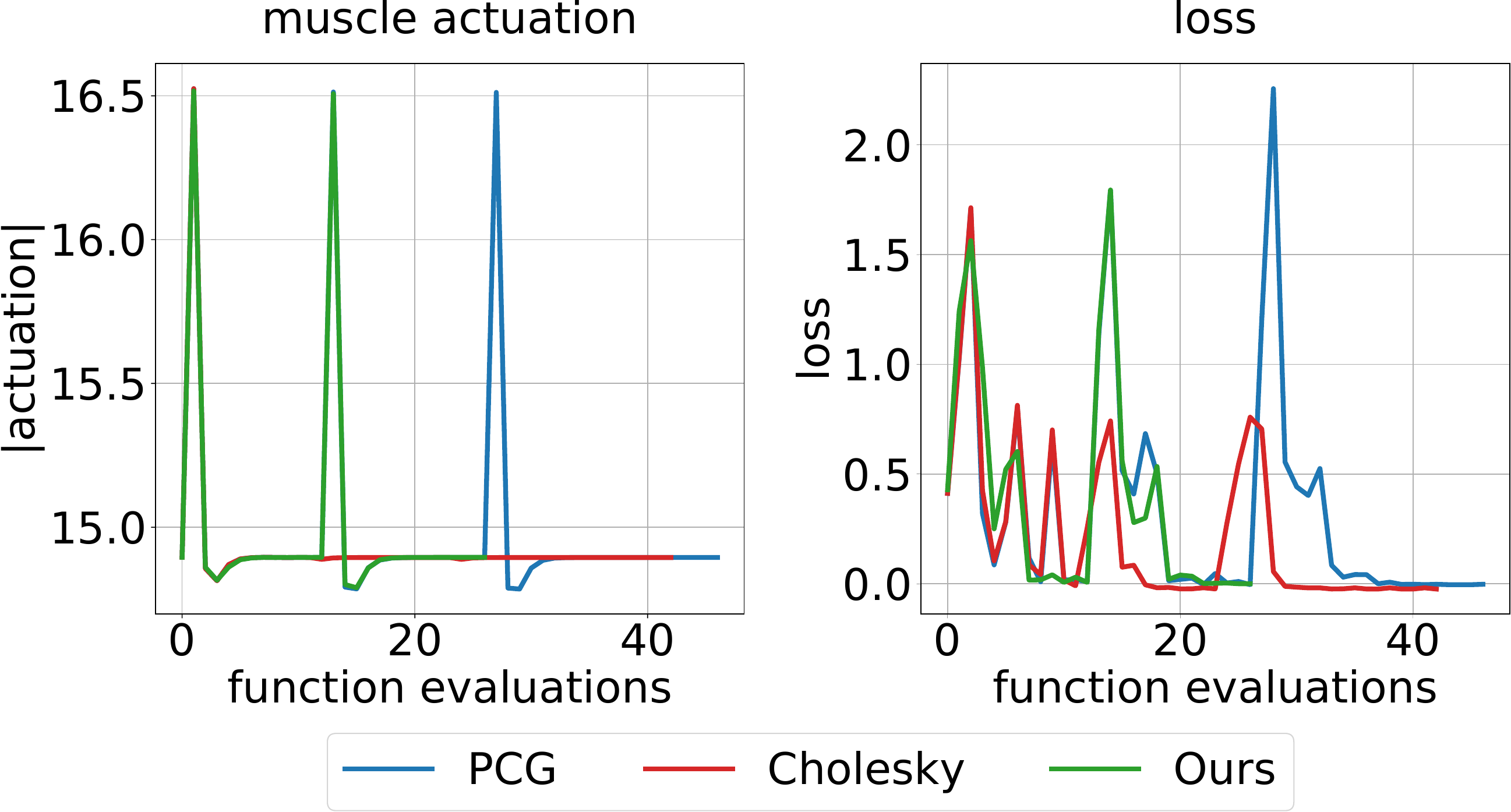}
    \caption{The progress of running L-BFGS optimization to solve the ``Torus'' example. The horizontal axis shows the number of function evaluations used, and the vertical axes plot the magnitude of the actuation (left) and loss (right).}
    \label{fig:supp:torus}
\end{figure}

Fig.~\ref{fig:supp:torus} reports the optimization progress in this example. During optimization, we noticed that the optimization result was more sensitive to the initial guess than previous examples, implying a challenging landscape of the loss function with more local minima than before.

\begin{figure}[htb]
    \centering
    \includegraphics[width=0.85\linewidth]{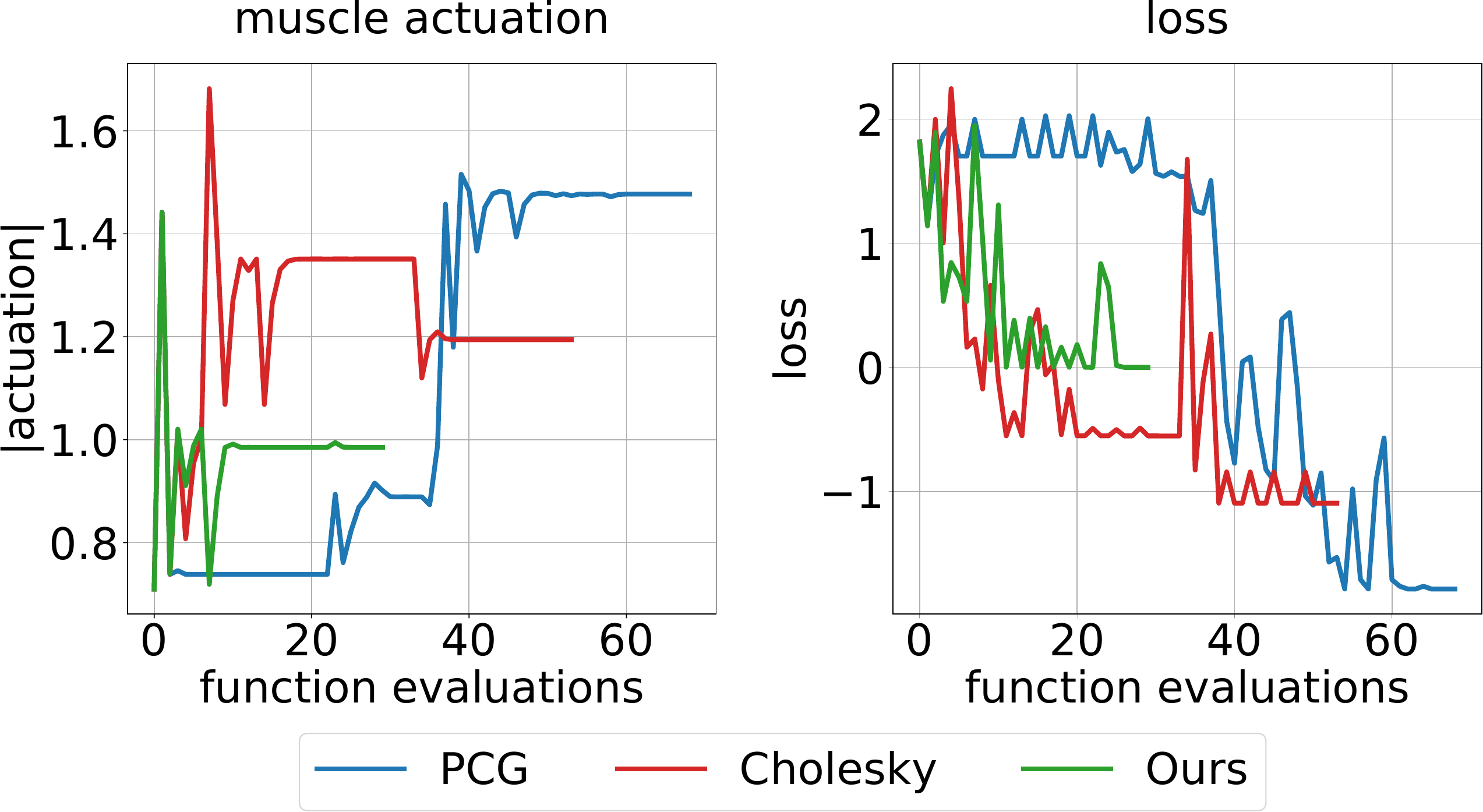}
    \caption{The progress of running L-BFGS optimization to solve the ``Quadruped'' example. The horizontal axis shows the number of function evaluations used, and the vertical axes plot the magnitude of the actuation (left) and loss (right).}
    \label{fig:supp:quadruped}
\end{figure}

\paragraph{Quadruped} For this example and the ``Cow'' example below, we consider optimizing an open-loop control sequence represented as a sinusoidal function. Unlike the ``Torus'' example above which is frequently trapped into local minima, optimizing the motion of this quadruped is much easier possibly due to two reasons: first, with its four legs at the four corners of the body, the quadruped is very stable and can hardly fall over; Second, the sinusoidal parameterization greatly reduces the dimension of the search space. Fig.~\ref{fig:supp:quadruped} displays the optimization progress, and we notice Newton's methods achieve better performance than DiffPD in this example.

\begin{figure}[htb]
    \centering
    \includegraphics[width=0.85\linewidth]{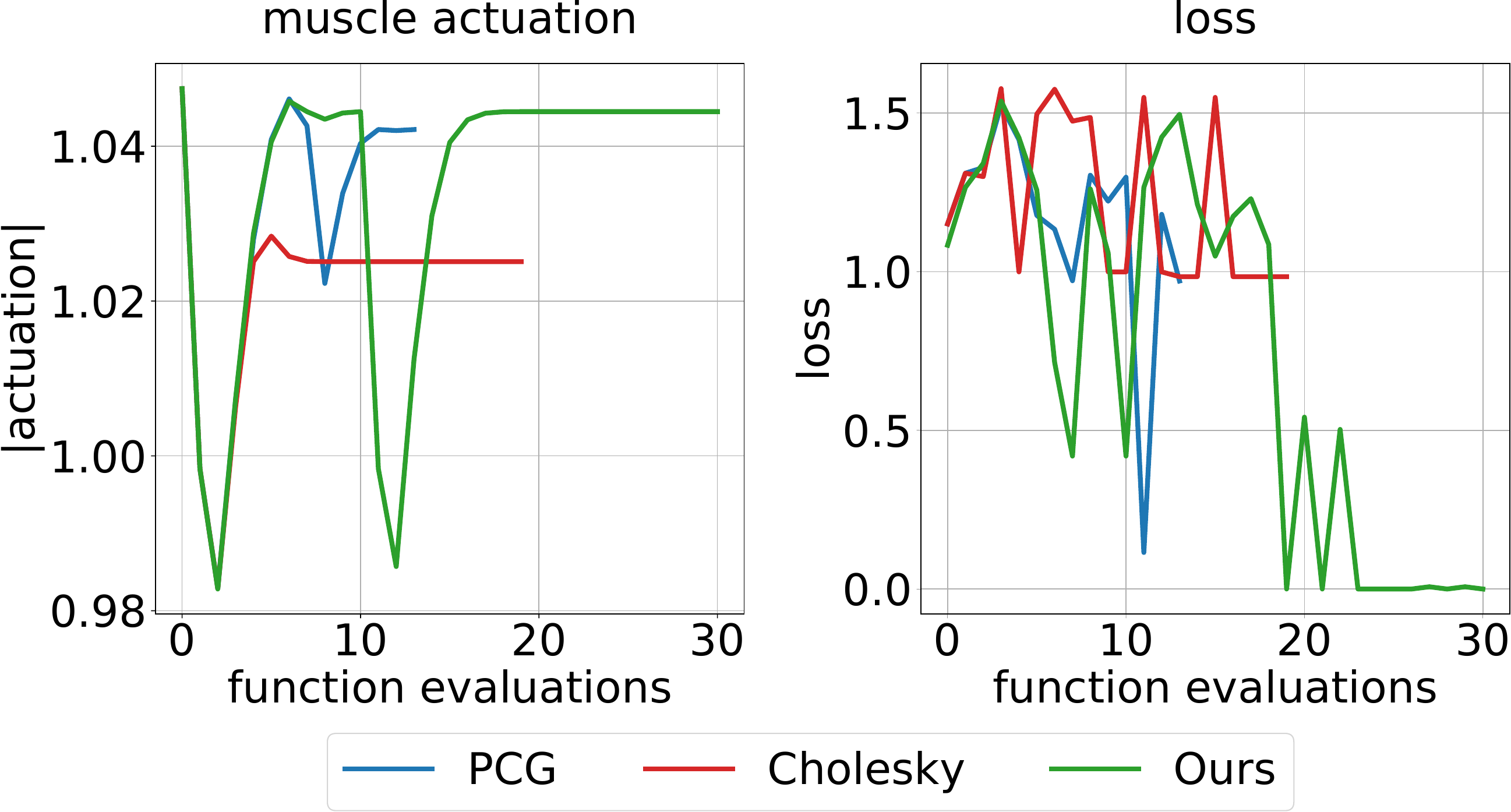}
    \caption{The progress of running L-BFGS optimization to solve the ``Cow'' example. The horizontal axis shows the number of function evaluations used, and the vertical axes plot the magnitude of the actuation (left) and loss (right).}
    \label{fig:supp:cow}
\end{figure}

\paragraph{Cow} The ``Cow'' example has a setup very similar to the previous example above except that it uses a different mesh model. The optimization progress can be found in Fig.~\ref{fig:supp:cow}

\subsection{Closed-Loop Control}

The neural network controller used in the ``Starfish'' and ``Shark'' examples takes as input the velocity of their centroid, the velocities of their limb endpoints, and the relative positions of their limb endpoints to their centroid. Each actuated limb (the arms of the starfish and the caudal fin of the shark) contains a pair of antagonistic muscle groups $m_{-}$ and $m_{+}$. At each frame $t$, a control signal is inferred from input states for each limb by a neural network parameterized with $\mathbf{\theta}$. We denote this control signal as $u_t$ for simplicity. Given $u_t$, the paired actuation signals $\mu_{m_{-}(t)}$ and $\mu_{m_{+}(t)}$ are generated by: 
\begin{align}
    \mu_{m_{-}(t)}=&\;\frac{1}{2}(\lvert u_t\rvert-u_t),\\
    \mu_{m_{+}(t)}=&\;\frac{1}{2}(\lvert u_t\rvert+u_t).
\end{align}
Within each muscle group, the actuation $a_i(t)$ is propagated along both the length of the limb $i=0,1,\cdots$ and time $t$:
\begin{equation}\label{eq:propagation}
\begin{aligned}
    a_i(t)=
    \begin{cases}
      \mu(t), & \text{if}\ i=0, \\
      a_{i-1}(t-1), & \text{otherwise}.
    \end{cases}
\end{aligned}
\end{equation}

\begin{figure}[htb]
    \centering
    \includegraphics[width=\linewidth]{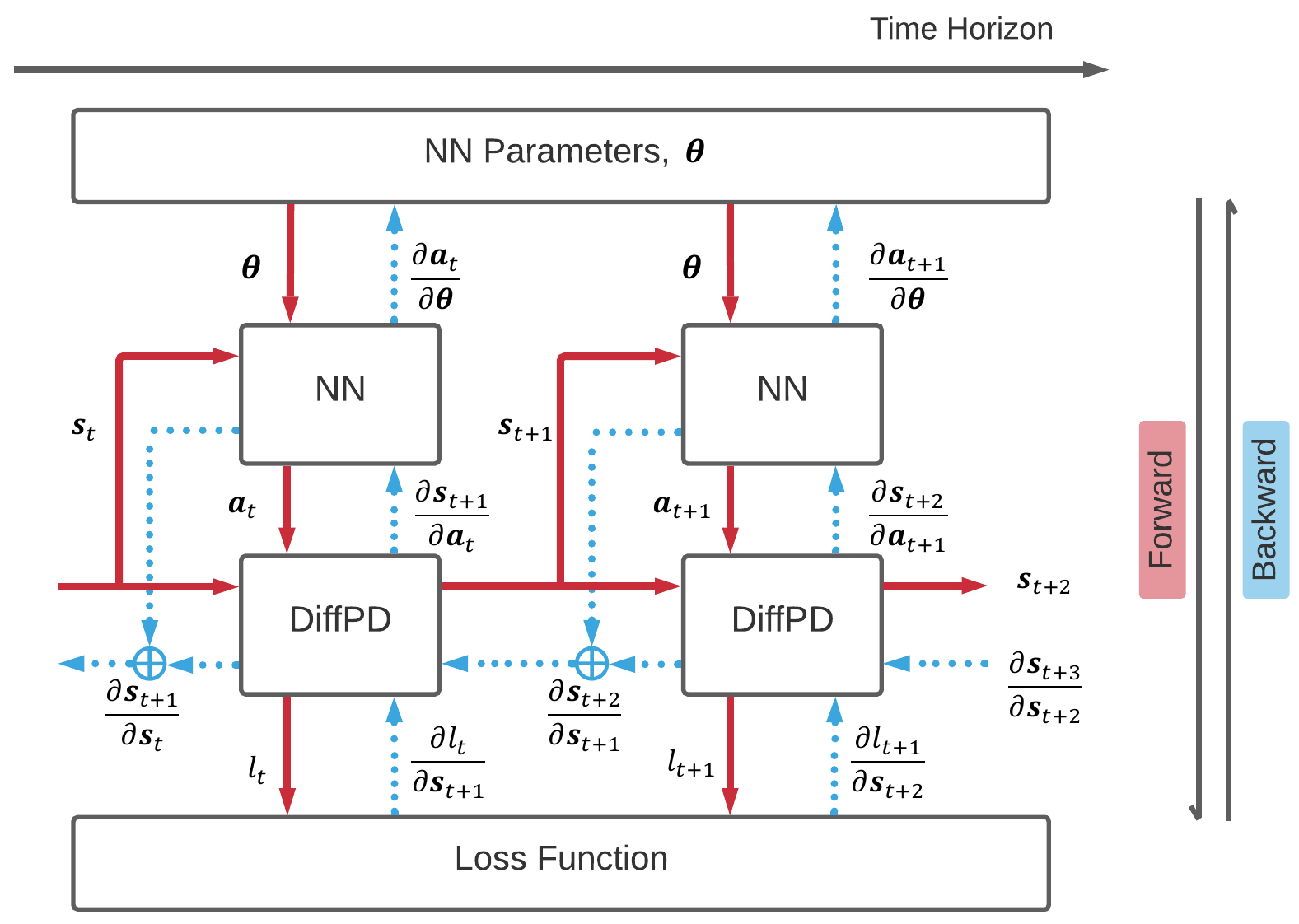}
    \caption{We present this diagram to show how the neural network (NN) controller and DiffPD are combined in forward simulation (red) and backpropagation (blue). $\mathbf{s}$ and $\mathbf{a}$ represent the state and action vectors respectively with their indices indicating the time step.}
    \label{fig:supp:nn}
\end{figure}

The goal of the controller is to cause a given marine creature to swim in a given direction as fast as possible. All creatures start from a rest pose with no speed. The loss function is composed of two parts: first, the dot product between the velocity of the centroid and a target directional unit vector to encourage movement along it; second, the norm of the cross product between those same vectors to penalize their deviation. With the loss function defined, Fig.~\ref{fig:supp:nn} outlines the function of the controller during forward and backward simulation. The optimization progress of DiffPD for both ``Starfish'' and ``Shark'' can be found in Fig. 16 in the main paper.

\bibliographystyle{ACM-Reference-Format}
\bibliography{reference}